\documentclass[preprint,12pt]{elsarticle}
\usepackage[pdfborder={0 0 0}]{hyperref} % from ther hyperlink 
\usepackage{amsthm,amsmath,amsfonts,latexsym,amssymb} %mathematic symbols 
\usepackage{graphics,fancyhdr,graphicx,subfigure} % for figures 
\usepackage[ruled,linesnumbered]{algorithm2e} % for algorithm insertion
\usepackage[margin=1in]{geometry}
\usepackage{multirow,booktabs} % consolidate table rows; table forms 
\usepackage{appendix}
\usepackage[table]{xcolor}
\usepackage{listings}
\usepackage{tabularx}
\usepackage{placeins}
\usepackage{comment}
\usepackage{lineno}
\usepackage{lscape}
\usepackage{bm}
\usepackage{hyperref}
\usepackage{mathrsfs}
\usepackage{arydshln}
\usepackage{appendix}
\usepackage{float} % the caption can float 
\usepackage{enumitem}
\usepackage{url}

\usepackage{makecell} %line width in table 
\usepackage[
singlelinecheck=false % <-- important
]{caption} % make caption at the left 
\usepackage{caption}
\usepackage{upgreek} % let greek upright
\newcommand*{\review}{\textcolor{black}}

\definecolor{custom-blue}{RGB}{3,69,173}
\hypersetup{
    colorlinks = true,
    urlcolor   = blue,
    citecolor  = custom-blue,
}

\biboptions{numbers,sort&compress} % for refenrece format 
\definecolor{listinggray}{gray}{0.9}
\definecolor{lbcolor}{rgb}{0.9,0.9,0.9}
\definecolor{Darkgreen}{RGB}{0,100,0}

\begin{document} % for all document 

\abovedisplayskip=6.0pt
\belowdisplayskip=6.0pt
\begin{frontmatter} % for the preface and abstract 

\title{Time-Marching Neural Operator–FE Coupling: AI-Accelerated Physics Modeling}

\author[1,2]{Wei Wang}
\ead{wwang198@jhu.edu}
\author[2]{Maryam Hakimzadeh}
\ead{mhakimz1@jhu.edu}
\author[1,3]{Haihui Ruan \corref{cor1}}
\ead{hhruan@polyu.edu.hk}
\author[2]{Somdatta Goswami\corref{cor1}}
\ead{sgoswam4@jhu.edu}

\address[1]{Department of Mechanical Engineering, The Hong Kong Polytechnic University}
\address[2]{Department of Civil and Systems Engineering, Johns Hopkins University}
\address[3]{PolyU-Daya Bay Technology and Innovation Research Institute}
\cortext[cor1]{Corresponding author.}

\begin{abstract}
\noindent \review{Numerical solvers for partial differential equations (PDEs) often struggle to balance computational efficiency with accuracy, especially in multiscale and time-dependent systems. Neural operators offer a promising avenue to accelerate simulations, but their practical deployment is hindered by several challenges: they typically require large volumes of training data generated from high-fidelity solvers, tend to accumulate errors over time in dynamical settings, and often exhibit poor generalization in multiphysics scenarios.} This work introduces a novel hybrid framework that integrates physics-informed deep operator network (PI-DeepONet) with finite element method (FEM) through domain decomposition and leverages numerical analysis for time marching. \review{The core innovation lies in efficient coupling FEM and DeepONet subdomains via a Schwarz alternating method, expecting to solve complex and nonlinear regions by a pre-trained DeepONet, while the remainder is handled by conventional FEM.} To address the challenges of dynamic systems, we embed the Newmark-Beta time-stepping scheme directly into the DeepONet architecture, substantially reducing long-term error propagation. Furthermore, an adaptive subdomain evolution strategy enables the ML-resolved region to expand dynamically, capturing emerging fine-scale features without remeshing. \review{The framework's efficacy has been rigorously validated across a range of solid mechanics problems---spanning static, quasi-static, and dynamic regimes including linear elasticity, hyperelasticity, and elastodynamics---demonstrating accelerated convergence rates (up to 20\% improvement in convergence rates compared to conventional FE coupling approaches) while preserving solution fidelity with error margins consistently below 3\%.} Our extensive case studies demonstrate that our proposed hybrid solver: (1) reduces computational costs by eliminating fine mesh requirements, (2) mitigates error accumulation in time-dependent simulations, and (3) enables automatic adaptation to evolving physical phenomena. \review{This work establishes a new paradigm for coupling state-of-the-art physics-based and machine learning solvers in a unified framework—offering a robust, reliable, and scalable pathway for high-fidelity multiscale simulations.}
\end{abstract}
\end{frontmatter}
\section{Introduction}
\label{sec:intro}

Partial differential equations (PDEs) are foundational to modeling a wide range of scientific and engineering phenomena. However, obtaining analytical solutions to these equations is often infeasible, especially for complex or nonlinear systems. As a result, numerical solvers have become indispensable tools. While highly accurate, these solvers can be computationally expensive, particularly when applied to multiscale, multiphysics systems. The solution accuracy of numerical methods is strongly influenced by mesh resolution - denser meshes yield higher accuracy but incur significantly higher computational costs. Moreover, many scientific and engineering applications - such as design optimization, sensitivity analysis, and uncertainty quantification - require repeated solutions of these equations for different parameter settings, further compounding the computational burden.

\review{ \textbf{Related prior work:} } Surrogate models have gained significant attention for their ability to emulate high-fidelity simulations at a fraction of the computational cost \cite{forrester2008engineering, bhosekar2018advances, samadian2024application, wang2024causality}. In the rapidly growing field of scientific machine learning (SciML), deep neural networks (DNNs) are increasingly used as emulators to analyze, solve, and optimize systems governed by PDEs \cite{raissi2019physics, cuomo2022scientific, huang2025partial}. These models are typically trained on a finite set of labeled data, generated from expensive traditional solvers such as finite difference methods, finite element methods (hereafter referred to as FE), or computational fluid dynamics methods. Once trained, these models offer fast and accurate predictions under varying initial conditions, boundary conditions, system parameters, and geometric configurations.

Among these surrogate approaches, neural operators, which learn mappings between infinite-dimensional function spaces, have emerged as a powerful class of emulators \cite{lu2021learning, goswami2023physics, fanaskov2023spectral, hao2023gnot}. However, neural operators often suffer from limited generalization capabilities unless they are trained as heavily overparameterized networks \cite{lee2024training, de2022generic, kovachki2021universal}. High-dimensional systems require models with large trainable parameters to capture the underlying physics, which can reduce their robustness and scalability. Additionally, these models struggle with multiscale, multiphysics problems, where capturing complex interactions across spatial and temporal scales is critical. They also tend to accumulate errors over long-time integrations, making them less reliable for dynamic systems \cite{michalowska2024neural, oommen2024rethinking, navaneeth2022koopman, yin2022interfacing}. 

\review{\textbf{Domain-Decomposition $+$ ML Models:}
Traditional numerical solvers and DNN-based surrogate models each offer distinct strengths, yet both face inherent limitations in scalability, reliability, and generalization when applied to large-scale, multiphysics problems. Domain decomposition methods (DDM) - long used to develop parallel, scalable solvers for PDEs - provide a promising framework to address these challenges \cite{chan1994domain, smith1997domain, dolean2015introduction, heinlein2021combining, snyder2023domain}. Non-overlapping substructuring techniques such as FETI (Finite Element Tearing and Interconnecting) \cite{farhat1991method, farhat2000scalable}, BDDC (Balancing Domain Decomposition by Constraints) \cite{dohrmann2003preconditioner}, and their boundary-element variants (BETI) \cite{langer2003boundary, langer2005coupled} have enabled scalable parallel PDE solvers for decades. These methods ``tear” the global mesh into independent subdomains and enforce inter-subdomain continuity via Lagrange multipliers or primal constraints, resulting in parallelizable, reduced interface systems. More recently, hybrid approaches that combine FE discretizations with neural network surrogates have gained traction \cite{mitusch2021hybrid, zhou2023transfer, yin2022interfacing, thel2024introducing, margenberg2024dnn, pantidis2023integrated}. For example, FEMIN \cite{thel2024introducing} replaces large portions of crash simulation meshes with neural networks that predict interface forces from kinematic data, significantly accelerating simulations. DNNMG \cite{margenberg2024dnn} employs neural networks to estimate fine-scale corrections to coarse FE solutions, enhancing the efficiency of nonlinear FE problems. I-FENN \cite{pantidis2023integrated} embeds physics-informed neural networks directly within the finite element stiffness matrix, enabling the solution of non-local damage mechanics problems at a computational cost comparable to local damage models. These hybrid models typically augment or replace specific components of FE solvers with learned surrogates, yielding substantial speedups in targeted regimes. However, most existing frameworks in the literature are demonstrated on static systems and fall into one of two categories: (i) purely data-driven models that require large volumes of high-fidelity simulation data for training before deployment, or (ii) task-specific models designed to solve a single downstream problem, rather than serving as general-purpose surrogates.}

\review{\textbf{{Contributions of this paper:}}
In this work, we introduce a domain decomposition (DDM)-based hybrid solver that couples traditional numerical methods with physics-informed operator learning surrogates to accelerate simulations while preserving high-fidelity accuracy. While DDMs have been widely applied in SciML for data parallelism, model parallelism, and pipeline optimization, our framework introduces a distinct paradigm: it tightly integrates pre-trained physics-informed neural operators with conventional solvers to enable seamless spatial coupling, rather than merely partitioning computational workloads. The central idea is to identify subdomains that are computationally intensive—due to fine-scale features or strong nonlinearities—and replace them with DeepONet-based neural operators. The remainder of the domain, where the physics can be accurately captured using coarser discretizations, is solved using standard finite element (FE) or numerical methods. These two solvers interact through an overlapping Schwarz-type coupling scheme \cite{morlet1997schwarz, mota2017schwarz, mota2022schwarz}, ensuring continuity and consistency across subdomain interfaces. We demonstrate the effectiveness of this hybrid framework on a series of elasticity problems: starting with a static linear elastic system and extending to more complex quasi-static hyperelastic scenarios. To address dynamic problems, we embed a classical time integrator—specifically the Newmark-Beta method—directly within the neural operator architecture \cite{fung1998complex, mohammadzadeh2014investigation}. This integration mitigates the issue of long-term error accumulation, a common shortcoming of purely data-driven models. Finally, we introduce an adaptive subdomain evolution strategy, where the ML-based subdomain expands dynamically in response to emerging fine-scale phenomena, such as stress concentrations. This reduces the reliance on dense meshing in critical regions, boosting computational efficiency while retaining accuracy.} Key innovations of our framework include: \vspace{-6pt}
\begin{itemize}
    \item Adaptive Coupling: A Schwarz alternating method facilitates efficient information exchange between FE and neural operator subdomains through overlapping interfaces, maintaining solution continuity while balancing accuracy and efficiency. \vspace{-6pt}
    \item Temporal Integration: A Newmark-Beta time-stepping scheme embedded within the neural operator to significantly reduce the long-time error accumulation in dynamic systems, addressing a key limitation of traditional data-driven ML approaches. \vspace{-6pt}
    \item Dynamic Subdomain Optimization: An adaptive algorithm allows the neural operator region to grow in response to evolving fine-scale features (e.g., stress localization), reducing dependence on dense mesh refinements and enhancing overall performance. \vspace{-6pt}
\end{itemize}

\section{Methodology}
\label{sec:methods}

This work focuses on solid mechanics problems under static, quasi-static, and dynamic loading conditions. The governing equations for momentum equilibrium over a computational domain $\Omega$ are given by:
\begin{equation} \text{Static or quasi-static:} 
\quad \nabla \cdot \boldsymbol{\sigma} + \mathbf{f} = 0, \label{static_equa} \end{equation}
\begin{equation} \text{Dynamic:} 
\quad \nabla \cdot \boldsymbol{\sigma} + \mathbf{f} = \rho \ddot{\boldsymbol{u}}, \label{dynamic_equa} \end{equation}
where $\boldsymbol{\sigma}$, $\mathbf{f}$, $\mathbf{u}$ and $\rho$ denote the stress tensor, body force, displacement vector, and density, respectively.

\noindent To develop a physics-informed coupling framework, we integrate neural operators - specifically Deep Operator Network (DeepONet) - with a finite elememet (FE) solver. All FE simulations are carried out using FEniCSx \cite{baratta2023dolfinx}, an open-source Python-based FE platform. The FE models utilize triangular meshes generated via Gmsh \cite{geuzaine2009gmsh}, and implement a first-order continuous Galerkin formulation \cite{logg2012automated} from Sections ~\ref{section_static} - \ref{section_puzzle}.

DeepONet is employed to learn nonlinear operators that map between infinite-dimensional function spaces defined over a bounded open set $\Omega^{*} \subset \mathbb{R}^{D}$, given a set of input-output function pairs. Let $\mathcal{U}$ and $\mathcal{S}$ be Banach spaces defined as:
\begin{equation} 
\mathcal{U} = \{\Omega^{*}; h: \mathcal{X} \rightarrow \mathbb{R}^{d_{x}}\}, \quad \mathcal{X} \subseteq \mathbb{R}^{d_{h}}, \label{U space} \end{equation}
\begin{equation} \mathcal{S} = \{\Omega^{*}; s: \mathcal{Y} \rightarrow \mathbb{R}^{d_{y}}\}, \quad \mathcal{Y} \subseteq \mathbb{R}^{d_{s}}, \label{S space} \end{equation}
where $\mathcal{U}$ and $\mathcal{S}$ denote the spaces of input and output functions, respectively. The goal is to approximate a nonlinear operator $G: \mathcal{U} \rightarrow \mathcal{S}$ via a parametric mapping:
\begin{equation} 
G: \mathcal{U} \times \boldsymbol{\Theta} \rightarrow \mathcal{S} \quad \text{or} \quad G_{\boldsymbol{\uptheta}}: \mathcal{U} \rightarrow \mathcal{S}, \quad \boldsymbol{\uptheta} \in \boldsymbol{\Theta}, \label{G mapping} 
\end{equation}
where $\boldsymbol{\Theta}$ is a finite-dimensional parameter space and $\boldsymbol{\uptheta}$ represents the tunable parameters (weights and biases) of the neural network.

To construct a physics-informed DeepONet (PI-DeepONet), we embed the governing Eqs. \eqref{static_equa} and \eqref{dynamic_equa} into the learning process to build static and dynamic surrogate models. In PI-DeepONet, physical systems governed by PDEs are described through multiple components: the PDE solution, forcing terms, and initial and boundary conditions. By minimizing a composite loss function incorporating these components, we learn the optimal parameters $\boldsymbol{\uptheta}^*$ to predict functions in $\mathcal{S}$, conditioned on variations in $\mathcal{U}$. This capability to learn function mappings governed by PDEs makes PI-DeepONe a powerful tool for building high-fidelity surrogate models.

We develop three distinct PI-DeepONet architectures - summarized in Table~\ref{tab:NO_architecture} - to handle static, quasi-static and dynamic regimes, as detailed in Sections~\ref{PI-DeepONet static} and \ref{Newmark PI-DeepONet}. The developed surrogate models are designed to be embedded into traditional numerical solvers using a DDM approach, as illustrated in Section~\ref{Domain decom}.

\subsection{Case 1: PI-DeepONet in static and quasi-static loading examples}
\label{PI-DeepONet static}

In static or quasi-static cases, the governing equations are time independent, meaning there is no inertia term, as shown in Eq. \ref{static_equa}. We consider two material models in this context: linear elasticity for small deformations and hyperelasticity for scenarios involving large deformations.

\noindent For linear elasticity with infinitesimal strains, the strain tensor is expressed as: 
\begin{equation}
 \boldsymbol{\varepsilon} = \frac{1}{2} \left(\nabla{\mathbf{u}} + \left(\nabla{\mathbf{u}}\right)^{T} \right), \label{strain}
\end{equation}
with the corresponding stress given by:
\begin{equation}
 \boldsymbol{\sigma} = \lambda \text{tr}(\boldsymbol{\varepsilon}) \boldsymbol{I}+ 2\mu \boldsymbol{\varepsilon}, \label{stress}
\end{equation}
where $\lambda$ and $\mu$ are the Lam\'e constants that characterize the elastic properties of the material. Substituting Eqs. \ref{strain} and \ref{stress} into Eq. \ref{static_equa} yields the strong form of the governing equations for static problems.
When solved via the standard FEM, these continuous equations are transformed into the algebraic system
\begin{equation}
\mathbf{K} \;\mathbf{u} = \mathbf{F}, \nonumber
\end{equation}
where $\mathbf{K}$ is the global stiffness matrix, $\mathbf{F}$ is the global external force vector, and $\mathbf{u}$ now denotes the vector of nodal displacement unknowns in the FE discretization. By notational reuse, we use the same bold symbol $\mathbf{u}$ for both the continuous displacement field and its discrete nodal representation; the intended meaning should be clear from the context.

\noindent For hyperelasticity, we must account for large deformations within the framework of continuum mechanics. In this case, the stress term in Eq. \ref{static_equa} is replaced by the first Piola-Kirchhoff stress tensor, $\mathbf{P_{1}}$, which normalizes forces in the current configuration over areas in the reference configuration. The key kinematic quantities are defined as follows:
\begin{align}
\mathbf{F}_g &= \mathbf{I} + \nabla{\mathbf{u}}, \label{F_grad} \\
\mathbf{C} &= \mathbf{F}_g^T \mathbf{F}_g, \label{RCG}\\
\mathbf{I}_{1} &= \text{tr}(\mathbf{C}), \label{first_invariant}\\ 
J &= \det{\mathbf{F}_g}, \label{volume change factor}
\end{align}
where $\mathbf{F}_g$ represents the deformation gradient tensor, $\mathbf{C}$ is the right Cauchy-Green deformation tensor, $\mathbf{I}_{1}$ denotes the first invariant of $\mathbf{C}$, and $J$ is the volume change factor (determinant of $\mathbf{F}_g$). In this work, we utilize the Neo-Hookean hyperelastic model with the strain energy density function expressed as:
\begin{equation}
 \Psi = \frac{1}{2}\mu (\mathbf{I}_{1} - 3) + \frac{\lambda}{2} \ln{J}^2 - \mu \ln{J}.\label{strain_energy}   
\end{equation}
The first Piola-Kirchhoff stress tensor is then derived through:
\begin{equation}
\mathbf{P_{1}} = \frac{\partial{\Psi}}{\partial{\mathbf{F}_g}}. \label{first_piola_kirchhoff}  
\end{equation}

For both linear elasticity and hyperelasticity in static or quasi-static conditions, we implement the conventional PI-DeepONet architecture as outlined in Table \ref{tab:NO_architecture}. This structure comprises two fully connected neural networks (FNNs): one functioning as the branch network and the other as the trunk network \cite{goswami2023physics}, as illustrated in Fig. \ref{Fig:DeepONet_structure}(a). For static and quasi-static loading scenarios, only the `Branch1' network is utilized, while the `Branch2' network shown in Fig. \ref{Fig:DeepONet_structure}(a) is exclusively used for dynamic models.
In the branch network (labeled `Branch1' in Fig. \ref{Fig:DeepONet_structure}(a)), we use the Dirichlet boundary condition $\mathbf{u}_{|\partial{\Omega}}$ as input functions from the space $\mathcal{U}$. To ensure robust generalization capabilities, these input functions are generated using the Gaussian Random Field (GRF) approach \cite{lu2021learning}. The trunk network (labeled Trunk in Fig. \ref{Fig:DeepONet_structure}(a)) takes spatial coordinates $\mathbf{x} \in \mathbb{R}^{d}$ as inputs, where $d$ represents the spatial dimensionality of the model.

By minimizing loss functions derived from the PDE residual loss and boundary conditions, we determine the optimal parameters to establish accurate neural operators for static problems, which we refer to as static PI-DeepONet. The specific PDEs and boundary conditions for linear elasticity and hyperelasticity are presented in Sections \ref{section_static} and \ref{section_quasi_static}, respectively. The training methodology for static PI-DeepONet in both linear elasticity and hyperelasticity contexts will be detailed in Section \ref{Newmark PI-DeepONet}.

\subsection{Case 2: PI-DeepONet in dynamic loading examples}
\label{Newmark PI-DeepONet}

PI-DeepONet can be extended beyond static problems to model time-dependent behaviors in structural dynamics by introducing time discretization in the trunk network. However, this approach has demonstrated the accumulation of errors for the problem over long time horizons \cite{michalowska2024neural}. To resolve the error accumulation issue, we propose a time-marching framework, which is inspired by the Newmark-Beta method \cite{newmark1959method} (hereafter referred to as the Newmark method for brevity), a widely-used time-discretization scheme in FE analysis. This hybridization leads to a time-advancing surrogate model that leverages concepts from numerical analysis, enabling the prediction of dynamic responses such as displacement, velocity, and acceleration fields across space and time. In this section, we formulate this dynamic PI-DeepONet, drawing on principles from FE solvers and convolutional architectures to encode spatial information efficiently.

Neglecting damping, the semi-discrete elastodynamic equations over a computational domain $\Omega$ are expressed using the Newmark method as:
\begin{align}
\mathbf{\dot{u}}^{n} &= \mathbf{\dot{u}}^{n-1} + (1-\gamma)\Delta{t}\mathbf{\ddot{u}}^{n-1} + \gamma\Delta{t}\mathbf{\ddot{u}}^{n}, \label{U_dot} \\
\mathbf{u}^{n} &= \mathbf{u}^{n-1} +\Delta{t}\mathbf{\dot{u}}^{n-1} + \frac{\Delta{t}^{2}}{2}\left((1-2\beta)\mathbf{\ddot{u}}^{n-1} + 2\beta\mathbf{\ddot{u}}^{n}\right), \label{U}
\end{align}
\begin{equation}
\mathbf{M}\mathbf{\ddot{u}}_{FE}^{n} + \mathbf{K} \mathbf{u}_{FE}^{n} = \mathbf{F}^{n} , \label{K_U=f}
\end{equation}
where $n-1$ and $n$ indicate the previous and current time steps, respectively; $\mathbf{u}$, $\mathbf{\dot{u}}$, and $\mathbf{\ddot{u}}$ are the displacement, velocity, and acceleration vectors; $\Delta{t}$ is the time increment; \review{$\mathbf{M}$ is the mass matrix;} and $\gamma, \beta$ are Newmark parameters with typical values in the range $\gamma \in [0,1]$, $\beta \in [0, \frac{1}{2}]$. To simplify, we use $\gamma = \beta = \frac{1}{2}$, resulting in:
\begin{align}
\mathbf{\dot{u}}^{n} &= \mathbf{\dot{u}}^{n-1} + \frac{\Delta{t}}{2}\left(\mathbf{\ddot{u}}^{n-1} +  \mathbf{\ddot{u}}^{n}\right), \label{U_dot_sim} \\
\mathbf{u}^{n} &= \mathbf{u}^{n-1} +\Delta{t}\mathbf{\dot{u}}^{n-1} + \frac{\Delta{t}^{2}}{2}\mathbf{\ddot{u}}^{n}, \label{U_sim}
\end{align}
From Eq. \ref{U_sim}, we can isolate acceleration, $\mathbf{\ddot{u}^{n}}$ as:
\begin{equation}
 \mathbf{\ddot{u}}^{n}= \frac{2}{\Delta{t}^{2}}(\mathbf{u}^{n} - \mathbf{u}^{n-1}) - \frac{2}{\Delta{t}}\mathbf{\dot{u}}^{n-1}   , \label{U_new}
\end{equation}
Substituting this into Eq. \ref{K_U=f} (assuming zero external force) yields:
\begin{equation}
\left(\mathbf{M}\frac{2}{\Delta{t}^{2}} + \mathbf{K} \right)\mathbf{u}_{FE}^{n} - \mathbf{M}\frac{2}{\Delta{t}^{2}} \mathbf{u}_{FE}^{n-1} - \frac{2}{\Delta{t}}\mathbf{\dot{u}}_{FE}^{n-1} = \mathbf{0}.\label{K_U=f_new}
\end{equation}
Here, the term $\left(\mathbf{M}\frac{2}{\Delta{t}^{2}} + \mathbf{K} \right)$ is recognized as the effective stiffness matrix.

\begin{figure}[H]
\begin{center}
\includegraphics[width=0.8\textwidth]{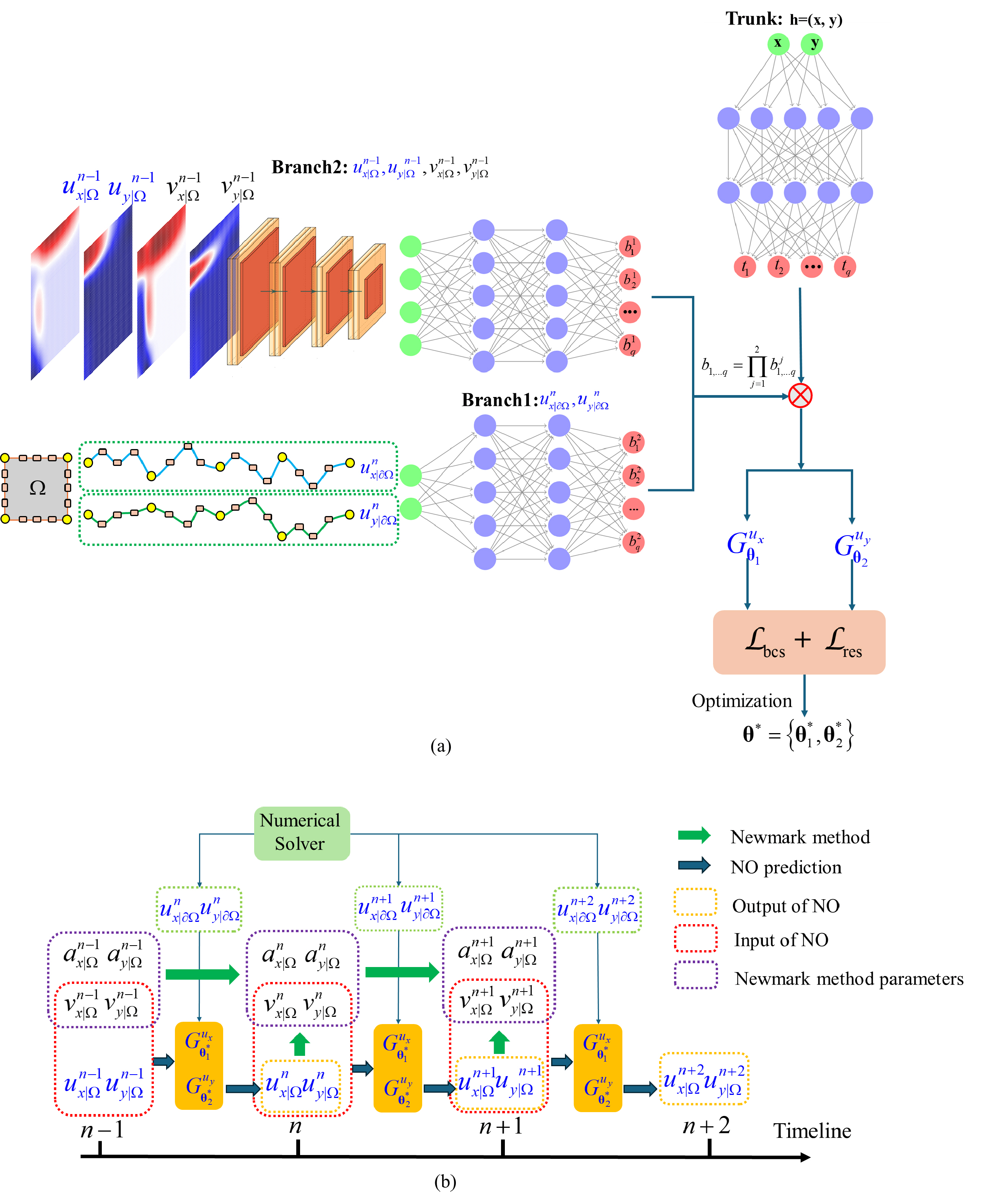}
\caption{Time-advancing DeepONet: (a) The DeepONet structure: Branch1 is the branch network with input functions of the Dirichlet boundary condition(\review{$u^{n}_{x|\partial\Omega}, u^{n}_{y|\partial\Omega}$}) from square edges at time step $n$; Branch2 for input functions of the displacement and velocity (\review{$u^{n-1}_{x|\Omega}, u^{n-1}_{y|\Omega}, v^{n-1}_{x|\Omega}, v^{n-1}_{y|\Omega}$}) over the entire domain $\Omega$ at time step $n-1$; Each branch net returns features embedding $\left[b^{i}_{1},b^{i}_{2}, ..., b^{i}_{q}\right]^{T} \in \mathbb{R}^{q}$ with $i = 1,2$ as output , and their dot product obtains  $\left[b_{1},b_{2}, ..., b_{q}\right]^{T}$; Trunk denotes the trunk networks taking the spatial coordinates $(x,y) \in \Omega$ as input, returning $\left[t_{1},t_{2}, ..., t_{q}\right]^{T} \in \mathbb{R}^{q}$  as output. Two identical sets of branch and trunk networks are merged via two element-wise dot products to output two solution operators \review{${G^{u_x}_{\boldsymbol{\uptheta}_1}}$} and \review{$G^{{u_y}}_{\boldsymbol{\uptheta}_2}$}, predicting \review{$u^{n}_{x|\Omega}$} and \review{$u^{n}_{y|\Omega}$}. Given the PDEs and boundary conditions, the residual loss ($\mathcal{L}_{res}$) and boundary conditions loss ($\mathcal{L}_{bcs}$) are defined. The optimal parameters $\boldsymbol{\uptheta}^*$ are obtained by minimizing the addition of these two loss functions; (b) Flowchart of integrating time-advancing DeepONet and Numerical solver in structural dynamics: the green arrow denotes the Newmark method predicting process, blue arrow indicates NO prediction, the orange-dashed and red-dashedbox denote the output and input of NO, and purple-dashed box shows the parameters needed for Newmark method.}
\label{Fig:DeepONet_structure}
\end{center}
\end{figure}

In standard FE methods, Eq. \ref{K_U=f_new} is solved for displacement $\mathbf{u}^n$, with velocity and acceleration subsequently updated via Eq. \ref{U_dot_sim} and \ref{U_new}. Instead, we embed this equation into the DeepONet architecture to form a time-marching PI-DeepONet. This modified architecture introduces a second branch network (`Branch2' as shown in Fig. \ref{Fig:DeepONet_structure}(a)), in addition to Branch1. Branch2 encodes prior-step displacement $\mathbf{u}^{n-1}$ and velocity $\mathbf{\dot{u}}^{n-1}$ across the domain $\Omega$. To extract domain-wide features efficiently, multi-channel CNNs are used in Branch2, as shown in Fig. \ref{Fig:DeepONet_structure}(a). CNNs not only reduce computational cost compared to FNNs, but also naturally align with the local interaction principles of the FE method-i.e., only neighboring nodes significantly affect computations at a given node.  We combine outputs from both branches (Branch1 FNNs and Branch2 CNN + FNNs) with the trunk network to form two operators, \review{$G^{u_x}_{\boldsymbol{\uptheta}_1}$} and \review{$G^{u_y}_{\boldsymbol{\uptheta}_2}$}, for displacements in $x$- and $y$-direction. \review{Although $G^{u_x}_{\boldsymbol{\uptheta}_1}$ and $G^{u_y}_{\boldsymbol{\uptheta}_2}$ share an identical DeepONet structure, they are parameterized by independent weight and bias sets ($\boldsymbol{\uptheta}_1 \neq \boldsymbol{\uptheta}_2$), enabling them to learn distinct mappings for $u_x$ and $u_y$.} The network is trained by minimizing the residual and boundary losses, $\mathcal{L}_{res}$ and $\mathcal{L}_{bcs}$, to accurately learn displacement fields at each time step without needing to include time in the trunk network. 

A schematic of this architecture and the prediction workflow with numerical solvers is shown in Fig. \ref{Fig:DeepONet_structure}(b). The trained PI-DeepONet receives inputs from the previous time step (red-dashed box) and current boundary displacements obtained from numerical solver (green-dashed box) to predict the full-field displacement (orange-dashed box). Using Newmark integration (green arrow), the velocity and acceleration are then inferred (purple-dashed box), enabling recursive forward prediction. This work is the first to incorporate classical time-advancing schemes into DeepONet, offering a novel route for hybrid time-marching surrogate modeling. Moreover, this framework can be extended to other time-integration methods such as the generalized-$\alpha$ method \cite{erlicher2002analysis} or Runge-Kutta schemes \cite{cartwright1992dynamics}. The concept can also be adapted to construct fully data-driven, time-advancing DeepONet.

%\begin{figure}[H]\ContinuedFloat
%\caption{(continued) Two identical sets of branch and trunk networks are merged via two element-wise dot products to output two solution operators ${G^{{u}}_{\boldsymbol{\uptheta}_1}}$ and $G^{{v}}_{\boldsymbol{\uptheta}_2}$, predicting $u^{n}_{|\Omega}$ and $v^{n}_{|\Omega}$. Given the PDEs and boundary conditions, the residual loss ($\mathcal{L}_{res}$) and boundary conditions loss ($\mathcal{L}_{bcs}$) are defined. The optimal parameters $\boldsymbol{\uptheta}^*$ are obtained by minimizing the addition of these two loss functions; (b) Flowchart of integrating time-advancing DeepONet and Numerical solver in structural dynamics: the green arrow denotes the Newmark method predicting process, blue arrow indicates NO prediction, the orange-dashed and red-dashedbox denote the output and input of NO, and purple-dashed box shows the parameters needed for Newmark method.}
%\label{Fig:DeepONet_structure_part2}
%\end{figure}

\noindent From a mathematical standpoint, both static and dynamic PI-DeepONet frameworks reduce to boundary value problems (BVPs) across spatial domains:
\begin{align}
    \mathcal{N}[s(\mathbf{x})](\mathbf{x}) &= f(\mathbf{x}),  s \in \mathcal{S}, \mathbf{x} \in \Omega, \label{Eq_res}\\
    \mathcal{B}[h(\mathbf{x})](\mathbf{x}) &= g(\mathbf{x}), h \in \mathcal{U}, \mathbf{x} \in \partial\Omega, \label{Eq_bcs}
\end{align}
where $\mathcal{N}$ and $\mathcal{B}$ are spatial differential operators, $f:\Omega \rightarrow \mathbb{R}^{d}$ is the driving force, $g: \partial\Omega \rightarrow \mathbb{R}^d$ is the boundary condition with the dimension $d$.  In this work, the driving force (i.e., body force) is not considered, all the boundary conditions are Dirichlet boundary conditions, and mechanical system is 2 dimensional. Thus, the $G(h, \boldsymbol{\uptheta})$ can be represented as:
\begin{equation}
    G(h, \boldsymbol{\uptheta}) = \{G(u_x, \boldsymbol{\uptheta}_1), G(u_y, \boldsymbol{\uptheta}_2)\},
\end{equation}
where \review{$u_x$} and \review{$u_y$} are $x$-displacement and $y$-displacement with \review{$h = (u_x,u_y)$}. For simplification, \review{$G(u_x, \boldsymbol{\uptheta}_1)$} and \review{$G(u_y,\boldsymbol{\uptheta}_2)$} are denoted as \review{$G^{u_x}_{\boldsymbol{\uptheta}_1}$} and \review{$G^{u_y}_{\boldsymbol{\uptheta}_2}$}.Consequently, Eq. \ref{Eq_res} and Eq. \ref{Eq_bcs} are expressed as:
\begin{align}
    \mathcal{N}_1\left(G^{u_x}_{\boldsymbol{\uptheta}_1},   G^{u_y}_{\boldsymbol{\uptheta}_2}\right) (\mathbf{x}) &= \mathbf{0},  \mathbf{x} \in \Omega, \boldsymbol{\uptheta}=\{\boldsymbol{\uptheta}_1, \boldsymbol{\uptheta}_2\} \in \mathbf{\Theta}, \label{Eq_res_new1}\\
    \mathcal{N}_2\left( G^{u_x}_{\boldsymbol{\uptheta}_1},   G^{u_y}_{\boldsymbol{\uptheta}_2}\right)(\mathbf{x}) &= \mathbf{0},  \mathbf{x} \in \Omega, \boldsymbol{\uptheta}=\{\boldsymbol{\uptheta}_1, \boldsymbol{\uptheta}_2\} \in \mathbf{\Theta}, \label{Eq_res_new2}\\
  G^{u_x}_{\boldsymbol{\uptheta}_1}(\mathbf{x}) &= s_1(\mathbf{x}), \mathbf{x} \in \partial\Omega , s= \{s_1, s_2\} \in \mathcal{S}, \label{Eq_bcs_new1} \\
 G^{u_y}_{\boldsymbol{\uptheta}_2}(\mathbf{x}) &= s_2(\mathbf{x}), \mathbf{x} \in \partial\Omega ,  s= \{s_1, s_2\} \in \mathcal{S}, \label{Eq_bcs_new2}
\end{align}
where $\mathcal{N}_1$ and  $\mathcal{N}_2$ are the spatial differential operators in $x$ and $y$-directions, respectively. It is noteworthy that Eq. \ref{Eq_res_new1}-\ref{Eq_bcs_new2} are the loss functions of the PI-DeepONet and can be defined as:
\begin{align}
    \mathcal{L}(\boldsymbol{\uptheta}) &= \mathcal{L}_{res}(\boldsymbol{\uptheta}) + \mathcal{L}_{bcs}(\boldsymbol{\uptheta}) \nonumber\\
    & = \frac{1}{N^r_1}\sum^{N^r_1}_{i=1}\left( \mathcal{N}_1( G^{u_x}_{\boldsymbol{\uptheta}_1},   G^{u_y}_{\boldsymbol{\uptheta}_2})(\mathbf{x}_r^i)\right)^2 + \frac{1}{N^r_2}\sum^{N^r_2}_{i=1}\left( \mathcal{N}_2( G^{u_x}_{\boldsymbol{\uptheta}_1},   G^{u_y}_{\boldsymbol{\uptheta}_2})(\mathbf{x}_r^i)\right)^2  \nonumber\\
&+\frac{1}{N^{b}_1}\sum^{N^b_1}_{i=1}\left( G^{u_x}_{\boldsymbol{\uptheta}_1}(\mathbf{x}_b^i) - s(\mathbf{x}_b^i)\right)^2+\frac{1}{N^{b}_2}\sum^{N^b_2}_{i=1}\left( G^{u_y}_{\boldsymbol{\uptheta}_2}(\mathbf{x}_b^i) - s(\mathbf{x}_b^i)\right)^2\label{loss_func},   \\
    \vspace{6pt}
    \text{for}\quad & \{\mathbf{x}^i_r\}^{N^r_1}_{i=0} \subsetneq \Omega,  \{\mathbf{x}^i_b\}^{N^b_1}_{i=0} \subsetneq \partial\Omega, N^r_1 = N^r_2, N^b_1 = N^b_2, \nonumber
\end{align}
where $\mathcal{L}_{res}$ and $\mathcal{L}_{bcs}$ are residual loss derived from Eq. \ref{Eq_res_new1}-\ref{Eq_res_new2} and boundary condition loss derived from \ref{Eq_bcs_new1}-\ref{Eq_bcs_new2};  $N^r_1$, $N^r_2$, $N^b_1$ and $N^b_2$ denote the number of collocation points and boundary points. By minimizing the loss function, the optimal parameter $\boldsymbol{\uptheta}^* =\{ \boldsymbol{\uptheta}^*_1, \boldsymbol{\uptheta}^*_2\}$ can be obtained as:
\begin{equation}
    \boldsymbol{\uptheta}^* = \arg\min_{\boldsymbol{\uptheta}} (\mathcal{L}).
\end{equation}

\subsection{DDM and coupling strategy}
\label{Domain decom}

We introduce a hybrid computational framework that couples a traditional FE solver with a PI-DeepONet to enable efficient and accurate simulations of complex physical phenomena. The coupling is achieved via a DDM approach based on the Schwarz alternating method, which facilitates iterative information exchange across overlapping subdomains in static and dynamic settings \cite{yu2018partitioned, yin2022interfacing}. To the best of our knowledge, this is the first study to demonstrate the integration of a PI-DeepONet with an FE solver, forming a cohesive FE–Neural Operator (FE–NO) architecture capable of simulating dynamic physical systems. While we present the methodology in a two-dimensional (2D) context, the approach is readily extendable to three dimensions.

\begin{figure}[H]
\begin{center}
\includegraphics[width=0.8\textwidth]{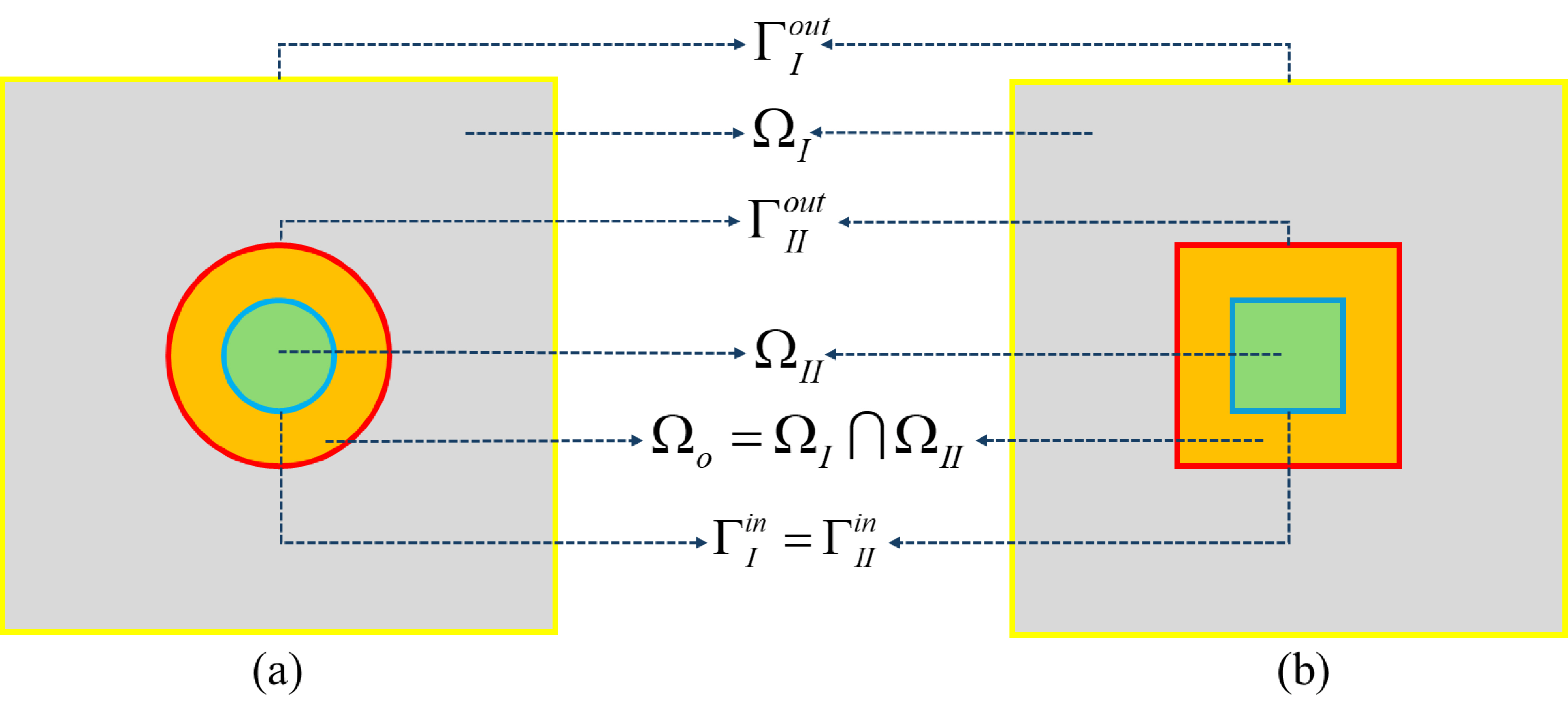}
\caption{Schematics of DDM: $\Omega_{I}$ denotes the gray region and orange region, $\Omega_{II}$ denotes the green region and the orange region, ($\Omega_{o}$) is the overlapping region in orange. $\Gamma_{I}^{out}$ (yellow lines) and $\Gamma_{II}^{out}$ (red lines) denote the outer boundary of $\Omega_{I}$ and $\Omega_{II}$ , respectively. $\Gamma_{II}^{in} = \Gamma_{II}^{out}$ (blue lines) is the inner boundary lines for both $\Omega_{I}$ and $\Omega_{II}$.}
\label{Fig:Schematics}
\end{center}
\end{figure}

The computational domain $\Omega \subset \mathbb{R}^2$ is decomposed into two overlapping subdomains, $\Omega_I$ and $\Omega_{II}$, such that $\Omega = \Omega_I \cup \Omega_{II}$ and the overlapping region is defined as $\Omega_o = \Omega_I \cap \Omega_{II}$. Figure~\ref{Fig:Schematics} illustrates the decomposition, where $\Omega_I$ (gray + orange) is governed by the FE model, and $\Omega_{II}$ (green + orange) is modeled by the PI-DeepONet.

We denote the interface boundary between subdomains as $\Gamma_{II}^{in}$ (blue line), the external boundary of  $\Omega_{II}$ as $\Gamma_{II}^{out}$ (red line), and the external boundary of $\Omega_I$ as  $\Gamma_{I}^{out}$ (yellow line).

Two configurations are considered in this work: Figure~\ref{Fig:Schematics}(a) corresponds to the static and quasi-static settings involving linear elastic and hyperelastic materials, where $\Omega_{II}$ is circular. Figure~\ref{Fig:Schematics}(b) depicts the dynamic setting involving elastodynamics, with a square-shaped $\Omega_{II}$. Importantly, this framework allows $\Omega_{II}$ to assume arbitrary geometry, even in cases where the interface $\Gamma_{I}^{in}$ lacks smoothness (i.e., $\Gamma_{I}^{in} \notin C^1(\mathbb{R})$), as shown in the dynamic case.
The algorithm of the Schwarz alternating method at the overlapping boundary in quasi-static or dynamic mode is illustrated in Algorithm \ref{algo_1}. The core of the coupling strategy lies in the Schwarz alternating method, outlined in Algorithm~\ref{algo_1}. 
At each iteration, the FE and NO models alternately solve their respective subdomains. Each solver uses the most recent interface data from the other, which ensures consistency and continuity across the overlapping region. Relaxation parameter (denoted as $\boldsymbol{\rho}_r$) is introduced at the interface to stabilize the iteration and accelerate convergence.

\begin{algorithm}[H]
\SetAlgoLined
\caption{Schwarz alternating method at overlapping boundary in the coupling framework}
\label{algo_1}
\textbf{Initialization}: Set $\mathbf{u}_{NO}^{0} = \mathbf{0}$ in $\Omega_{II}$ and $\mathbf{u}_{FE}^{0} = \mathbf{0}$ in $\Omega_{I}$\;

\textbf{Main loop}:\\
\textbf{for} $n = 0 : n_{max} - 1$ \textbf{do}:\\
\hspace*{1em}Set $j = 0$, and $\epsilon$ a critical value\\
\hspace*{1em}\textbf{while} $j \geq 0$ \textbf{do}:\\
\hspace*{2em} $j = j+1$\\
\hspace*{2em}\textbf{Model FE}:\\
\hspace*{3em} 1.  Receive the interface information $\mathbf{u}_{|\Gamma_{I}^{in}}^{n, j-1}$  from Model NO and $\mathbf{u}_{|\Gamma_{I}^{out}}^{n,j-1}$ \hspace*{3em}from the external sources.  \\
\hspace*{3em} 2. Solve $\mathbf{u}_{|\Omega_{I}}^{n,j}$ based on the boundary conditions.\\
\hspace*{3em} 3. Obtain the  $\mathbf{u}^{n, j}$  at $\Gamma_{II}^{out}$ and pass it to Model NO.\\

\hspace*{2em}\textbf{Model NO}:\\
\hspace*{3em} 1. Receive the interface information $\mathbf{u}_{|\Gamma_{II}^{out}}^{n, j}$ from Model FE.\\
\hspace*{3em} 2. Solve $\mathbf{u}_{|\Omega_{II}}^{n,j}$ based on the boundary conditions and obtain $\mathbf{u}_{|\Gamma_{II}^{in}}^{n, j}$.\\
\hspace*{3em} 3. Calculate the relaxation formula: $\tilde{\mathbf{u}}_{|\Gamma_{II}^{in}}^{n,j} = (1-\boldsymbol{\rho}_r)\mathbf{u}_{|\Gamma_{II}^{in}}^{n,j} + \boldsymbol{\rho}_r\mathbf{u}_{|\Gamma_{I}^{in}}^{n,j}$;
\hspace*{4em} and pass it back to Model FE.\\
\hspace*{2em} \textbf{If} $\|\mathbf{u}^{n, j}_{|\Omega_{I}} - \mathbf{u}^{n, j-1}_{|\Omega_{I}}\|_{L^2} + \| \mathbf{u}^{n, j}_{|\Omega_{II}} - \mathbf{u}^{n, j-1}_{|\Omega_{II}}\|_{L^2}< \epsilon$, \textbf{end while}\\
\hspace*{1em} \textbf{End for}
\end{algorithm}

In static problems, time stepping is unnecessary and only the inner iteration loop is executed. The convergence criterion is defined by the $L^2$ norm of successive iterates:
\begin{equation} L^2 \text{ error} = \|\mathbf{u}^{n, j}_{|\Omega_{I}} - \mathbf{u}^{n, j-1}_{|\Omega_{I}}\|_{L^2} + \| \mathbf{u}^{n, j}_{|\Omega_{II}} - \mathbf{u}^{n, j-1}_{|\Omega_{II}}\|_{L^2}.
\label{L2 error} \end{equation}
Once convergence is achieved, the solutions are assembled for each subdomain at time step $n$, as:
\begin{align}
\mathbf{u}_{|\Omega_{I}}^n &:= \mathbf{u}_{|\Omega_{I}}^{n, j} \label{u_domain1} \\
\mathbf{u}_{|\Omega_{II}}^n &:= \mathbf{u}_{|\Omega_{II}}^{n, j}. \label{u_domain2}
\end{align}
This results in a global solution $\mathbf{u}_{|\Omega}^n$ over the entire domain. The Schwarz method, long used in classical DDM for PDEs \cite{lions1988schwarz}, proves equally effective in this hybrid FE–NO context. 

\subsection{Expansion of ML-subdomain in dynamic simulations}
\label{puzzle}
A key objective of the FE–NO coupling framework is to accelerate the simulation of complex mechanical systems, particularly under dynamic loading conditions. In many such problems, the regions requiring the most computational effort evolve spatially and temporally throughout the simulation. For example, during crack propagation, the highest stress and computational intensity is localized near the moving crack front. As the crack advances or branches, the surrogate model must adapt accordingly to maintain efficiency and accuracy. In such cases, one needs to either shift the ML-subdomain to the desired location or to adaptively expand its size.

\begin{figure}[H]
\begin{center}
\includegraphics[width=0.8\textwidth]{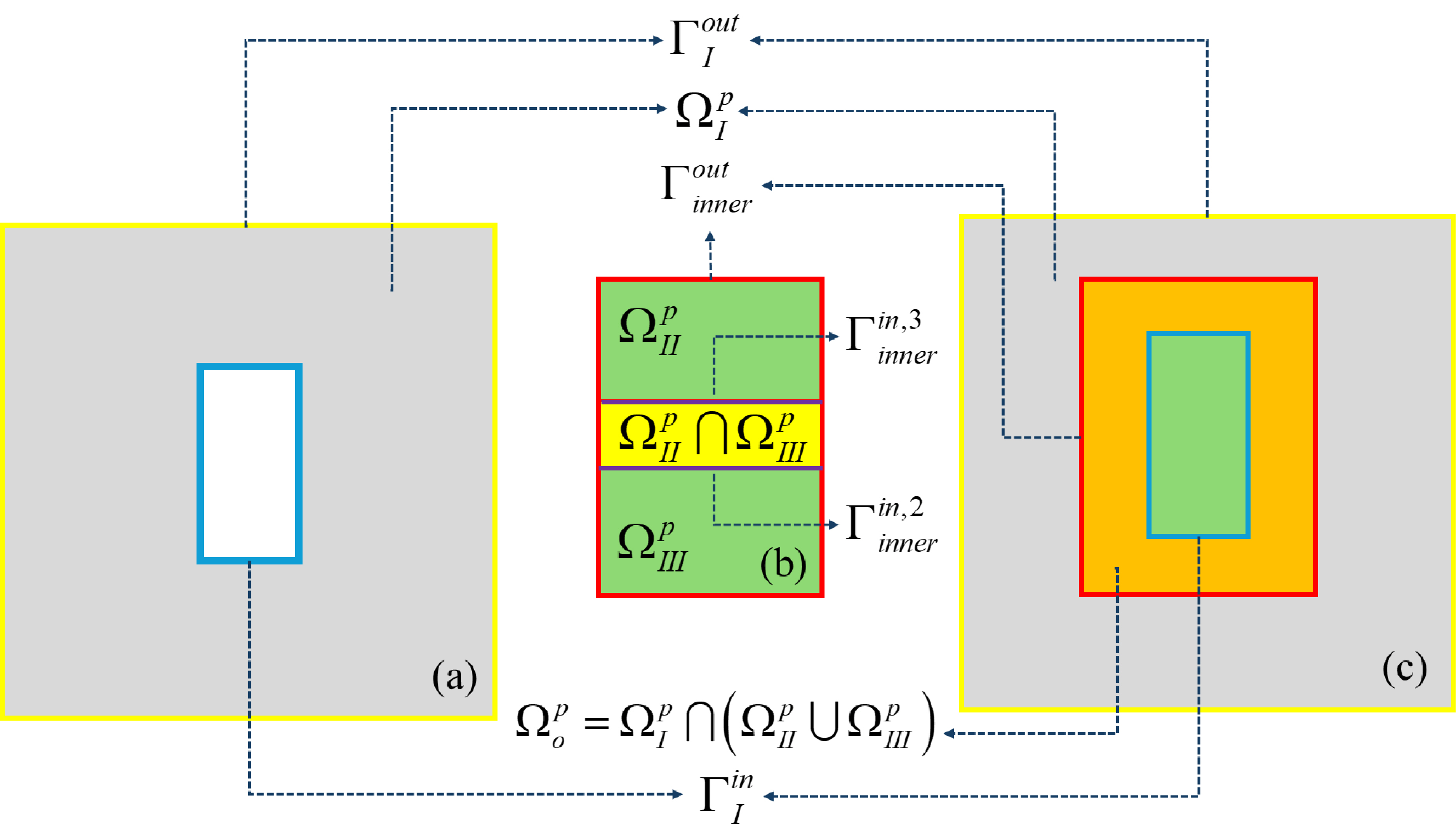}
\caption{Schematic of adaptive DeepONet subdomain expansion:
(a) $\Omega^{p}_{I}$ is the FE domain (gray region) with a central hole representing the initial DeepONet region.
(b) Two PI-DeepONet regions $\Omega^{p}_{II}$ and $\Omega^{p}_{III}$ (green squares) are activated and combined to cover a larger area. The yellow area represents their overlap.
(c) The full domain showing the overlapping region (orange) between the FE domain $\Omega^{p}_{I}$ and the union of the expanded DeepONet domains.}
\label{Fig:Schematics2}
\end{center}
\end{figure}

To address this challenge, we propose a strategy for ML-subdomain expansion, which enables the PI-DeepONet to follow and encapsulate the most demanding computational regions during the simulation. This is achieved by expanding or repositioning the PI-DeepONet subdomain dynamically, replacing the original region with a larger or differently located one as needed. Initially, only a small region-typically centered in the domain-is modeled by the PI-DeepONet, as shown in Fig. \ref{Fig:Schematics}. After a certain number of time steps, this region may be expanded to include more area of interest, such as a propagating crack tip. As illustrated in Fig. \ref{Fig:Schematics2}, two DeepONet subdomains are combined to form a larger surrogate domain. To efficiently train the PI-DeepONet for the expanded subdomain, we apply transfer learning \cite{goswami2020transfer} by fine-tuning only the trunk network of the original PI-DeepONet while keeping the branch network fixed (frozen). These DeepONet models are used in parallel to model the extended region. This dynamic strategy maintains high accuracy while reducing the burden on the FE solver, which continues to handle the remainder of the computational domain.

The underlying algorithm follows a similar structure to the Schwarz alternating method introduced earlier (Algorithm~\ref{algo_1}), with the main difference being the multiple DeepONet regions that must be synchronized. These DeepONet models are treated as a combined surrogate system, exchanging information with each other and with the FE solver. The modified Schwarz algorithm is outlined in Algorithm~\ref{algo_2}, where the inner iteration handles FE–NO coupling and an additional layer handles synchronization between the two NO regions.

\begin{algorithm}[H]
\SetAlgoLined
\caption{Schwarz alternating method for multi-region DeepONet coupling}
\label{algo_2}
\textbf{Initialization}: Set $\mathbf{u}_{NO1}^{0} = \mathbf{0}$ in $\Omega_{II}^{p}$,  $\mathbf{u}_{NO2}^{0} = \mathbf{0}$ in $\Omega_{III}^{p}$, and $\mathbf{u}_{FE}^{0} = \mathbf{0}$ in $\Omega_{I}^{p}$\\
\textbf{Main loop}:\\
\textbf{for} $n = 0 : n_{max} - 1$ \textbf{do}:\\
\hspace*{1em}Set $j = 0$, $k = 0$, $\epsilon$ and $\epsilon'$ the critical values\\
\hspace*{1em}\textbf{while} $j \geq 0$ \textbf{do}:\\
\hspace*{2em} $j = j+1$\\
\hspace*{2em}\textbf{Model FE}:\\
\hspace*{3em} 1.  Receive the interface information $\mathbf{u}_{|\Gamma_{I}^{in}}^{n, j-1}$ from Model combined NOs and  \hspace*{3em} $\mathbf{u}_{|\Gamma_{I}^{out}}^{n,j-1}$ from the external sources.  \\
\hspace*{3em} 2. Solve $\mathbf{u}_{|\Omega^{p}_{I}}^{n,j}$ based on the boundary conditions.\\
\hspace*{3em} 3. Obtain the  $\mathbf{u}^{n, j}$  at $\Gamma_{inner}^{out}$ and pass it to Model combined NOs.\\

\hspace*{2em}\textbf{Model combined NOs}:\\
\hspace*{3em}\textbf{while} $k \geq 0$ \textbf{do}:\\
\hspace*{4em} $k = k+1$\\
\hspace*{4em} \textbf{Model NO1}:\\
\hspace*{5em} \textbf{If} $n = 0, j = 0, k = 0$ \textbf{do}:\\
\hspace*{6em}1. Receive the boundary conditions of Model NO1  from interface 
\hspace*{6em}information $\mathbf{u}_{|\Gamma_{inner}^{out}}^{n, j}$ obtained in Model FE; and assume $\mathbf{u}^{n,j,k}_{|\Gamma_{inner}^{in, 1}} = \mathbf{0}$.\\
\hspace*{5em}\textbf{Else do}:\\
\hspace*{6em}1. Receive the boundary conditions of Model NO1  from interface \\
\hspace*{6em}information $\mathbf{u}_{|\Gamma_{inner}^{out}}^{n, j}$  and  $\mathbf{u}^{n,j,k-1}_{|\Gamma_{inner}^{in, 1}}$ obtained in Model FE and Model \\
\hspace*{6em}NO2, respectively.\\
\hspace*{6em} 2. Solve $\mathbf{u}_{|\Omega^{p}_{II}}^{n,j,k}$ based on the boundary conditions and obtain $\mathbf{u}_{|\Gamma_{inner}^{in,2}}^{n, j,k}$\\
\hspace*{6em} and pass it to Model NO2.\\

\hspace*{4em} \textbf{Model NO2}:\\
\hspace*{5em} 1. Receive the interface information $\mathbf{u}_{|\Gamma_{innner}^{out}}^{n, j}$  from Model FE and  \\
\hspace*{5em}$\mathbf{u}_{|\Gamma_{inner}^{in,2}}^{n, j,k}$ from Model NO1 \\

\hspace*{5em} 2. Solve $\mathbf{u}_{|\Omega^{p}_{III}}^{n,j,k+1}$ based on the boundary conditions, then obtain $\mathbf{u}_{|\Gamma_{inner}^{in,1}}^{n, j,k+1}$,\\
\hspace*{5em} and pass it to Model NO2.\\
\hspace*{3em} \textbf{If} $\|\mathbf{u}^{n, j, k}_{|\Omega_{II}} - \mathbf{u}^{n, j, k-1}_{|\Omega_{II}}\|_{L^2} + \| \mathbf{u}^{n, j, k}_{|\Omega_{III}} - \mathbf{u}^{n, j, k-1}_{|\Omega^{p}_{III}}\|_{L^2}< \epsilon'$,  \textbf{end while}, obtain the 
\hspace*{3em}boundary conditions $\mathbf{u}^{n, j}_{|\Gamma^{in}_{I}} $ and pass it to Model FE.\\
\hspace*{2em} \textbf{If} $\|\mathbf{u}^{n, j, k}_{|\Omega_{I}} - \mathbf{u}^{n, j-1, k}_{|\Omega_{I}}\|_{L^2} + \| \mathbf{u}^{n, j, k}_{|\Omega_{II} \cup \Omega_{III}} - \mathbf{u}^{n, j-1, k}_{|\Omega_{II} \cup \Omega_{III}} \|_{L^2}< \epsilon$, \textbf{end while}\\
\hspace*{1em} \textbf{End for}

\end{algorithm}
To ensure consistency between the FE and ML subdomains and among the ML subdomains themselves, we define two $L^2$-based convergence metrics: $L^2_{p_1}$ measures convergence between the DeepONet subdomains: \begin{equation} L^2_{p_1} = \|\mathbf{u}^{n,j,k}_{|\Omega_{II}} - \mathbf{u}^{n,j,k-1}_{|\Omega_{II}}\|_{L^2} + \|\mathbf{u}^{n,j,k}_{|\Omega_{III}} - \mathbf{u}^{n,j,k-1}_{|\Omega_{III}}\|_{L^2} \label{L2_p1 error} \end{equation}
$L^2_{p_2}$ evaluates convergence between the combined DeepONet region and the FE domain: \begin{equation} L^2_{p_2} = \|\mathbf{u}^{n,j,k}_{|\Omega_{I}} - \mathbf{u}^{n,j-1,k}_{|\Omega_{I}}\|_{L^2} + \|\mathbf{u}^{n,j,k}_{|\Omega_{II} \cup \Omega_{III}} - \mathbf{u}^{n,j-1,k}_{|\Omega_{II} \cup \Omega_{III}}\|_{L^2} \label{L2_p2_error} \end{equation}

Convergence is achieved when both $L^2_{p_1} < \epsilon'$ and $L^2_{p_2} < \epsilon$. The final solution at time step $n$ across all three subdomains is then defined as:
\begin{align} 
\mathbf{u}_{|\Omega_{I}}^n &:= \mathbf{u}_{|\Omega_{I}}^{n,j,k} \\ \mathbf{u}_{|\Omega_{II}}^n &:= \mathbf{u}_{|\Omega_{II}}^{n,j,k} \\ \mathbf{u}_{|\Omega_{III}}^n &:= \mathbf{u}_{|\Omega_{III}}^{n,j,k} \end{align}

The global solution $\mathbf{u}_{|\Omega}^n$ is assembled by stitching together these subdomain solutions. In this work, we manually expanded the ML subdomain to demonstrate the feasibility of adaptive surrogate integration. In future work, we aim to develop a fully automated scheme that dynamically expands the ML subdomain based on parametric thresholds derived from physical or error-based indicators. Furthermore, we demonstrate the method using two DeepONet subdomains; however, the strategy naturally extends to more regions. By incrementally activating and repositioning DeepONet subdomains during simulation, the surrogate model can continuously track and adapt to evolving regions of interest. Additionally, due to the interchangeable roles of the FE and DeepONet solvers, the PI-DeepONet region can be translated across the domain as needed, allowing dynamic redistribution of computational resources. This adaptive strategy offers a powerful way to accelerate large-scale simulations without sacrificing accuracy-enabling simultaneous model refinement and physical prediction throughout the simulation.
\begin{table}[H]
\centering
\caption{PI-DeepONet architectures for different material models.}
\label{tab:NO_architecture}
\begin{tabular}{c c c c}
\Xhline{3\arrayrulewidth}
\textbf{Model} & \textbf{Branch Net} & \textbf{Trunk Net} & \textbf{Activation} \\
\Xhline{3\arrayrulewidth}

Linear Elastic 
& [200$\times$2, 100$\times$4, 800] 
& [2, 100$\times$4, 800] 
& \texttt{tanh} \\ \hline

Hyper-elastic 
& [200$\times$2, 100$\times$4, 800] 
& [2, 100$\times$4, 800] 
& \texttt{tanh} \\ \hline

Elasto-dynamic 
& \begin{tabular}[c]{@{}c@{}}CNN + [82$^2$$\times$4, 256, 800]\\ and [82$\times$4$\times$2, 100$\times$4, 800]\end{tabular}
& [2, 100$\times$4, 800] 
& \texttt{tanh} \\

\Xhline{3\arrayrulewidth}
\end{tabular}
\end{table}

\section{Numerical results and discussion}

In this section, we present five progressively complex case studies to demonstrate the robustness and versatility of the proposed FE--PI-DeepONet coupling framework. Each case highlights different material models, loading regimes and mesh distribution, with increasing complexity in physics, coupling strategy, and meshed geometry. \vspace{-6pt}
\begin{itemize}
    \item \textbf{Case 1} (Section~\ref{section_static}): A linear elastic model under static loading, where displacement responses are mapped from prescribed boundary conditions.\vspace{-6pt}
    \item \textbf{Case 2} (Section~\ref{section_quasi_static}): A hyperelastic model subjected to quasi-static loading, capturing large deformations in nonlinear materials.\vspace{-6pt}
    \item \textbf{Case 3} (Section~\ref{section_elasto_dynamic}): A linear elastic model under dynamic loading, incorporating time-dependent behavior and field evolution.\vspace{-6pt}
    \item \textbf{Case 4} (Section~\ref{section_puzzle}): A dynamic case that also demonstrates adaptive expansion of the ML-resolved subdomain within the simulation.\vspace{-6pt}
    \item \textbf{Case 5} (Section~\ref{thick_cylinder}): A static case with non-uniform mesh distribution over complex geometry, demonstrating the potential of our coupling framework for multi-scale problems that require adaptive mesh refinement.\vspace{-6pt}
\end{itemize}
In all five examples, the PI-DeepONet is coupled with the FE solver across a designated spatial subdomain. Each architecture has a single trunk network, which encodes the spatial coordinates $(x, y)$. For each case, two identical PI-DeepONet are trained independently to predict the \review{$u_x$}- and \review{$u_y$}-displacements, enabling efficient and accurate inference during coupled simulations. The Adam optimizer~\cite{kingma2014adam} is used for training, with a learning rate of $10^{-3}$ for the linear elastic models and $10^{-4}$ for the hyperelastic model. Once trained, the resulting network parameters $\boldsymbol{\uptheta}^*$ define a displacement-predicting surrogate that integrates seamlessly within the FE--NO framework. The branch network design varies between static/quasi-static and dynamic cases:\vspace{-6pt}
\begin{itemize}
    \item For the \textit{static and quasi-static} cases, the branch network encodes Dirichlet boundary conditions from $\Gamma_{II}^{out}$ using 200 boundary points and 2 displacement components per point, yielding a total of 400 input features.\vspace{-6pt}
    \item For the \textit{dynamic} cases, the square domain allows for richer input features. The branch network incorporates both field variables (e.g., displacement, velocity) and boundary data. A CNN or a large fully connected network is used to encode \review{$u_x$}, \review{$u_y$}, $v_x$, and $v_y$ on an $82 \times 82$ grid, boundary data are provided separately as input.\vspace{-6pt}
\end{itemize}
Time integration for the dynamic cases (Sections \ref{section_elasto_dynamic} and \ref{section_puzzle}) is performed using the Newmark method. \review{To simplify the analysis and reduce dimensionality, we employ a non-dimensionalized formulation with fixed material properties: density $\rho = 5$, time increment $\Delta t = 1$, Young’s modulus $E = 1000$, and Poisson’s ratio $\nu = 0.3$. All governing equations used in this study are presented in non-dimensionalized form, with corresponding reference parameters, in \ref{APP_II}.} All simulations in Sections \ref{section_static} - \ref{section_puzzle} are performed on a square domain of length 2 units. To demonstrate the prediction accuracy of the training of DeepONet, we have provided additional plots in Appendix A. 

\review{Furthermore, our choice of employing neural networks based surrogates in stress-concentrated regions (although NN typically struggle in such regimes) was informed by two factors: \vspace{-6pt}
\begin{itemize}
    \item Boundary data propagation: In our setup, if the NN is used in the outer/coarse region, its predicted displacements serve as boundary input to the FEM domain. Any approximation error from the NN then propagates directly into the high-fidelity FEM region, degrading accuracy where it matters most. \vspace{-6pt}
    \item Computational cost: FEM simulations are more expensive in high-resolution, localized domains (e.g., around stress concentrations). Assigning these regions to the NN surrogate helps reduce computational cost, even if some accuracy is sacrificed. \vspace{-6pt}
\end{itemize}}

\subsection{Hybrid framework for static loading}
\label{section_static}

\begin{figure}[H]
\begin{center}
\includegraphics[width=1\textwidth, height=12cm]{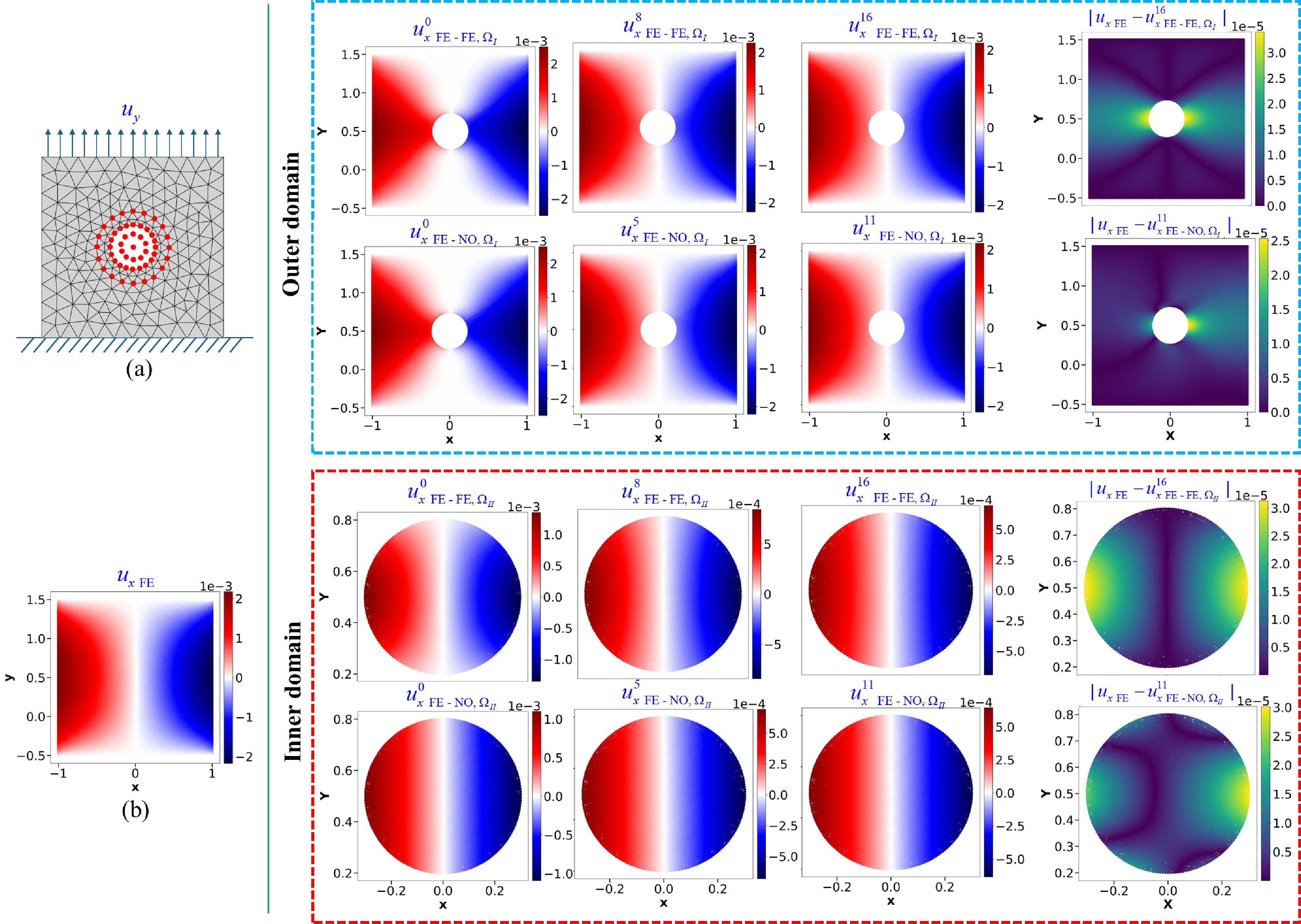}
\caption{Response in $x$-direction (\review{$u_x$}) of the linear elastic coupling model under static loading conditions: (a) Schematic of decomposed domains for the spatial coupling framework, where the bottom edge has fixed boundary conditions and the top edge is subjected to an applied displacement in the $y$-direction (\review{$u_y = 0.01$}); (b) Ground truth displacement \review{$u_x$} obtained by solving the intact domain using FEniCSx. The blue-dashed box contains: Columns 1-3 showing the evolution of \review{$u_x$} in $\Omega_{I}$ for FE-FE coupling (top row) at iterations $j = 0, 8, 16$ and FE-NO coupling (bottom row) at iterations $j = 0, 5, 11$, with column 4 displaying the absolute error between the converged solution ($j = 16$ for FE-FE and  $j = 11$ for FE-NO) and the ground truth. The red-dashed box contains: Columns 1-3 showing the evolution of \review{$u_x$} in $\Omega_{II}$ for both coupling frameworks at the same iterations, with column 4 similarly displaying the absolute error relative to the ground truth.}
\label{Fig:static_u}
\end{center}
\end{figure}

\begin{figure}[H]
\begin{center}
\includegraphics[width=1\textwidth, height= 12cm]{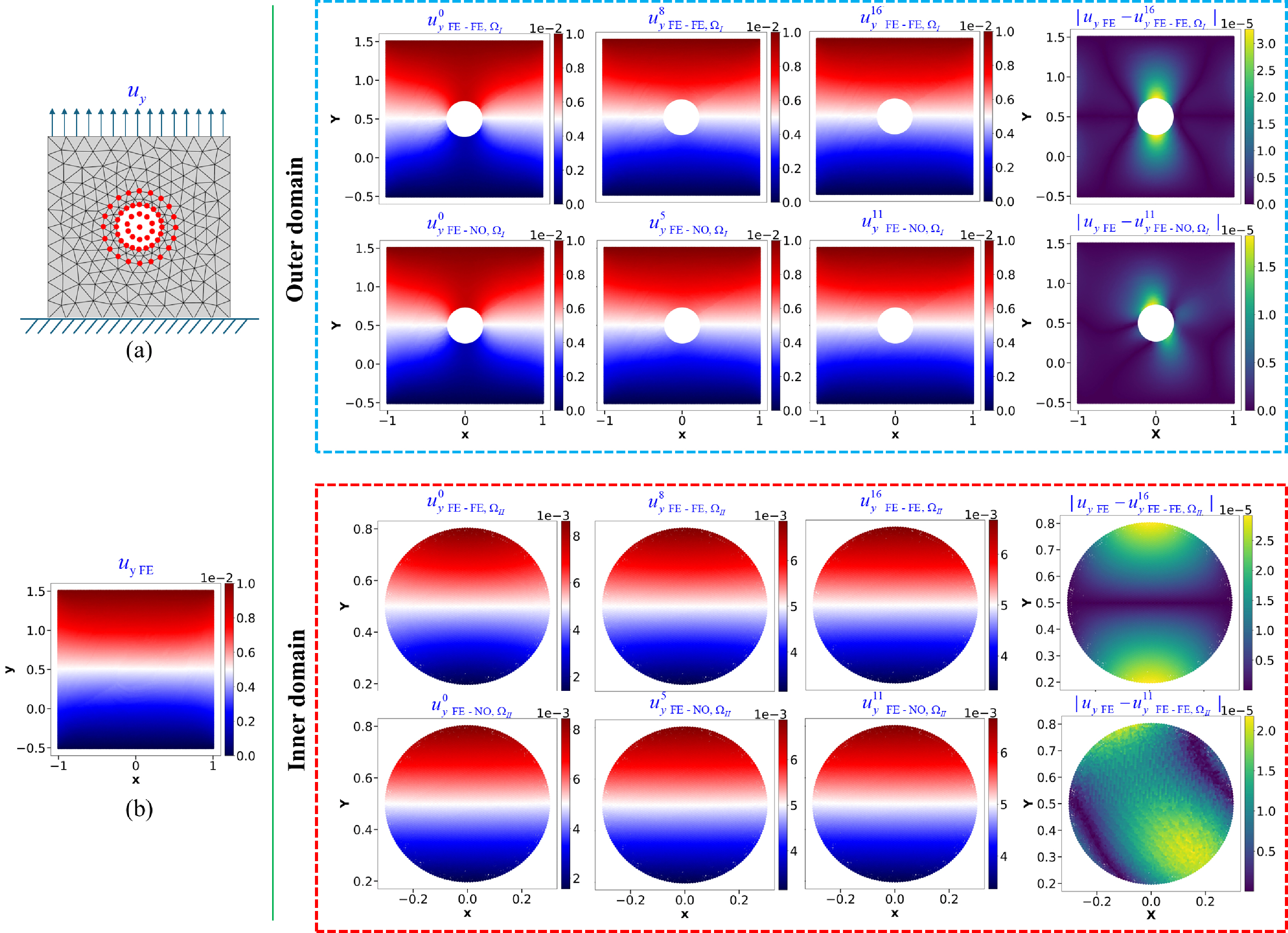}
\caption{Response in $y$-direction (\review{$u_y$}) of the linear elastic coupling model under static loading conditions: (a) Schematic of decomposed domains for the spatial coupling framework, where the bottom edge has fixed boundary conditions and the top edge is subjected to an applied displacement in the $y$-direction (\review{$u_y = 0.01$}); (b) Ground truth displacement \review{$u_y$} obtained by solving the intact domain using FEniCSx. The blue-dashed box contains: Columns 1-3 showing the evolution of \review{$u_y$} in $\Omega_{I}$ for FE-FE coupling (top row) at iterations $j = 0, 8, 16$ and FE-NO coupling (bottom row) at iterations $j = 0, 5, 11$, with column 4 displaying the absolute error between the converged solution ($j = 16$ for FE-FE and $j = 11$ for FE-NO) and the ground truth. The red-dashed box contains: Columns 1-3 showing the evolution of \review{$u_y$} in $\Omega_{II}$ for both coupling frameworks at the same iterations, with column 4 similarly displaying the absolute error relative to the ground truth.}
\label{Fig:static_v}
\end{center}
\end{figure}

In the first example, we consider a linear elastic problem defined on a square domain subject to displacement-controlled loading applied to the top face of the domain.  \review{The schematic representation of the decomposed domains for the FE-NO coupling is shown in Fig. \ref{Fig:static_u}(a), where the inner domain $\Omega_{II}$ is a disk with radius 0.35 and the outer domain $\Omega_{I}$ is a square with a circular hole of radius 0.3}. The displacement along the bottom face of the domain is fixed, while a constant displacement \review{$u_y = 0.01$} in the $y$-direction is uniformly imposed on the top face. Figures \ref{Fig:static_u}(b) and \ref{Fig:static_v}(b) present the results for displacement in $x$ direction and $y$ direction respectively, obtained by FE simulation on the entire domain. These results serve as the ground truth against which the accuracy of the DDM based approaches is evaluated. Our first task in implementing the proposed spatial coupling framework is to train the PI-DeepONet. 
For this purpose, we generate 1000 boundary conditions as a Gaussian random field with length scale parameters $l_u = l_v = 2$ for displacements \review{$u_x$} and \review{$u_y$}, respectively. By substituting these displacements into the PI-DeepONet operators \review{$G^{u_x}_{\boldsymbol{\uptheta}_1}$} and \review{$G^{u_y}_{\boldsymbol{\uptheta}_2}$} in Eq. \ref{strain}, \ref{stress}, and \ref{static_equa}, and considering the case with no body force ($\mathbf{f} = \mathbf{0}$), the physics-informed neural operator is trained using the following governing equations:

\begin{align}
&(\lambda + 2\mu)\frac{\partial^2 G^{u_x}_{\boldsymbol{\uptheta}_1}}{\partial x^2} + \mu \frac{\partial^2 G^{u_x}_{\boldsymbol{\uptheta}_1}}{\partial y^2} + (\lambda + \mu)\frac{\partial^2 G^{u_y}_{\boldsymbol{\uptheta}_2}}{\partial x \partial y} = 0 , (x,y)\in \Omega_{II} \label{static_PI_NO_res0}\\
&(\lambda + 2\mu)\frac{\partial^2 G^{u_y}_{\boldsymbol{\uptheta}_2}}{\partial y^2} + \mu \frac{\partial^2 G^{u_y}_{\boldsymbol{\uptheta}_2}}{\partial x^2} + (\lambda + \mu)\frac{\partial^2 G^{u_x}_{\boldsymbol{\uptheta}_1}}{\partial x \partial y} = 0, (x,y)\in \Omega_{II} \label{static_PI_NO_res1}\\
&G^{u_x}_{\boldsymbol{\uptheta}_1} = u_x(x,y), (x,y) \in \Gamma^{out}_{II} \label{static_PI_NO_bcs0}\\
&G^{u_y}_{\boldsymbol{\uptheta}_2}= u_y(x,y), (x,y) \in \Gamma^{out}_{II} \label{static_PI_NO_bcs1}
\end{align}
After training this PI-DeepONet for $2\times 10^{6}$ iterations, the loss plot is obtained as shown in Fig. \ref{Fig:static_DeepONet}(a) with the boundary conditions loss ($\mathcal{L}_{bcs}$)and the residual loss ($\mathcal{L}_{res}$), which are based on the Eq.\ref{static_PI_NO_bcs0}-\ref{static_PI_NO_bcs1} and  Eq.\ref{static_PI_NO_res0}-\ref{static_PI_NO_res1}, respectively.

 To evaluate the performance of the trained DeepONet prior to the coupling process, we test the trained DeepONet with one unseen boundary conditions of $\Omega_{II}$. For example, we apply boundary conditions extracted from the ground truth in Fig. \ref{Fig:static_u}(b) and \ref{Fig:static_v}(b).  \review{Comparing the prediction results with ground truth (FE), absolute errors for $u_x$ and $u_y$ are both below $1.75\times10^{-5}$ (Appendix Fig.~\ref{Fig:static_DeepONet}}. This indicates that our ML model is trained well for accurate generalization.

The domain $\Omega_I$ shown in Fig. \ref{Fig:static_u}(a) is modeled in FEniCSx with uniform displacement on the top face, fixed bottom face, and responding displacement from $\Omega_{II}$ at the interface $\Gamma_I^{in}$. The domain $\Omega_I$ is discretized into 38,812 elements, with 200 nodes uniformly distributed along the interfaces $\Gamma_{II}^{out}$ and $\Gamma_{II}^{in}$ to facilitate information transfer.

We first present our results for the FE-NO coupling before comparing its convergence with the FE-FE coupling approach. The coupling process begins by solving for $\Omega_I$ using the FEniCSx solver and transferring the displacement information (\review{$u_x$} and \review{$u_y$}) at $\Gamma_{II}^{out}$ to the pre-trained PI-DeepONet. The neural operator then predicts the displacements throughout $\Omega_{II}$, providing values for \review{$u_{x|\Gamma_{II}^{in}}$} and \review{$u_{y|\Gamma_{I}^{in}}$} that are passed back to $\Omega_I$. This exchange of information continues until convergence between the solutions of the two subdomains is achieved, as calculated using Eq. \ref{L2 error}. 

For the FE-FE coupling, we follow the same process, except that subdomain $\Omega_{II}$ is simulated using the FE solver instead of the PI-DeepONet. In this case, $\Omega_{II}$ is discretized into 21,490 elements with 200 nodes on $\Gamma_{II}^{in}$ and $\Gamma_{II}^{out}$. The nodes are positioned at the same coordinates as those in $\Omega_{I}$ to ensure efficient boundary condition exchange. \review{In Figures \ref{Fig:static_u} and \ref{Fig:static_v}, we use the notations ${u_x^{j}}_{FE-FE}$ and ${u_y^{j}}_{FE-FE}$ to denote the $x$ and $y$ displacements obtained using the FE-FE coupling at the $j$-th iteration of the Schwarz coupling strategy.  For the FE-NO coupling, we use the same notation with the subscript modified to FE-NO. It should be emphasized that the superscript $n$ in \autoref{section_quasi_static} - \autoref{section_puzzle} indicates the time step. }

Figure \ref{Fig:static_u} presents the $x$-displacement results for FE-FE at inner iterations $j=\{0, 8, 16\}$ and for FE-NO at $j = \{0, 5, 11\}$, along with the computed errors at the final iteration for both $\Omega_I$ and $\Omega_{II}$. At $j=0$, both \review{${u_x^0}_{FE-FE, \Omega_I}$} and \review{${u_x^0}_{FE-NO, \Omega_I}$} exhibit sharp gradients in displacement field near $\Omega_{II}$, which appears as a hole for the FE solver of $\Omega_I$.
% A large portion of the upper and lower regions appears white (close to 0), while dark red and dark blue occupy significant areas in the left and right regions, extending to the edge of $\Omega_{II}$. While this displacement gradient presents an arc-shaped distribution at the left and right portions of $\Omega_{\text{II}}$ in FE-FE coupling, in $u^0_{\text{FE-NO},\Omega_{\text{II}}}$, the displacement distributes uniformly. In the inner domain, the displacement extrema for FE-FE  reach approximately $\pm 1.3 \times 10^{-3}$,  while for FE-NO close to $1 \times 10^{-3}$.
After 9 iterations ($j = 8$), the sharp gradients in the displacement field in \review{${u_x^8}_{FE-FE}$} are significantly reduced across both domains. A similar mitigation is also observed in \review{${u_x^5}_{FE-NO}$}.
% The white area in $\Omega_I$ reduces to a vertical line, and the horizontal edges of $\Omega_{II}$ transition to lighter red and blue, indicating attenuation of the concentration. This phenomenon is particularly evident in $u^8_{FE-FE, \Omega_{II}}$and $u^5_{FE-NO,\Omega_{II}}$ , where a significant reduction in the magnitude of $u_x$ is observed, and the arc-shaped distribution weakens in FE-FE coupling.

After additional iterations, the difference between \review{${u_x^{8}}_{FE-FE,\Omega_{I}}$} and \review{${u_x^{16}}_{FE-FE,\Omega_{I}}$} becomes nearly indistinguishable, similar to that between \review{${u_x^5}_{FE-NO,\Omega_{I}}$} and \review{${u_x^{11}}_{FE-NO, \Omega_{I}}$}.

% The magnitudes of the last two $x$-displacement in $\Omega_{II}$ for FE-FE and FE-NO slightly decrease.

The final column in Fig. \ref{Fig:static_u} shows the absolute error between the FE reference solution $u_{FE}$ and final converged solutions in both domains. For both coupling methods, the largest errors occur near the interface between $\Omega_{I}$ and $\Omega_{II}$. While the FE-NO coupling yields slightly lower errors than the FE-FE coupling, all absolute errors remain within the same order of magnitude $10^{-5}$.

Similarly, Fig. \ref{Fig:static_v} illustrates the evolution of the $y$-displacement for both coupling frameworks. 
% In $\Omega_{I}$, the white region along the horizontal direction narrows to a line, and the $u_y$ distribution becomes more uniform in the vertical direction. For $v_{\Omega_{II}}$, the maximum value decreases as the iterations progress, and the arc-shaped distribution dissipates in FE-FE. 
The absolute error between the final converged solution and ground truth, depicted in the last column of Fig. \ref{Fig:static_v}, is comparable to that observed for \review{$u_x$}, also on the order of $10^{-5}$.

Given the uniaxial loading in the $y$-direction, the \review{$u_y$} displacement is approximately $10$ times greater than \review{$u_x$}, the latter arising primarily from the Poisson effect. Consequently, similar absolute error magnitudes translate into different relative errors for \review{$u_x$} (less than 5\%) and \review{$u_y$} (less than 0.5\%).

Due to symmetry in both loading and geometry, the displacement distribution and the absolute error in the FE-FE coupling exhibit symmetric patterns. The maximum error in \review{$|{u_x}_{FE} - {u_x}^{16}_{FE-FE}|$} appears along the $x$-direction, while for \review{$|{u_y}_{FE} - {u_y}^{16}_{FE-FE}|$}, it occurs along the $y$-direction. In contrast, the FE-NO coupling introduces asymmetric error distributions, leading to slight asymmetries in the final \review{$u_x$} and \review{$u_y$} fields, as shown in Fig. \ref{Fig:static_u} and Fig. \ref{Fig:static_v}. However, due to the small magnitude of the errors, these asymmetries are not visually noticeable.  
% Compared with the absolute error in Fig. \ref{Fig:static_DeepONet}, it is found that in FE-NO coupling the error value is within $1.5 \times 10^{-5}$ higher due to the information transferring.

\begin{figure}[H]
\begin{center}
\includegraphics[width=0.8\textwidth, height= 8cm]{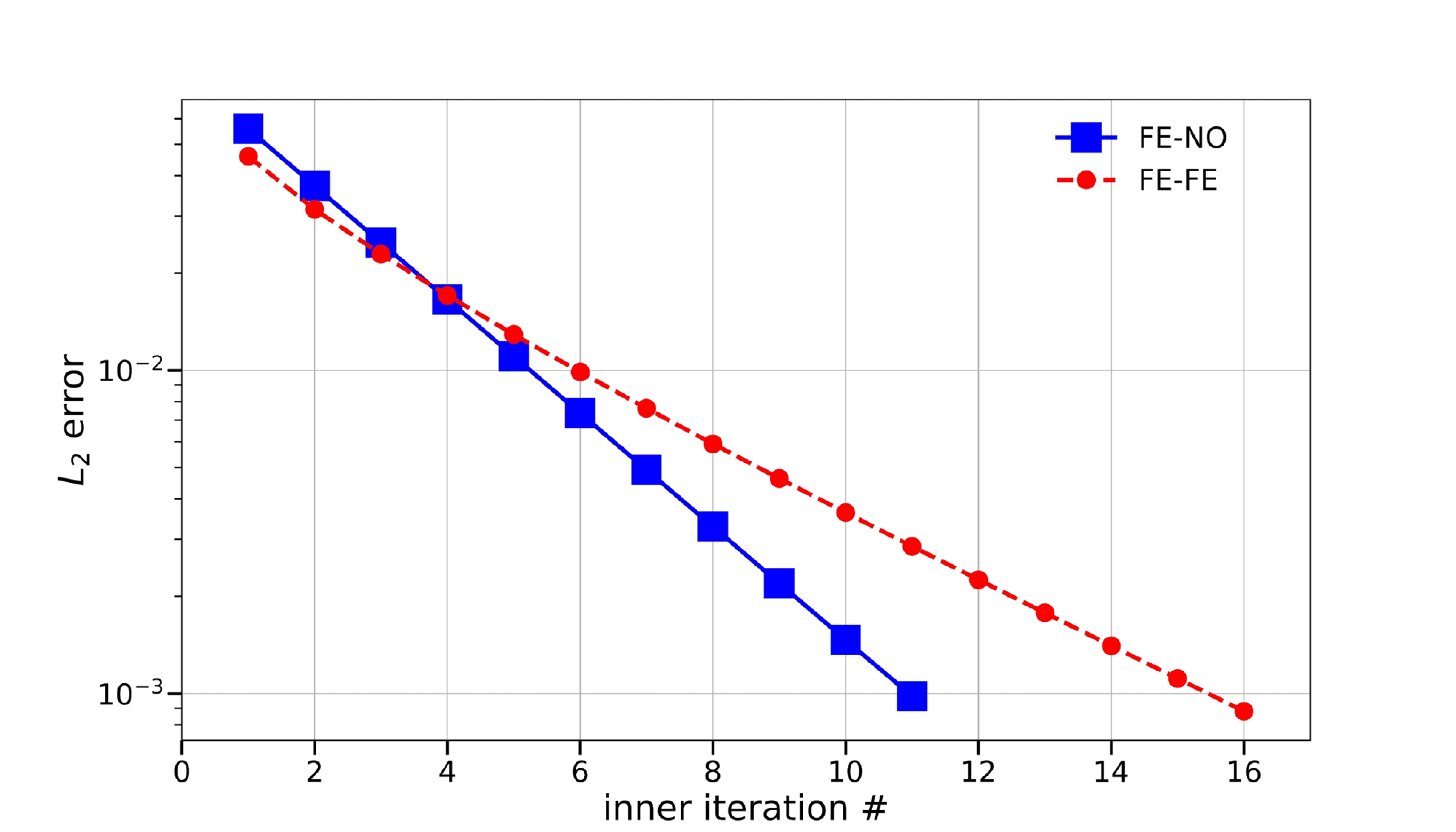}
\caption{$L^2$ error norm evolution across iterations for the linear elastic model under static loading conditions.}
\label{Fig:static_error}
\end{center}
\end{figure}

Following Algorithm \ref{algo_1}, the Schwarz alternating method terminates based on the $L^2$ error defined in Eq. \ref{L2 error}. The error profile for the linear elastic material in the static regime is shown in Fig. \ref{Fig:static_error}. For both coupling approaches, the $L^2$ error decreases with each iteration, indicating a continuous reduction in the displacement difference between successive iterations. Eventually, the $L^2$ error falls below the critical threshold $\epsilon = 10^{-3}$, signifying convergence in both domains as defined by Eq. \ref{u_domain1} and \ref{u_domain2}. With this threshold, absolute errors remain on the order of $10^{-5}$, as demonstrated in Fig. \ref{Fig:static_u} and Fig. \ref{Fig:static_v}. While a smaller $\epsilon$ would require more iterations, it does not significantly improve the absolute error.

Figure~\ref{Fig:static_error} presents the error $L^2$ versus the number of inner iterations for the FE-FE and FE-NO coupling. As shown, the FE-NO solution starts with a higher $L^2$ error in the initial iteration ($j = 1$), which can be attributed to the limitations of the PI-DeepONet model. Specifically, the model was trained using relatively large length parameters ($l_u = l_v = 2$), while the radius of $\Omega_{II}$ is only 0.35. This mismatch hinders the model’s ability to accurately capture the complex boundary conditions in this region.

This behavior is evident in Figs.~\ref{Fig:static_u} and \ref{Fig:static_v} at $j = 0$: due to the sharp displacement gradient in $\Omega_{I}$, the FE-FE coupling yields an arc-shaped displacement distribution, while the FE-NO solution remains nearly uniform.

Since the inner iteration process in linear elasticity serves to smooth sharp displacement gradients, the FE-NO coupling converges faster than the FE-FE approach, requiring five fewer inner iterations to meet the convergence criterion. As a result, the total computation time is reduced from 102 seconds for FE-FE to 80 seconds for FE-NO-yielding a 20\% improvement in efficiency for this simple, small-scale linear elastic example.

It should be emphasized that at each iteration of the FE-FE coupling, the inner domain $\Omega_{II}$ must be completely recalculated. In contrast, for the FE-NO coupling, \review{${u_x}_{|\Omega_{II}}$} is obtained through direct evaluation of the neural operator at spatial coordinates ($x$, $y$), without requiring additional calculations or training. It is anticipated that highly nonlinear materials and complex geometries would pose greater challenges to the FE solver, making FE-FE significantly more time-consuming than FE-NO in such scenarios.

\subsection{Hybrid framework for quasi-static loading}
\label{section_quasi_static}
\begin{figure}[H]
\begin{center}
\includegraphics[width=1\textwidth, height= 12cm]{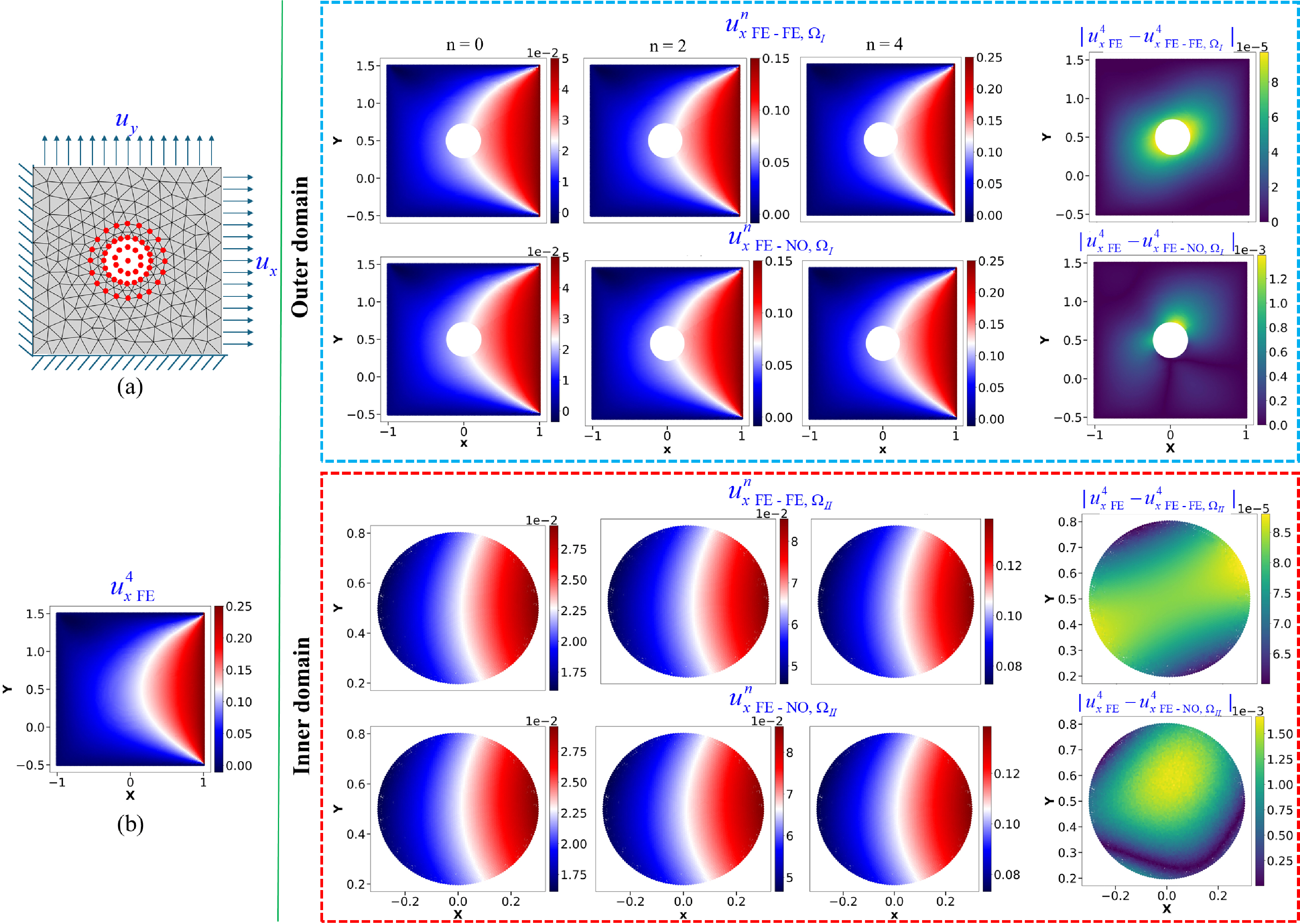}
\caption{Response in $x$-direction (\review{$u_x$}) of the hyperelastic coupling model under quasi-static loading conditions: (a) Schematics of decomposed domains for spatial coupling framework, where the left and bottom edges have fixed boundary conditions and the top and right edges are subjected to an applied displacement at time step $n$ in $y$-direction (\review{$u_y = 0.05(n+1) $}) and $x$-direction (\review{$u_x = 0.05(n+1)$}); (b) Ground truth displacement \review{$u_x$} at time step $n=4$ obtained by solving the intact domain using FEniCSx;  The blue-dashed box contains: Columns 1-3 showing the evolution of  \review{$u_x$} in $\Omega_{I}$ for FE-FE coupling (top row) and FE-NO coupling (bottom row) at time step $n = 0, 2 , 4$, with column 4 displaying the absolute error between the converged solution at time step $n = 4$ and ground truth. The red-dashed box contains: Columns 1-3 showing the evolution of \review{$u_x$} in $\Omega_{II}$ for both coupling frameworks at the same time steps, with column 4 displaying the absolute error relative to the ground truth.}
\label{Fig:quasi_static_hyper_u}
\end{center}
\end{figure}

\begin{figure}[H]
\begin{center}
\includegraphics[width=1\textwidth, height= 11cm]{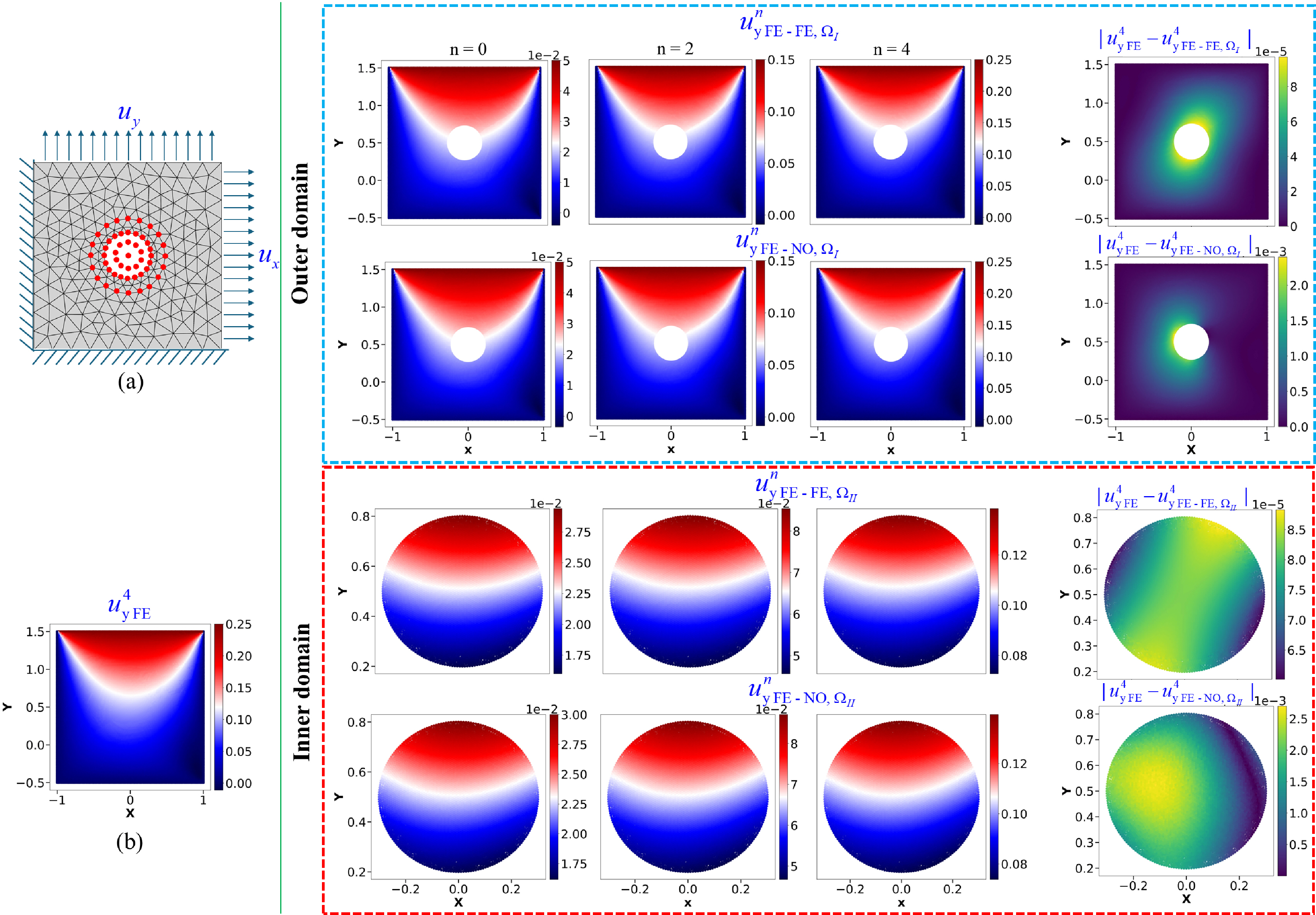}
\caption{Response in $y$-direction (\review{$u_y$}) of the hyperelastic coupling model under quasi-static loading conditions: (a) Schematics of decomposed domains for spatial coupling framework, where the left and bottom edges have fixed boundary conditions and the top and right edges are subjected to an applied displacement at time step $n$ in $y$-direction (\review{$u_y = 0.05(n+1) $}) and $x$-direction (\review{$u_x = 0.05(n+1)$}); (b) Ground truth displacement \review{$u_y$} at time step $n=4$ obtained by solving the intact domain using FEniCSx; the blue-dashed box contains: Columns 1-3 showing the evolution of  \review{$u_y$} in $\Omega_{I}$ for FE-FE coupling (top row) and FE-NO coupling (bottom row) at time step $n = 0, 2, 4$, with column 4 displaying the absolute error between the converged solution at time step $n = 4$ and ground truth. The red-dashed box contains: Columns 1-3 showing the evolution of \review{$u_y$} in $\Omega_{II}$ for both coupling frameworks at the same time steps, with column 4 displaying the absolute error relative to the ground truth.}
\label{Fig:quasi_static_hyper_v}
\end{center}
\end{figure}

In this subsection, a hyperelastic material in a square domain is examined under quasi-static loading. The schematic of the decomposed domains for the FE-NO coupling is similar to that of the previous Section \ref{section_static}, as displayed in Fig. \ref{Fig:quasi_static_hyper_u}(a). The discretization in FE analysis is also the same as before; however, the boundary conditions are different. The left and bottom edges are fixed, and the monotonically increasing displacement is uniformly applied on the top (\review{$u_y = 0.05 \times (n+1) $}) and right (\review{$u_x = 0.05 \times (n +1)$}) edges at the time step $n$ (with $n$ denoting the quasi-static time step). The ground truth values are computed using FE over the entire domain. The displacement components \review{$u_x$} and \review{$u_y$} in $n = 4$ are shown in Fig.\ref{Fig:quasi_static_hyper_u}(b) and Fig.\ref{Fig:quasi_static_hyper_v}(b), respectively. It is worth noting that in this example, the FE solvers employ Newton's nonlinear solver in FEniCSx \cite{petsc-web-page}, as each loading step involves solving a nonlinear equation due to the Neo-Hookean material model.

To enable FE-NO coupling, a PI-DeepONet is trained for the hyperelastic system, following the same approach as for the linear elastic case in the static setting. A total of 1000 boundary conditions are sampled from a Gaussian random field with length scale parameters $l_u = l_v = 1$. These are used as input to the DeepONet operators \review{$G^{u_x}_{\boldsymbol{\uptheta}_1}$} and \review{$G^{u_y}_{\boldsymbol{\uptheta}_2}$} (see Eq.~\ref{F_grad}–\ref{first_piola_kirchhoff}). Substituting these predicted displacements into the constitutive equations yields the first Piola-Kirchhoff stress tensor $\boldsymbol{P_1}^n$ at time step $n$ under quasi-static conditions. The equilibrium condition in the reference configuration (neglecting body forces) is: 
\begin{equation} 
\nabla \cdot \boldsymbol{P_1}^n = 0 \label{static_equa_P1} \end{equation}

Substituting the PI-DeepONet predictions into this equation, the governing system for the hyper-elastic model at time step $n$ becomes:
\begin{align}
 &\frac{\partial}{\partial x} \left[ \mu \left(1 + \frac{\partial G^{u_x}_{\boldsymbol{\uptheta}_1}}{\partial x}\right) + \left(\lambda \log(J(G^{u_x}_{\boldsymbol{\uptheta}_1}, G^{u_y}_{\boldsymbol{\uptheta}_2})) - \mu\right) \frac{1}{J(G^{u_x}_{\boldsymbol{\uptheta}_1}, G^{u_y}_{\boldsymbol{\uptheta}_2})} \left(1 + \frac{\partial G^{u_y}_{\boldsymbol{\uptheta}_2}}{\partial y}\right) \right] \nonumber\\
&+ \frac{\partial}{\partial y} \left[ \mu \frac{\partial G^{u_y}_{\boldsymbol{\uptheta}_2}}{\partial x} + \left(\lambda \log(J(G^{u_x}_{\boldsymbol{\uptheta}_1}, G^{u_y}_{\boldsymbol{\uptheta}_2})) - \mu\right) \frac{1}{J(G^{u_x}_{\boldsymbol{\uptheta}_1}, G^{u_y}_{\boldsymbol{\uptheta}_2})} \left(-\frac{\partial G^{u_x}_{\boldsymbol{\uptheta}_1}}{\partial y}\right) \right] = 0, (x,y) \in \Omega_{II}\label{hyper_equi_0}, \\\nonumber \\
& \frac{\partial}{\partial y} \left[ \mu \left(1 + \frac{\partial G^{u_y}_{\boldsymbol{\uptheta}_2}}{\partial y}\right) + \left(\lambda \log(J(G^{u_x}_{\boldsymbol{\uptheta}_1}, G^{u_y}_{\boldsymbol{\uptheta}_2})) - \mu\right) \frac{1}{J(G^{u_x}_{\boldsymbol{\uptheta}_1}, G^{u_y}_{\boldsymbol{\uptheta}_2})} \left(1 + \frac{\partial G^{u_x}_{\boldsymbol{\uptheta}_1}}{\partial x}\right) \right] \nonumber\\
&+\frac{\partial}{\partial x} \left[ \mu \frac{\partial G^{u_x}_{\boldsymbol{\uptheta}_1}}{\partial y} + \left(\lambda \log(J(G^{u_x}_{\boldsymbol{\uptheta}_1}, G^{u_y}_{\boldsymbol{\uptheta}_2})) - \mu\right) \frac{1}{J(G^{u_x}_{\boldsymbol{\uptheta}_1}, G^{u_y}_{\boldsymbol{\uptheta}_2})} \left(-\frac{\partial G^{u_y}_{\boldsymbol{\uptheta}_2}}{\partial x}\right) \right]= 0, (x,y) \in \Omega_{II}\label{hyper_equi_1}, \\\nonumber 
\end{align}
where the Jacobian determinant $J$ represents the local volume change: \begin{equation} J = \left( 1+ \frac{\partial G^{u_x}_{\boldsymbol{\uptheta}_1}}{\partial x} \right)\left( 1 + \frac{\partial G^{u_y}_{\boldsymbol{\uptheta}_2}}{\partial y} \right) - \frac{\partial G^{u_x}_{\boldsymbol{\uptheta}_1}}{\partial y} \frac{\partial G^{u_y}_{\boldsymbol{\uptheta}_2}}{\partial x}. \end{equation}
The boundary conditions were imposed as: 
\begin{align}
G^{u_x}_{\boldsymbol{\uptheta}_1} &= u_x^n(x,y), (x,y) \in \Gamma^{out}_{II} \label{HY_PI_NO_bcs0}\\
G^{u_y}_{\boldsymbol{\uptheta}_2} &= u_y^n(x,y), (x,y) \in \Gamma^{out}_{II} \label{HY_PI_NO_bcs1}
\end{align}
The PI-DeepONet described by Eq.\ref{hyper_equi_0}–\ref{HY_PI_NO_bcs1} is trained over 2 million iterations to optimize the parameters $\boldsymbol{\uptheta}^* = \{\boldsymbol{\uptheta}_1, \boldsymbol{\uptheta}_2\}$. Figure~\ref{Fig:hyper_DeepONet}(a) shows the loss reduction over iterations, accounting for both the residual and the boundary components. To assess generalization, the trained PI-DeepONet is tested on an unseen boundary condition at time step $n = 4$, corresponding to uniform displacement loading. The ground truth FE solutions \review{${u_x^4}_{FE,\Omega_{II}}$} and \review{${u_y^4}_{FE,\Omega_{II}}$} are compared with PI-DeepONet predictions \review{${u_x^4}_{NO,\Omega_{II}}$} and \review{${u_y^4}_{NO,\Omega_{II}}$}. \review{The absolute errors, $|{u_x^4}_{FE,\Omega_{II}} - {u_x^4}_{NO,\Omega_{II}}|$ and $|{u_y^4}_{FE,\Omega_{II}} - {u_y^4}_{NO,\Omega_{II}}|$, are both below $1.2 \times 10^{-3}$, as presented in Fig.~\ref{Fig:hyper_DeepONet}}. Given the maximum displacement magnitude of 0.139, the relative error is below 1\%, which is acceptable for engineering applications. These results confirm that the PI-DeepONet accurately generalizes to unseen inputs and is well suited for use in FE-NO coupling frameworks.

The FE-NO coupling framework under quasi-static conditions follows the same setup as described in Section \ref{section_static}, with convergence determined by the $L^2$ error criterion in Eq. \ref{L2 error}, using $\epsilon = 10^{-3}$. A key difference here is that for each time step, the inner iterations must converge before advancing to the next step. This time-marching scheme and the information exchange procedure are outlined in Algorithm \ref{algo_1}. For comparison, the FE-FE coupling uses the same algorithm, with $\Omega_{II}$ solved by FE rather than a neural operator. We denote the converged displacements in time step $n$ as\review{ ${u_x^n}_{FE-FE}$} and \review{${u_y^n}_{FE-FE}$} for the FE-FE case, and \review{${u_x^n}_{FE-NO}$} and \review{${u_y^{n}}_{FE-NO}$} for the FE-NO case.

The last columns of Fig. \ref{Fig:quasi_static_hyper_u} display the absolute error of the coupling frameworks against the reference solution of FE at n = 4 in both domains. This FE-FE coupling presents a comparable absolute error to that in the linear elastic example, at the same order of magnitude $10^{-5}$. Due to the large deformation in the hyperelastic model, the displacement is at magnitude $10^{-1}$, and the relative error is below $0.1\%$. Although the absolute error in FE-NO is an order of magnitude higher than that in FE-FE, it remains negligible in relative terms, below $1\%$ compared to the reference solution of FE.

Similarly, in Fig. \ref{Fig:quasi_static_hyper_v}, the diagonal-symmetric results of \review{$u_y$} with respect to \review{$u_x$}  are shown in both domains, resulting from the diagonal symmetry of the loading and fixed boundary conditions.  The arc-shaped distribution of \review{$u_y$} can be found in $\Omega_{I}$ and $\Omega_{II}$ at $n = \{0,2,4\}$ with loading conditions increasing from $0.05$ to $0.25$.  Relative to edge length 2, it can be considered as a large deformation. The absolute errors in $y$-displacement for FE-FE are almost in the same range as those in $x$-displacement. Meanwhile, in FE-NO, the absolute error is slightly larger than in \review{$u_x$}.

Except for the NO predictions in $\Omega_{II}$, all absolute errors obtained by the FE solver reach the maximum at the edges of $\Omega_{II}$. The higher errors within $\Omega_{II}$ in \review{$|{u_x^4}_{FE} - {u_x^{4}}_{FE-NO, \Omega_{II}}|$} and \review{$|{u_y^4}_{FE} - {u_y^{4}}_{FE-NO, \Omega_{II}}|$} are probably derived from the intrinsic error of our trained PI-DeepONet. Due to coupling, \review{$|{u_x^4}_{FE} -{ u_x^4}_{FE-NO, \Omega_{II}}|$} and \review{$|{u_y^4}_{FE} - {u_y^{4}}_{FE-NO, \Omega_{II}}|$}  are slightly higher,  while at the same order of magnitude $10^{-3}$. Moreover, the large error region (indicated in light yellow) appears along the edges of $\Omega_{II}$, rather than being completely isolated, as observed in \review{$|{u_x^4}_{FE} - {u_x^{4}}_{FE\text{-}NO, \Omega_{II}}|$} and \review{$|{u_y^4}_{FE} - {u_y^{4}}_{FE\text{-}NO, \Omega_{II}}|$}.

% Given the diagonally symmetric loading, the absolute errors in FE-FE coupling also exhibit quasi-diagonal symmetry, though not strictly diagonal due to the influence of the Poisson effect. $|u^4_{FE} - u^{4, 48}_{FE-FE}|$ is inclined towards $x$-direction, while $|v^{4}_{FE} - v^{4,50}_{FE-FE}|$ towards $y$-direction.
\begin{figure}[H]
\begin{center}
\includegraphics[width=1\textwidth, height= 8.5cm]{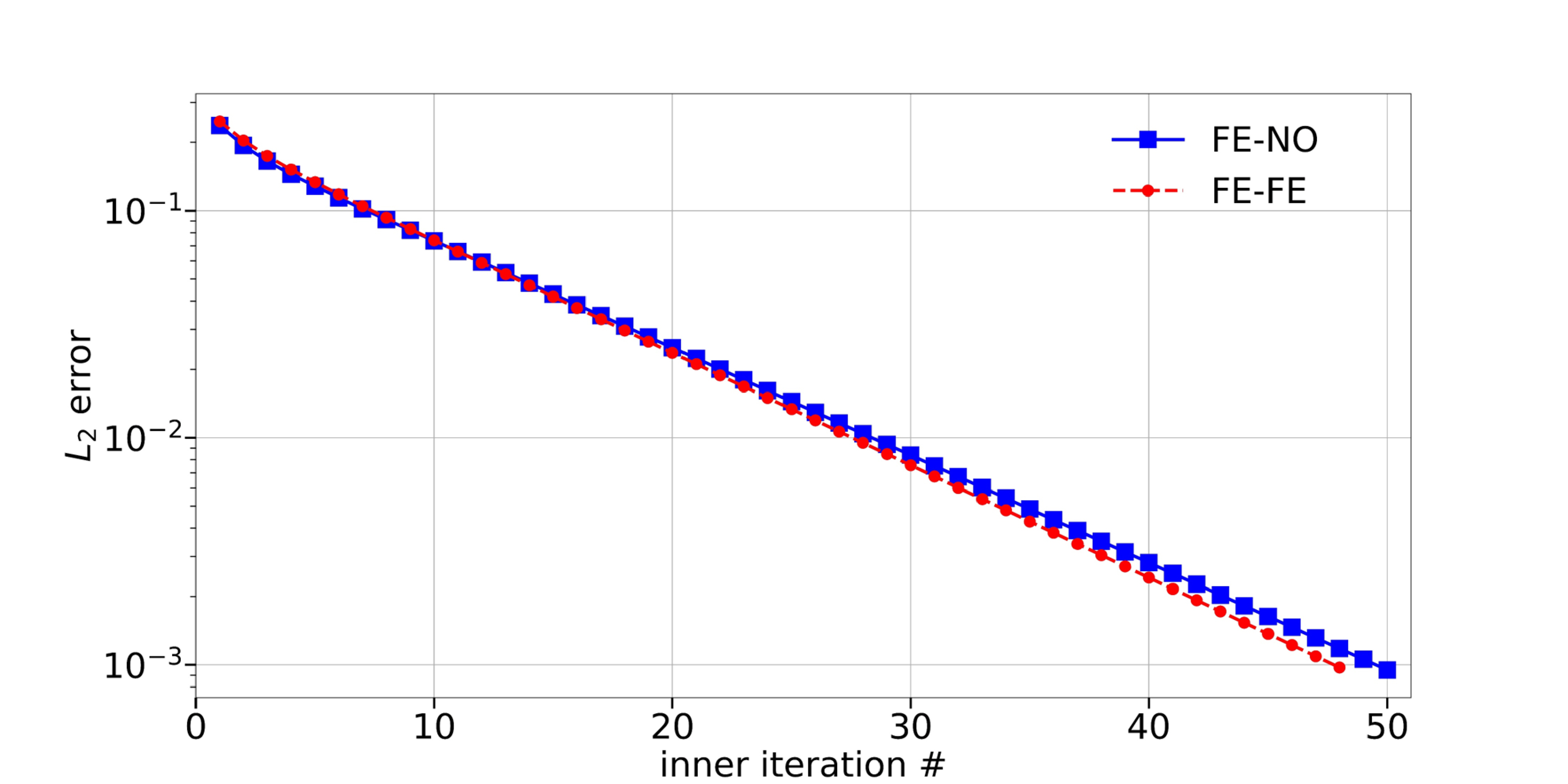}
\caption{$L^2$ error norm evolution across iterations for the hyper-elastic model under quasi-static loading conditions at time step $n =4$. }
\label{Fig:quasi_static_hyper_error_profile}
\end{center}
\end{figure}

Figure \ref{Fig:quasi_static_hyper_error_profile} presents the $L^2 $ error profiles of FE-FE and FE-NO in the hyperelastic model in quasi-static regime at time step $n =4$. The $L^2$ error is defined in Eq. \ref{L2 error}. As shown in Fig.~\ref{Fig:quasi_static_hyper_error_profile}, the error $L^2$ drops below the threshold $\epsilon = 10^{-3}$, indicating convergence of the coupling.

Thus, the displacement over the entire domain is obtained, as illustrated in Eq. \ref{u_domain1}-\ref{u_domain2}.  During the first 20 inner iterations, the FE-FE and FE-NO error profiles overlap closely. As the inner iterations progress, the deviation between them becomes increasingly noticeable. Ultimately, FE-FE reaches the convergence condition at 48 inner iterations, while FE-NO converges at 50 iterations. This indicates the comparable convergence behaviors and the robustness of both coupling frameworks. 

However, the computational efficiency in FE-FE and FE-NO is different. As mentioned in the linear elastic model in a static regime, the FE-FE coupling method needs recalculation by the FE solver in $\Omega_{II}$ at each inner iteration step. In the hyperelastic model, this process involves the Newton solver, necessitating additional iteration steps to obtain the solution at each inner iteration. In particular, at $n = 4$, the FE-FE approach takes 184.8 seconds for 49 inner iterations, whereas the FE-NO method completes 51 inner iterations in 160.8 seconds. Moreover, once trained, FE-NO leverages the direct evaluation of PI-DeepONet without requiring recalculations or further training. 

In addition to computational efficiency, the stability of the FE-FE framework is also challenged by the implementation of the nonlinear FE solver.  The nonlinear FE solver is prone to divergence due to various factors, such as large load increments, element distortion, and other highly nonlinear behaviors. In contrast, neural operators are specifically designed for nonlinear mappings in a mesh-free form, offering significant stability in simulating highly nonlinear models. 

\subsection{Neural operator coupling for dynamic loading}
\label{section_elasto_dynamic}

\begin{figure}[H]
\begin{center}
\includegraphics[width=1\textwidth, height= 10cm]{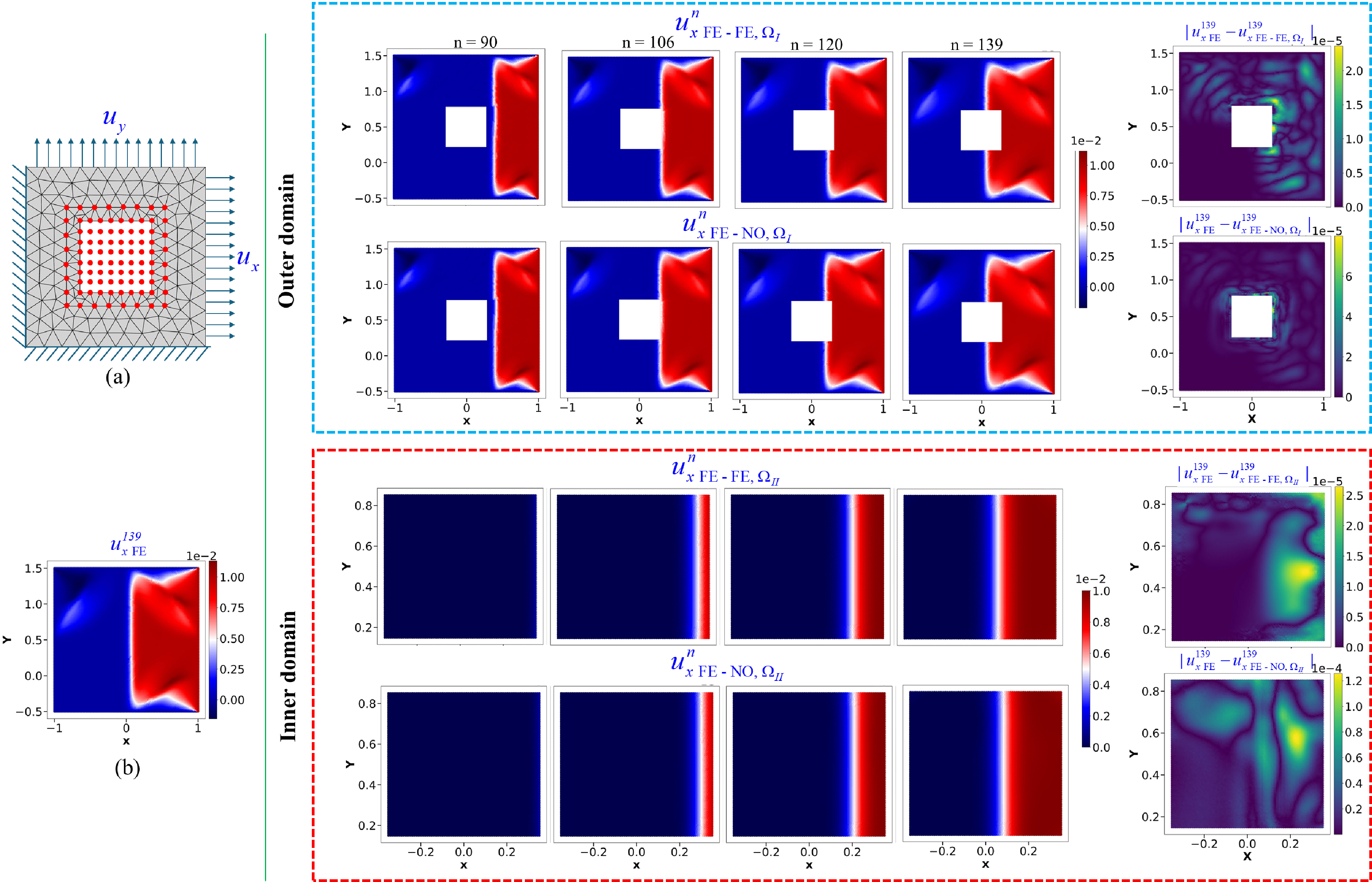}
\caption{Response in $x$-direction (\review{$u_x$}) of the linear elastic coupling model under dynamic loading conditions: (a) Schematics of decomposed domains for spatial coupling framework, where the left and bottom edges have fixed boundary conditions and the top and right edges are subjected to an applied displacement in $y$-direction (\review{$u_y = 0.01 $}) and $x$-direction (\review{$u_x = 0.01$}); (b) Ground truth displacement \review{$u_y$} at time step $n=139$ obtained by solving the intact domain using FEniCSx;  The blue-dashed box contains: Columns 1-4 showing the evolution of  \review{$u_x$} in $\Omega_{I}$ for FE-FE coupling (top row) and FE-NO coupling (bottom row) at time step $n = 90, 106, 120, 139$, with column 5 displaying the absolute error between the converged solution at time step $n = 139$ and ground truth. The red-dashed box contains: Columns 1-4 showing the evolution of \review{$u_x$} in $\Omega_{II}$ for both coupling frameworks at the same time steps, with column 5 displaying the absolute error relative to the ground truth.}
\label{Fig:elasto_dynamic_u}
\end{center}
\end{figure}

\begin{figure}[H]
\begin{center}
\includegraphics[width=1\textwidth, height= 10.5cm]{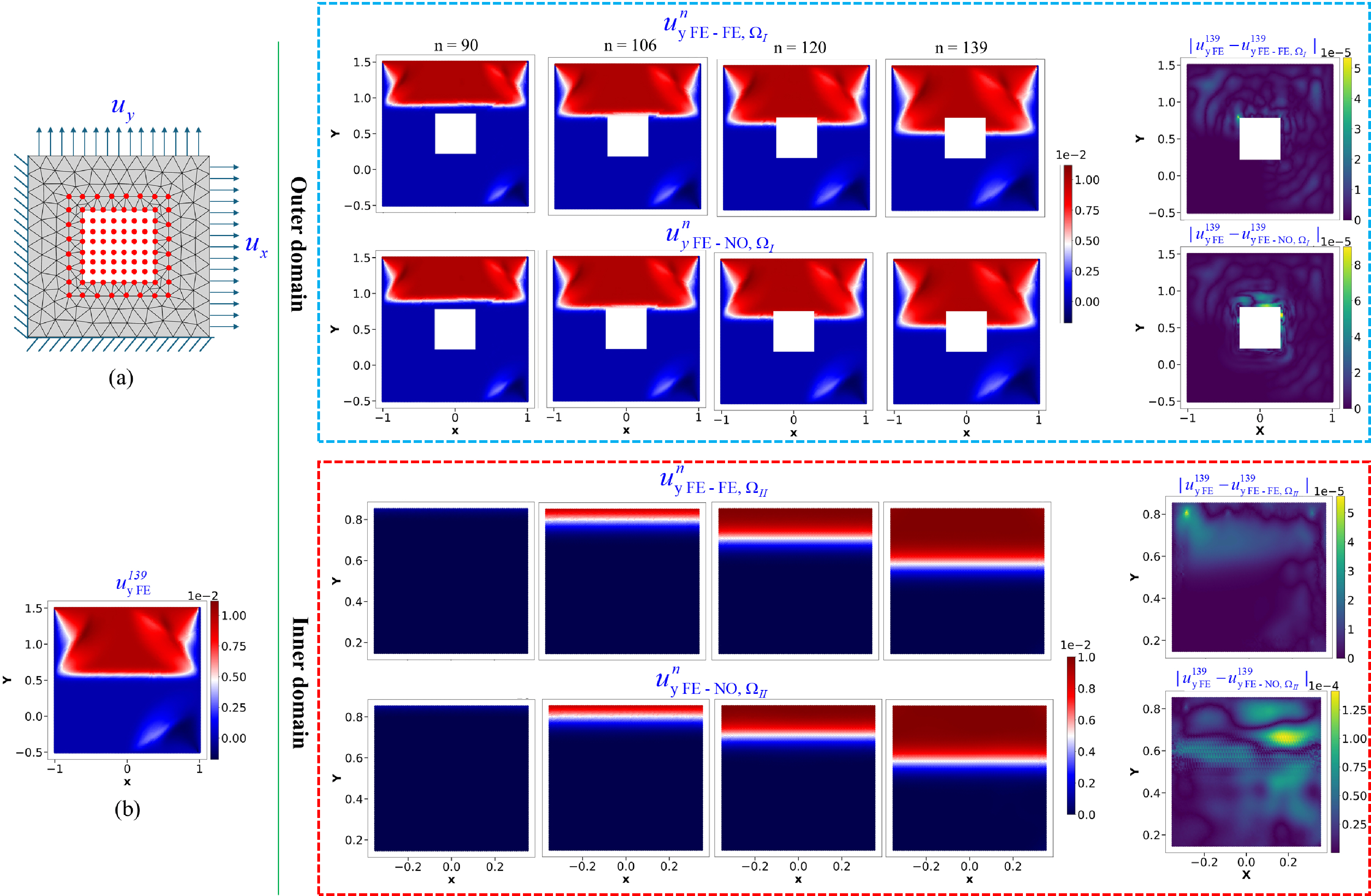}
\caption{Response in $y$-direction (\review{$u_y$}) of the linear elastic coupling model under dynamic loading conditions: (a) Schematics of decomposed domains for spatial coupling framework, where the left and bottom edges have fixed boundary conditions and the top and right edges are subjected to an applied displacement in $y$-direction (\review{$u_y = 0.01 $}) and $x$-direction (\review{$u_x = 0.01$}); (b) Ground truth displacement \review{$u_x$} at time step $n=139$ obtained by solving the intact domain using FEniCSx; the blue-dashed box contains: Columns 1-4 showing the evolution of  \review{$u_y$} in $\Omega_{I}$ for FE-FE coupling (top row) and FE-NO coupling (bottom row) at time step $n = 90, 106, 120, 139$, with column 5 displaying the absolute error between the converged solution at time step $n = 139$ and ground truth. The red-dashed box contains: Columns 1-4 showing the evolution of \review{$u_y$} in $\Omega_{II}$ for both coupling frameworks at the same time steps, with column 5 displaying the absolute error relative to the ground truth.}
\label{Fig:elasto_dynamic_v}
\end{center}
\end{figure}

In this elasto-dynamic example, the entire computational domain is a square. As shown in Fig. \ref{Fig:elasto_dynamic_u}(a), the schematic illustrates the DDM used in the FE-NO coupling framework. \review{The outer domain, $\Omega_{I}$, is a $2 \times 2$ square with an embedded $0.6 \times 0.6$ square hole, while the inner domain, $\Omega_{II}$, is a square with side length 0.7.} The boundary conditions remain consistent with those described in Section \ref{section_quasi_static}, except that a constant displacement is applied, with \review{$u_x = 0.01$} and \review{$u_y = 0.01$}. Figure \ref{Fig:elasto_dynamic_u}(b) and \ref{Fig:elasto_dynamic_v}(b) show the displacement results for the intact square domain at the final time step $n = 139$, obtained from the FE simulation in the intact domain. These results serve as the ground truth against which the accuracy of DDM-based methods is evaluated.

\begin{figure}[H]
\begin{center}
\includegraphics[width=1\textwidth, height= 13cm]{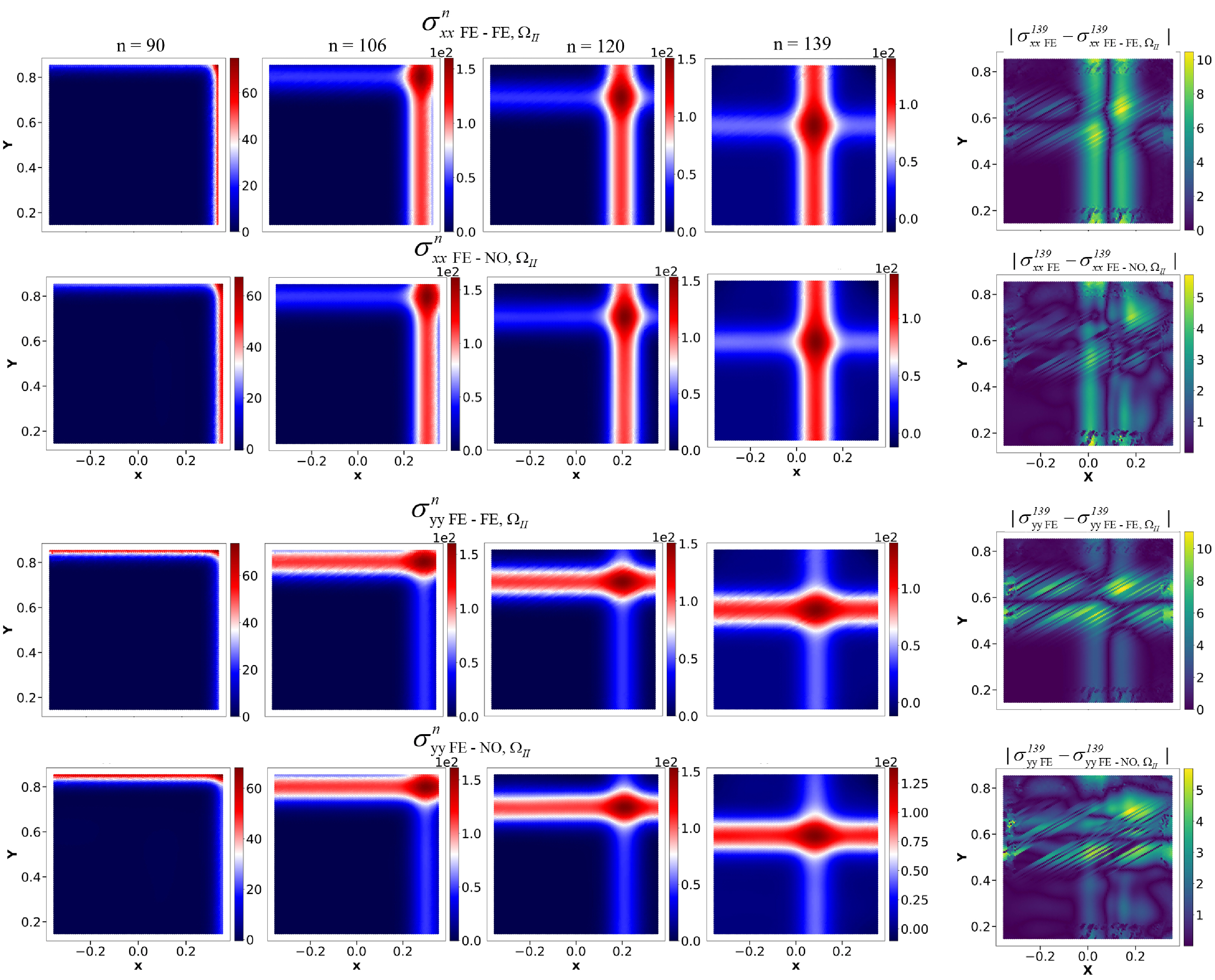}
\caption{Response in $x$-direction  ($\sigma_{xx}$) and $y$-direction ($\sigma_{yy}$) of the linear elastic coupling model under dynamic loading in $\Omega_{II}$: First two rows are the evolution of $\sigma_{xx}$ at $n = 90, 106, 120, 139$, obtained in FE-FE and FE-NO respectively, with column 5 displaying the absolute error between the converged solution at time step $n = 139$ and ground truth.  The last two rows are $ \sigma_{yy}$ for both coupling frameworks, with column 5 displaying the absolute error relative to the ground truth.}
\label{Fig:elasto_dynamic_stress}
\end{center}
\end{figure}

To establish an elastodynamic FE-NO coupling framework, we begin by implementing the FE-FE coupling approach. The domain is divided into two regions: $\Omega_I$ and $\Omega_{II}$. 
% These are discretized using FEniCSx into 61,114 and 21,490 elements, respectively. Both domains are equipped with 82 uniformly distributed nodes along the boundaries $\Gamma^{in}_{II}$ (and equivalently $\Gamma^{in}_{I}$) and $\Gamma^{out}_{II}$ to enable seamless data exchange. 
At each time step, data is transferred between the subdomains using the Schwarz alternating method, as outlined in Algorithm \ref{algo_1}. The discretization, computation, and information exchange closely follow the FE-FE coupling procedure used in the quasi-static hyperelastic model. The key difference under dynamic loading is the incorporation of the time dependency, which is handled using the Newmark method, as described in Eq. \ref{K_U=f_new}. The results of the FE-FE coupling under dynamic conditions are shown in Fig. \ref{Fig:elasto_dynamic_u}, \ref{Fig:elasto_dynamic_v}, and \ref{Fig:elasto_dynamic_stress}, demonstrating high accuracy and validating the effectiveness of this approach.

For the FE-NO coupling, a pretrained PI-DeepONet is used to transfer displacement information (\review{$u_x$} and \review{$u_y$}) across $\Gamma^{in}_{II}$ and $\Gamma^{out}_{II}$ at each time step. Since the system is dynamic and involves both acceleration and velocity, the time dimension must be considered. Therefore, the inner domain's PI-DeepONet is designed as an auto-regressive model. Moreover, to ensure compatibility with the Newmark method used in the FE analysis of the outer domain, the PI-DeepONet must also reflect the underlying time integration scheme. This motivates the development of a time-advancing PI-DeepONet architecture, inspired by the Newmark method, as illustrated in Fig. \ref{Fig:DeepONet_structure}(a). In this architecture, the Branch1 network incorporates current time step's boundary conditions, while the Branch2 network uses a convolutional neural network (CNN) to extract velocity and displacement features from the previous time step. The trunk network then predicts the current displacement at spatial coordinates $(x, y) \in \Omega_{II}$.

The time-advancing prediction process is further detailed in the flowchart in Fig. \ref{Fig:DeepONet_structure}(b). Functionally, this PI-DeepONet acts as a Newmark-integrated surrogate for the FE solver, allowing consistent prediction of displacements at each time step. Thus, the time-advancing PI-DeepONet and the FE solver are naturally coupled in the time domain. Meanwhile, spatial coupling is achieved via the Schwarz alternating method through boundary data exchange. Importantly, the time-advancing PI-DeepONet is not restricted to single-step predictions; it can sequentially predict multiple future time steps, thereby broadening its applicability to long-term dynamic simulations.

The next step involves training the time-advancing PI-DeepONet. Unlike earlier DeepONet models that rely solely on boundary conditions, the time-advancing variant requires full domain information as input to the branch network. Therefore, constructing a suitable dataset for PI-DeepONet pretraining is essential. This dataset is generated from a square domain solved using a FE solver, as shown in the reference ground truth results in Fig.~\ref{Fig:elasto_dynamic_u}(b) and Fig.~\ref{Fig:elasto_dynamic_v}(b). To begin, we generate $200$ Gaussian random field samples with length scale parameters $l_u = l_v = 0.2$ to replace the constant displacements \review{$u_x$} and \review{$u_y$}. The boundary conditions applied to the intact square are identical to those shown in Fig.~\ref{Fig:elasto_dynamic_u}(a), except that the constant values of \review{$u_x$} and \review{$u_y$} are substituted by the generated Gaussian random field data. We employ the Newmark method within the FE solver to simulate the dynamic response of this square domain. For each of the 100 random field samples, we extract the boundary conditions on $\Gamma^{\text{out}}_{II}$, as well as the domain displacements and velocities ($v_x$ and $v_y$) within $\Omega_{II}$, across 11 time steps. To train the PI-DeepONet for prediction over 10 specific time steps, we select the corresponding segment of the dataset. The domain displacements and velocities from time steps 0 through 10 are used as inputs to the Branch2 network, while the boundary conditions from time steps 1 through 11 serve as inputs to the Branch1 network. \review{As the domain displacements and velocities are generated from FEM with unstructured girds, the radial basis function interpolation in Scipy library \cite{2020SciPy-NMeth} should be utilized to obtain the structured data, which can be used as inputs of CNN.}This process yields 1,000 total training samples (10 time steps $\times$ 100 samples) for the displacement components \review{$u_x$}, \review{$u_y$}, and velocity components $v_x$, $v_y$, along with their corresponding boundary values on $\Gamma^{\text{out}}_{II}$. During training, batches of size 100 are randomly sampled at each step.

The PI-DeepONet is trained using the governing equations in strong form, as defined in Eq.~\ref{dynamic_equa}, by substituting \review{$u_x^n$} and \review{$u_y^n$} into the PI-DeepONet networks \review{$G^{u_x}_{\boldsymbol{\uptheta}_1}$} and \review{$G^{u_y}_{\boldsymbol{\uptheta}_2}$}, respectively. The acceleration is computed using Eq.~\ref{U_new}. The resulting physics-informed residual equations used for training are:
\begin{align}
&(\lambda + 2\mu)\frac{\partial^2 G^{u_x}_{\boldsymbol{\uptheta}_1}}{\partial x^2} + \mu \frac{\partial^2 G^{u_x}_{\boldsymbol{\uptheta}_1}}{\partial y^2} + (\lambda + \mu)\frac{\partial^2 G^{u_y}_{\boldsymbol{\uptheta}_2}}{\partial x \partial y} = \frac{-2}{(\beta\text{d}t)^2} \left( u_x^{n-1} + v_x^{n-1}\text{d}t -G^{u_x}_{\boldsymbol{\uptheta}_1}\right), (x,y)\in \Omega_{II} \label{ED_PI_NO_res0}\\
 &(\lambda + 2\mu)\frac{\partial^2 G^{u_y}_{\boldsymbol{\uptheta}_2}}{\partial y^2} + \mu \frac{\partial^2 G^{u_y}_{\boldsymbol{\uptheta}_2}}{\partial x^2} + (\lambda + \mu)\frac{\partial^2 G^{u_x}_{\boldsymbol{\uptheta}_1}}{\partial x \partial y} =  \frac{-2}{(\beta\text{d}t)^2} \left( u_y^{n-1} + v_y^{n-1}\text{d}t -G^{u_y}_{\boldsymbol{\uptheta}_2}\right), (x,y)\in \Omega_{II} \label{ED_PI_NO_res1}\\
&G^{u_x}_{\boldsymbol{\uptheta}_1} = u_x^n(x,y), (x,y) \in \Gamma^{out}_{II} \label{ED_PI_NO_bcs0}\\
&G^{u_y}_{\boldsymbol{\uptheta}_2}= u_y^n(x,y), (x,y) \in \Gamma^{out}_{II} \label{ED_PI_NO_bcs1}
\end{align}

It is important to note that the choice of 11 time steps in this study is arbitrary and can be extended to a larger number of steps, depending on the application requirements. After training the PI-DeepONet for 1 million iterations, the corresponding loss history is shown in Fig.~\ref{Fig:elasto_dynamic_DeepONet}(a). In this plot,  \(\mathcal{L}_{bcs}\) refers to the boundary condition loss, and \(\mathcal{L}_{res}\) denotes the residual loss, which correspond to Eq.~\ref{ED_PI_NO_bcs0}–\ref{ED_PI_NO_bcs1} and Eq.~\ref{ED_PI_NO_res0}–\ref{ED_PI_NO_res1}, respectively. Before proceeding to the coupling process, we evaluate the trained PI-DeepONet using an unseen test case. Specifically, we apply the domain information at the time step $n = 112$ and the boundary conditions at $n = 113$, taken from the ground truth solution of an intact square domain under constant dynamic loading simulated using FEniCSx. \review{When comparing the predicted results to the ground truth, the absolute errors in displacement components are found to be \review{\( |{u_x^{113}}_{\mathrm{FE}, \Omega_{II}} - {u_x^{113}}_{\mathrm{NO}, \Omega_{II}}| < 5 \times 10^{-5} \)} and \review{\( |{u_y^{113}}_{\mathrm{FE}, \Omega_{II}} - {u_y^{113}}_{\mathrm{NO}, \Omega_{II}}| < 7 \times 10^{-5} \)}, which is depicted in Fig. \ref{Fig:elasto_dynamic_DeepONet}}. These results demonstrate that the PI-DeepONet is well-trained and capable of accurate generalization to unseen dynamic conditions.

After training the PI-DeepONet surrogate model, we can couple it with the FE solver over $\Omega_{I}$. At each time step, the information transfer is carried out following the Algorithm. \ref{algo_1}. The coupling process for each inner iteration at each time step remains the same as described in Section \ref{section_quasi_static}. For the FE-FE coupling, the process remains the same, with the only difference being that $\Omega_{II}$ is solved using an FE solver. Instead of repeating the inner iteration process at each time step in the elastodynamic case, we present only the converged result at every time step, requiring that $L^2$ error falls below the critical threshold $\epsilon = 10^{-5}$. For elastodynamic results, we use the same notations for converged results \review{${u_x^{n}}_{FE-FE}$, ${u_y^{n}}_{FE-FE}$, ${u_x^{n}}_{FE-NO}$} and \review{${u_y^{n}}_{FE-NO}$} as in Section \ref{section_quasi_static}, with $n$  the superscript indicating the time step.

Fig. \ref{Fig:elasto_dynamic_u} shows the evolution of displacement \review{$u_x$} for both coupling frameworks from time step $n = {90, 106, 120, 139}$.  Note that the domain information at $n = 89$ is obtained from the FE-FE coupling framework. Based on the \review{${u_x^{89}}_{FE-FE,\Omega_{II}}, {u_y^{89}}_{FE-FE,\Omega_{II}}, {v_x}^{89}_{FE-FE,\Omega_{II}},{v_y}^{89}_{FE-FE,\Omega_{II}},\\ {u_x^{90}}_{|\Gamma^{out}_{II}}$} and  \review{${u_y^{90}}_{|\Gamma^{out}_{II}}$},  we can calculate and predict domain displacements \review{${u_x^{90}}_{\Omega_{II}}$} and \review{${u_y^{90}}_{\Omega_{II}}$} in FE-FE and FE-NO, respectively.

At $n = 90$, \review{${u_x^{90}}_{FE-FE, \Omega_{I}}$} and \review{${u_x^{90}}_{FE-NO, \Omega_{I}}$} both show that the high displacement values have not reached the domain $\Omega_{II}$. The entire blue region in \review{${u_x^{90}}_{FE-FE, \Omega_{II}}$} and \review{${u_x^{90}}_{FE-NO, \Omega_{II}}$} confirms that displacements are small in $\Omega_{II}$. Due to the negligible displacement value in $\Omega_{II}$, the inner iteration until time step $90$ only requires 5 steps in FE-NO and 6 steps in FE-FE. However, as soon as high displacement values reach $\Omega_{II}$, a higher number of inner iterations (e.g., $9-10$) are required for convergence.

After 16 time steps ($n = 106$), it is found in both \review{${u_x^{106}}_{FE-FE, \Omega_{I}}$} and \review{${u_x^{106}}_{FE-NO,\Omega_{I}}$} that high displacement values \review{$u_x$} reach the right edge of $\Gamma^{in}_{I}$, and have already occupied the right area of the overlap region ($\Omega_{o}$). 

% As shown in $u^{106,10}_{FE-FE,\Omega_{II}}$ and $u^{106,9}_{FE-NO,\Omega_{II}}$, the red region exists from $x=0.3$ to $0.35$, while the highest value representing by dark red can be hardly found.

% A straight transition interface splitting the blue and red (indicated by color white) is observed.

After another 14 time steps ($n = 120$), the $x$-displacement surpasses the $\Gamma^{in}_{I}$, stepping into the center hole of $\Omega_{I}$. The upper and lower wings of the displacement in \review{${u_x^{120}}_{FE-FE,\Omega_{I}}$} and \review{${u_x^{120}}_{FE-NO,\Omega_{I}}$} uniformly propagate, and no reflection or sharp gradients of displacement field can be observed. This indicates the continuity between the two domains and the convergence of the inner iterations. 

% The dark red is observed in $u^{120,10}_{FE-FE,\Omega_{II}}$ and $u^{120,10}_{FE-NO,\Omega_{II}}$ along with the straight transition interface. 

After an additional 20 time steps ($n = 139$), the higher values of \review{$u_x$} propagate further toward the left edge of the square. 
% Comparing the displacement $u_x$ in both frameworks with the ground truth in Fig. \ref{Fig:elasto_dynamic_u}(b), the corresponding absolute errors are exhibited in column 5 of the blue-dashed and red-dashed box.  

The absolute error \review{$|{u_x^{139}}_{FE} - {u_x^{139}}_{FE-FE}|$} in both $\Omega_{I}$ and $\Omega_{II}$ is on the order of magnitude $10^{-5}$, while \review{$|{u_x^{139}}_{FE} - {u_x^{139}}_{FE-NO}|$} is slightly higher in both domains. Specifically, the highest value reaches $8\times 10^{-5}$ in $\Omega_{I}$ and $1.25 \times 10^{-4}$ in $\Omega_{II}$.  In both coupling frameworks, the highest absolute errors in $\Omega_{I}$ can be found around $\Gamma^{in}_{I}$, especially for FE-NO coupling. 

% It is consistent that the maximum of the absolute error in $\Omega_{II}$ also shows up near the right edge $\Gamma^{in}_{I}$ ($x = 0.3$), where is occupied by the highest displacement value and the boundary displacements passed back to $\Omega_{I}$. 
 
In Fig. \ref{Fig:elasto_dynamic_v}, the evolution of displacement in $y$ direction, \review{$u_y$}, from $n = 90$ to $139$ is illustrated. In the first time steps, the red region of displacement uniformly propagates towards the bottom of the domain, then it gradually extends into $\Omega_{II}$ and propagates along with the transition interface as shown by a white line. Column 5 of the blue-dashed box and red-dashed box show the \review{$|{u_y^{139}}_{FE} - {u_y^{139}}_{FE-FE} |$} and  \review{$|{u_y^{139}}_{FE} - {u_y^{139}}_{FE-NO} |$}. The order magnitude of the absolute error is the same as that in displacement \review{$u_x$}. 

Regarding the absolute error in $\Omega_{I}$, the maximum error occurs in $\Gamma^{in}_{I}$. Similarly, the highest error in \review{$|{u_y^{139}}_{FE} - {u_y^{139}}_{FE-FE, \Omega_{II}} |$} is also observed near $\Gamma^{in}_{I}$. In contrast, the maximum error in \review{$|{u_y^{139}}_{FE} - {u_y^{139}}_{FE-NO, \Omega_{II}} |$} appears near the transition interface between the blue and red regions. As noted in \cite{wang2024causality}, such relatively sharp interfaces are particularly challenging to accurately approximate in scientific machine learning frameworks.
 
\begin{figure}[H]
\begin{center}
\includegraphics[width=0.9\textwidth, height= 10cm]{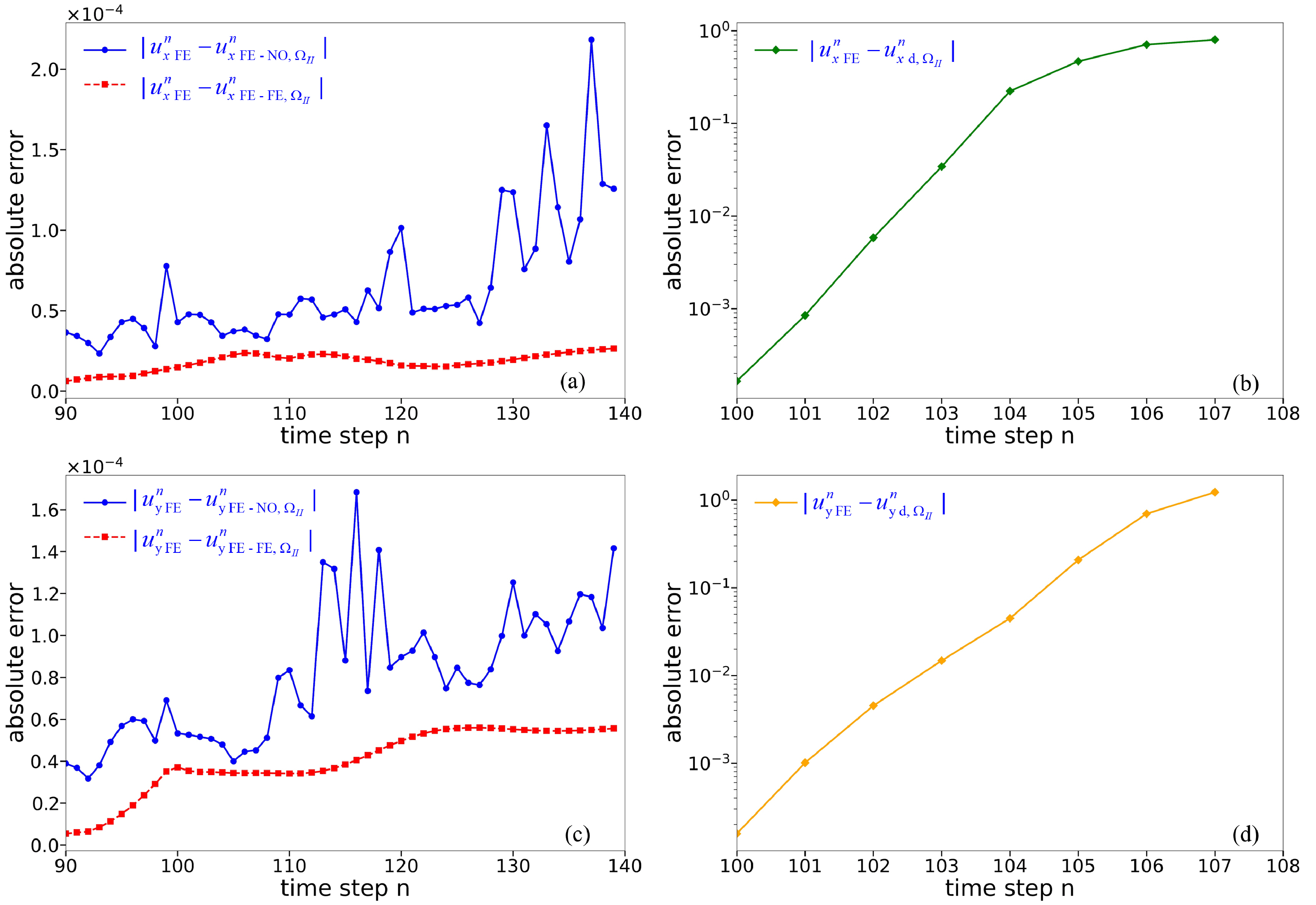}
\caption{The maximum absolute error value in $\Omega_{II}$ : (a)  absolute error of \review{$u_x$} from time step $n =90$ to $139$ for FE-NO (\review{$|{u_x^n}_{FE} - {u_x^n}_{FE-NO}| $}) and FE-FE (\review{$|{u_x^n}_{FE} - {u_x^n}_{FE-FE}| $}) coupling  (b) absolute error of \review{$u_x$} by auto-regressive data-driven DeepONet. (c) and (d) are for \review{$u_y$}}
\label{Fig:error_dynamic_uv}
\end{center}
\end{figure}

The detailed evolution of the absolute error for both coupling frameworks is depicted in Fig.\ref{Fig:error_dynamic_uv}. It is noteworthy that the auto-regressive method is used in both coupling frameworks, meaning that information from the previous time step is used to derive the results for the next time step. As a result, if there is an error in the previous time step, it will be carried over to the next time step's result. This is a commonly observed issue in auto-regressive models within scientific machine learning, where error accumulation often leads to an exponential increase in error, potentially causing it to exceed several orders of magnitude \cite{michalowska2024neural}.  In Fig. \ref{Fig:error_dynamic_uv}(a, c), the absolute errors for both FE-FE and FE-NO have an increased tendency. The \review{$|{u_x^n}_{FE} - {u_x^n}_{FE-FE}|$} changes smoothly, remaining within a small value range (below $5 \times 10^{-5}$), while the \review{$|{u_x^n}_{FE} - {u_x^n}_{FE-NO}|$} varies more drastically, exhibiting abrupt fluctuations, specifically in the last 10 time steps. Similarly, the \review{$|{u_y^n}_{FE} - {u_y^n}_{FE-FE}|$} keeps increasing in smooth manner, and the \review{$|{u_y^n}_{FE} - {u_y^n}_{FE-NO}|$} oscillates abruptly. Notably, in \review{$|{u_y^n}_{FE} - {u_y^n}_{FE-NO}|$} from $n = 113$ to $n =118$, the error value jumps steeply, showing three spikes with the highest value $1.68 \times 10^{-4}$ at $n=116$. Despite these error spikes, the absolute error \review{$|{u_y^n}_{FE} - {u_y^n}_{FE-NO}|$} self-corrects to lower values (approximately $1.02\times10^{-4}$) during time steps $n = 119-129$, demonstrating the resilience of the framework. This self-heal phenomenon indicates that the auto-regressive error accumulation is not dominant in FE-NO coupling, though it is an auto-regressive surrogate model. We hypothesize that the significant error attenuation is because of the accurate boundary conditions provided by the FE solver in $\Omega_{I}$. 

We compare our model against a purely data-driven DeepONet to demonstrate the substantial accuracy improvements achieved through the incorporation of boundary conditions. This data-driven DeepONet implements the identical dataset utilized in time-advancing PI-DeepONet. The 100 samples of domain information (\review{$u_x$} and \review{$u_y$}) at each time step from $n = 99$ to $108$  ($100 \times  10 $ samples in total) are provided as input of the branch network. The corresponding 100 samples for each time step from $n = 100$ to $109$ are set as the output. Notably, the architecture of the data-driven model is identical to the elastodynamic model in Table \ref{tab:NO_architecture}, except that the Branch1 network responsible for enforcing boundary conditions is removed. We train this data-driven DeepONet to predict the next time step's displacements in $x$-direction and $y$-direction. Consequently, we obtain two auto-regressive models for $x$-displacement and $y$-displacement, which predict the displacement in $\Omega_{II}$ only depending on the domain displacement from the previous time step without boundary conditions. The accurate results for $n = 99$ are provided by the FE solver. The data-driven DeepONet predicts the displacements in $\Omega_{II}$, denoted as \review{${u_x^n}_{d,\Omega_{II}}$} and \review{${u_y^n}_{d,\Omega_{II}}$}. The corresponding absolute error \review{$|{u_x^n}_{FE} - {u_x^n}_{d, \Omega_{II}}|$} and \review{$|{u_y^n}_{FE} - {u_y^n}_{d, \Omega_{II}}|$} are shown in Fig. \ref{Fig:error_dynamic_uv}(b, d). Both exhibit enormous increases of the order of magnitude $10^{-4}$ to $10^0$. In the first step at $n = 100$, both absolute errors are smaller than $1 \times 10^{-4}$ exhibiting the accuracy of the data-driven DeepONet for the trained timestep. However, when this model is used to predict the next 7 steps (i.e., from $n = 100$ to $107$), the auto-regressive error accumulates significantly, resulting in an increase of five orders of magnitude. This highlights the superiority of our time-advancing PI-DeepONet over the data-driven DeepONet.

Although the error accumulation in our FE-NO framework seems unavoidable in the current method, the absolute error for \review{$u_x$} and \review{$u_y$} are still lower than $3 \times 10^{-4}$ after 49 time-steps, while the displacement values of \review{$u_x$} and \review{$u_y$} are at $10^{-2}$ orders. To restrict the FE-NO error within a bounded threshold, we perform FE-FE coupling for a few timesteps after a few FE-NO coupling for significantly large timesteps, as a recalibration for the coupling framework.  As displayed in Fig. \ref{Fig:error_dynamic_uv}(a,c), after 139-time steps, the absolute error in FE-FE is still in $10^{-5}$, varying continuously and smoothly without any spikes.   

Figure~\ref{Fig:elasto_dynamic_stress} presents the stress fields for the elastodynamic example. The stress distributions in both coupling frameworks (i.e, FE-FE and FE-NO) are consistent with the displacement fields, with $\sigma_{xx}$ and $\sigma_{yy}$ propagating toward the left and bottom edges, respectively.
The displacements \review{${u_x^n}_{\Omega_{II}}$} and \review{${u_y^n}_{\Omega_{II}}$}, shown in Fig. \ref{Fig:elasto_dynamic_u} and Fig. \ref{Fig:elasto_dynamic_v} remain nearly constant within the red and blue regions, separated by a single transition interface. As a result, $\sigma_{xx}$ and $\sigma_{yy}$ exhibit stripe-like patterns centered around this displacement interface. Due to the Poisson effect, a light blue stripe intersects a red stripe in both $\sigma_{xx}$ and $\sigma_{yy}$, leading to a localized maximum in the intersection region.

Due to the use of the first-order continuous Galerkin method, stress discontinuities appear in the FE-FE coupling results. These are particularly evident in the white regions of $\sigma^{106}_{yy, FE-FE,\Omega_{II}}$, $\sigma^{120}_{yy, FE-FE,\Omega_{II}}$ and $\sigma^{139}_{yy, FE-FE,\Omega_{II}}$. In contrast, FE-NO coupling stress distributions are smoother because of the high-order differentiability. Therefore, in absolute errors, $|\sigma^{139}_{xx,FE} - \sigma^{139}_{xx,FE-NO,\Omega_{II}}|$ and $|\sigma^{139}_{yy,FE} - \sigma^{139}_{yy,FE-NO,\Omega_{II}}|$ are slightly lower than $|\sigma^{139}_{xx,FE} - \sigma^{139}_{xx,FE-FE,\Omega_{II}}|$ and $|\sigma^{139}_{yy,FE} - \sigma^{139}_{yy,FE-FE,\Omega_{II}}|$.  

It is worth noting that increasing the element order in finite element methods can mitigate the stress discontinuity issue. However, this comes at the cost of significantly increased computational effort, resulting in reduced efficiency. Instead of compromising between accuracy and efficiency in traditional FE solvers, employing neural operators to solve high-order PDEs presents a promising alternative.

\subsection{Expansion of the ML-subdomain}
\label{section_puzzle}
In Section \ref{section_elasto_dynamic}, the computational domain was pre-partitioned into two subdomains, $\Omega_I$ and $\Omega_{II}$, with $\Omega_{II}$ replaced by a DeepONet surrogate. Therefore, during the simulation in Section \ref{section_elasto_dynamic}, the DeepONet region remained fixed. However, in realistic multi-scale and dynamic systems, computationally intensive regions often evolve over time. To optimize the performance of our framework in such scenarios, we propose the dynamic expansion of the DeepONet region, which allows the DeepONet domain to expand adaptively in a tile-like manner following Algorithm~\ref{algo_2}. 

In this subsection, we consider the same global computational domain and boundary conditions as in Section \ref{section_elasto_dynamic}. As shown in Fig.~\ref{Fig:Schematics2}, the DeepONet region $\Omega_{II}$ is now extended to include two additional square domains, $\Omega_{II}$ and $\Omega_{III}$, positioned 0.3 units above and below $\Omega_{II}$, respectively, and having the same length (0.7 units). This configuration results in a larger rectangular surrogate composed of three adjacent DeepONet domains. \review{The outer domain $\Omega_I$ is a $2 \times2$ square with a $0.12 \times 0.6$ rectangle hole.}

To train these two time-advancing PI-DeepONet, we implement the same method as illustrated in previous section. For $\Omega_{II}$, the DeepONet is trained with suitable dataset extracted from the intact square domain solved by FE solver. For the DeepONet on $\Omega_{III}$, we have implemented a transfer learning strategy\cite{goswami2020transfer}, leveraging the trained model for $\Omega_{II}$. These two PI-DeepONet can also be trained using Eq. \ref{ED_PI_NO_res0}, \ref{ED_PI_NO_res1}, \ref{ED_PI_NO_bcs0} and \ref{ED_PI_NO_bcs1}.  We refer to the DeepONet for $\Omega_{II}$ as upper DeepONet, $\Omega_{III}$ as lower DeepONet. After training the two PI-DeepONets in the FE-NO coupling framework, we first couple them together. The combined DeepONet model is then treated as a single solver and coupled with the FE solver in $\Omega_{I}$, as outlined in Algorithm~\ref{algo_2}. The two different $L^2$ errors, $L^2_{p_1}$ and $L^2_{p_2}$, defined in Eq. \ref{L2_p1 error} and Eq. \ref{L2_p2_error}, should be lower than the critical thresholds $\epsilon= 1\times10^{-5}$ and $\epsilon'= 1\times10^{-5}$, respectively. In this FE-NO coupling framework, FE is discretized into 47,828 elements, with 162 nodes uniformly distributed along the left and right edges of $\Gamma^{out}_{inner}$ and 82 nodes on the top and bottom edges. 

For FE-NO coupling, we first solve $\Omega_I$ using FEniCSx and pass the displacement information at $\Gamma^{out}_{inner}$ to $\Omega_{II} \cup \Omega_{III}$. Having received the boundary conditions provided from $\Omega_{I}$ and $\Omega_{III}$ (at first coupling iteration of $\mathbf{u}_{|\Gamma^{in,2}_{inner}}=\mathbf{0}$, as detailed in Algorithm \ref{algo_2}), the upper DeepONet predicts the displacement throughout $\Omega_{II}$, giving the values of \review{$u_{x|\Gamma^{in,3}_{inner}}$} and \review{$u_{y|\Gamma^{in,3}_{inner}}$} that are passed to $\Omega_{III}$. Then, based on the boundary conditions from $\Omega_{I}$ and $\Omega_{II}$, the displacement over $\Omega_{III}$ is obtained, which provides the value \review{$u_{x|\Gamma^{in,2}_{inner}}$} and \review{$u_{y|\Gamma^{in,2}_{inner}}$} to $\Omega_{II}$. Information exchange between $\Omega_{II}$ and $\Omega_{III}$ continues until the iteration converges, as calculated in Eq. \ref{L2_p1 error}.  Then, the $\Omega_{II} \cup \Omega_{III}$ predicted by the DeepONet passes the \review{$u_{x|\Gamma^{out}_{inner}}$} and \review{$u_{y|\Gamma^{out}_{inner}}$} back to the $\Omega_I$. Consequently, the FE solver updates the displacement in $\Omega_I$, providing the new boundary conditions to $\Omega_{II} \cup \Omega_{III}$. It is important to note that once convergence is achieved for the coupling between $\Omega_{II}$ and $\Omega_{III}$, each subsequent update from $\Omega_{I}$ requires only one additional iteration to satisfy the convergence criterion $L^2_{p_1} < 1 \times 10^{-5}$. Information is continuously transferred between $\Omega_I$ and $\Omega_{II} \cup \Omega_{III}$, until the inner iteration convergence condition $L^2_{p_2} < 1 \times 10^{-5}$ is achieved. Eventually, we can get $\mathbf{u}^n_{|\Omega}$ by composing the subdomains of the results $I-III$. In this example, we use the same notations as Section \ref{section_elasto_dynamic} \review{${u_x^{n}}_{FE-NO}$} and \review{${u_y^{n}}_{FE-NO}$} to denote the converged $x$ and $y$ displacement at time step $n$ obtained using FE-NO coupling.

\begin{figure}[H]
\begin{center}
\includegraphics[width=0.85\textwidth, height= 10cm]{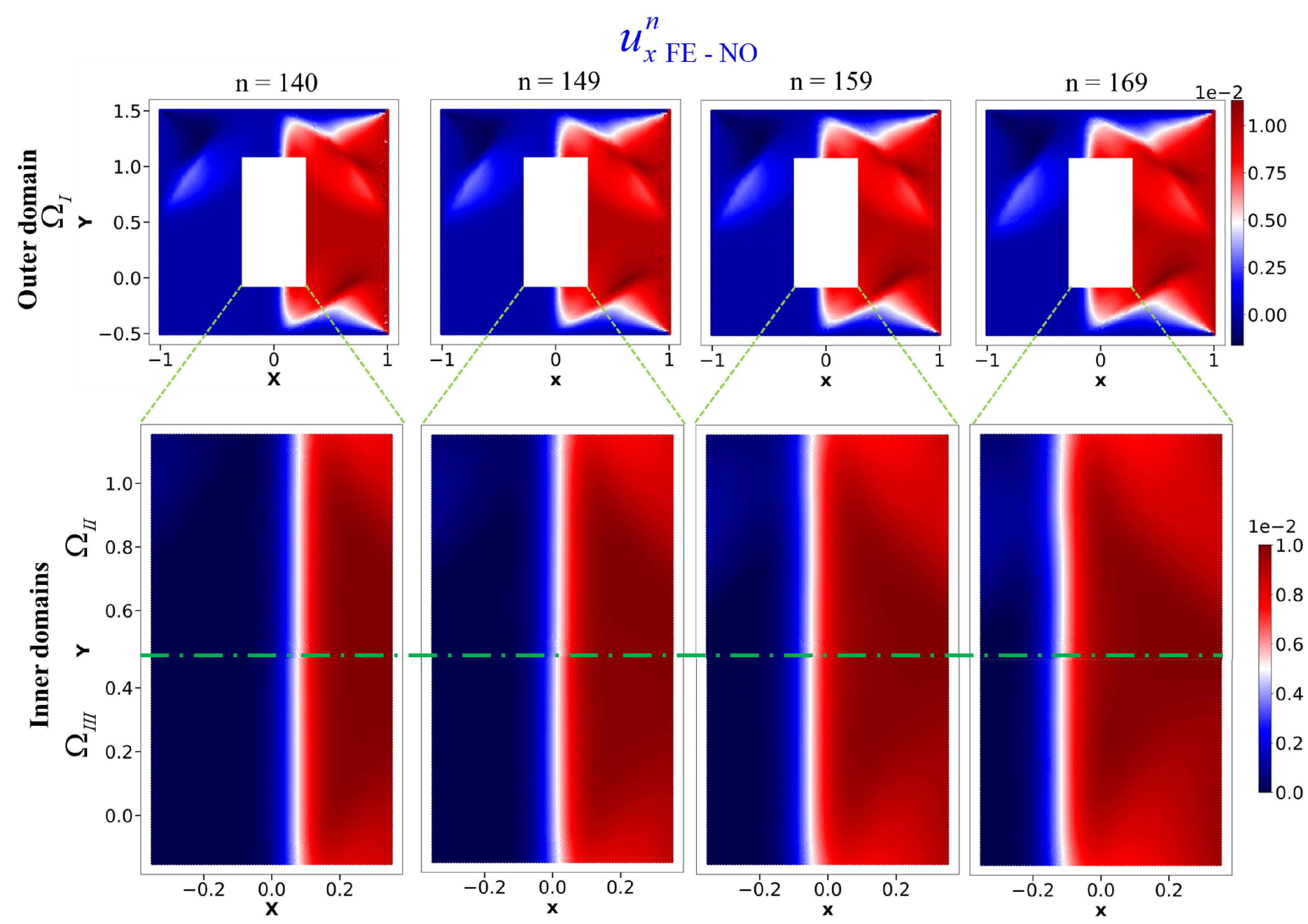}
\caption{Response in $x$-direction (\review{$u_x$}) of NO replacement for linear elastic FE-NO coupling model under dynamic loading conditions: first row showing the evolution of  \review{$u_x$} at time step $n = \{140, 149,159,169\}$ in  $\Omega^{p}_{I}$, second row and third row displaying the evolution in $\Omega_{II}$  and $\Omega^{p}_{III}$, respectively.}
\label{Fig:puzzle_u}
\end{center}
\end{figure}

\begin{figure}[H]
\begin{center}
\includegraphics[width=0.95\textwidth, height= 10.5cm]{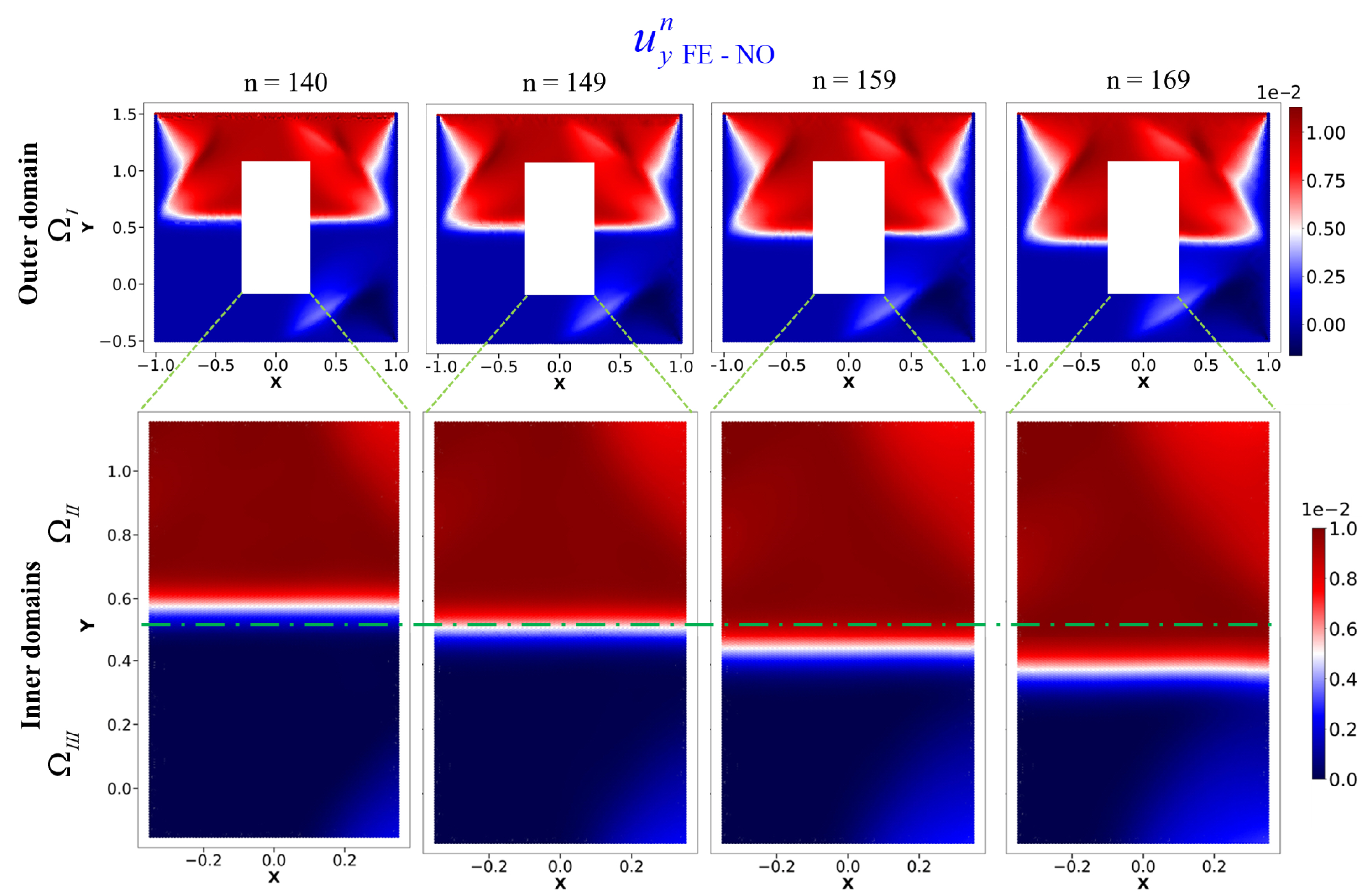}
\caption{Response in $y$-direction (\review{$u_y$}) of NO replacement for linear elastic FE-NO coupling model under dynamic loading conditions: first row showing the evolution of \review{$u_y$} at time step $n = \{140, 149,159,169\}$ in $\Omega^{p}_{I}$, second row and third row displaying the evolution in $\Omega_{II}$  and $\Omega^{p}_{III}$, respectively.}
\label{Fig:puzzle_v}
\end{center}
\end{figure}

Applying the domain information obtained at $n = 139$ by FE-FE coupling in previous Section \ref{section_elasto_dynamic}, further displacement propagation of the dynamic system is illustrated in Fig. \ref{Fig:puzzle_u} - \ref{Fig:puzzle_v}. 

Fig. \ref{Fig:puzzle_u} illustrates the evolution of \review{${u_x}_{FE-NO}$} in $\Omega_{I}$, $\Omega_{II}$ and $\Omega_{III}$ at time step $n = 140, 149, 159, 169$. At $n = 140$, the displacement \review{$u_x$} in $\Omega_{I}$ is similar to that in Fig. \ref{Fig:elasto_dynamic_u}, with two triangle wings on the upper and lower regions. In $\Omega_{II}$ and $\Omega_{III}$, the almost constant results show up, uniformly distributed red and blue regions along with a transition interface. After 9 time steps ($n = 149$), the transition interface of displacement \review{$u_x$} almost reaches the central axis ($x = 0$). The non-uniformity can be observed in the red region of $\Omega_{II}$, where the right top is slightly lighter than the others.  The transition interface is still relatively straight. After another 10 time steps (n = 159), the displacement \review{$u_x$} has already surpassed the central axis ($x = 0$), and the non-uniformity becomes more pronounced. Moreover, the transition interface is no longer a straight line; instead, it evolves into a curved shape-particularly near the upper end in $\Omega_{II}$ and the lower end in $\Omega_{III}$. This phenomenon becomes more pronounced after an additional 10 time steps at $n = 169$, where the transition interface in \review{${u_x^{169}}_{{FE-NO}, \Omega_{II}}$} exhibits a more prominent curvature. Additionally, the light red region extends toward the upper-right corner. In $\Omega_{III}$, a similar light red area appears symmetrically in the lower-right corner; however, the transition interface in $\Omega_{III}$ is not symmetrically curved. 

The difference of the transition interface in $\Omega_{II}$ and $\Omega_{III}$ originates from the Poisson's effect. The dark blue propagates from the left-top region to the right-bottom region. After red region passing the central axis, the dark blue influences the transition interface. Due to fixed boundary conditions on the left and bottom edges, there is no displacement wave from the left-bottom region. Therefore, the transition interface is not greatly influenced.

The evolution of \review{$u_y$} from $n = 140$ to $169$ is shown in Fig.~\ref{Fig:puzzle_v}, where the non-uniformity becomes clearly noticeable. Despite the non-uniformity and transition interface variance for \review{$u_y$}, the more intriguing process is the displacement propagation from $\Omega_{II}$ to $\Omega_{III}$. 

At $n = 140$, the displacement \review{$u_y$} in $\Omega_{II}$ is observed to remain nearly constant at approximately $1 \times 10^{-2}$. In contrast, the value of \review{${u_y^{140}}_{{FE-NO}, \Omega_{III}}$} is significantly lower, with a maximum around $3 \times 10^{-3}$.  The light blue region in the lower-right corner is attributed to the Poisson effect, as previously discussed. After 9 additional time steps ($n = 149$), the displacement \review{$u_y$} continues to propagate toward the bottom edge, with its maximum value in $\Omega_{III}$ reaching $6.7 \times 10^{-3}$-more than twice the value observed at $n = 140$. After another 9 time steps ($n = 159$), the maximum value reaches $1 \times 10^{-2}$, indicating that the uniform displacement field has extended into $\Omega_{III}$. Finally, at $n = 169$, the minimum of \review{$u_y$} in $\Omega_{II}$ is higher than $7.5\times 10^{-3}$, indicating that the transition interface leaves $\Omega_{II}$ to $\Omega_{III}$. 

\begin{figure}[H]
\begin{center}
\includegraphics[width=1\textwidth, height= 15cm]{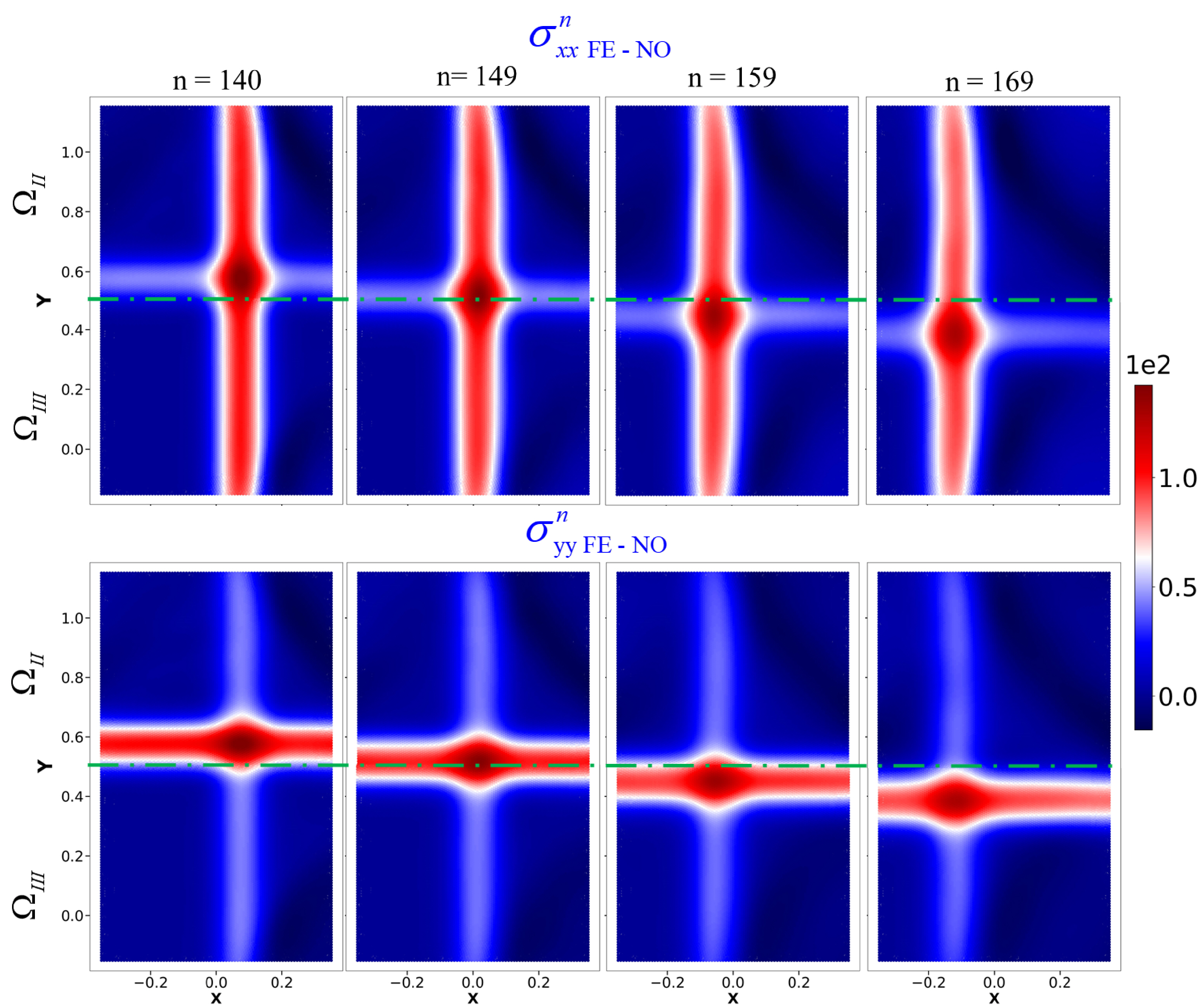}
\caption{Response in $x$-direction ($\sigma_{xx}$) and $y$-direction ($\sigma_{yy}$) of NO replacement for linear elastic coupling model under dynamic loading in $\Omega_{II}$: First two rows are the evolution of $\sigma_{xx}$ at $n = 140, 149 , 159, 169$ in $\Omega_{II}$ and $\Omega_{III}$, respectively, last two rows are $ \sigma_{yy}$ at corresponding time steps in $\Omega_{II}$ and $\Omega_{III}$ }
\label{Fig:puzzle_stress}
\end{center}
\end{figure}

In Fig. \ref{Fig:puzzle_stress}, the evolution of stress $\sigma_{xx}$ and $\sigma_{yy}$ in $\Omega_{II}$ and $\Omega_{III}$ is depicted. It is evident that the $\sigma_{xx}$ is not straight at the upper end in $\Omega_{II}$ and lower end in $\Omega_{III}$, confirming the curved transition interface obtained in Fig. \ref{Fig:elasto_dynamic_u}. This adaptive ML-subdomain expansion strategy offers a powerful tool for enabling real-time optimization of computational resources. This approach is particularly promising for dynamic multi-physics systems, such as phase-field models \cite{moelans2008introduction, wang2023modeling, hakimzadeh2022phase, hakimzadeh2025crack}, which require high-resolution computations near evolving interfaces. For instance, in phase-field fracture mechanics, mesh refinement is necessary at the crack tip, where localized stress singularities demand high resolution, which requires significant increase in computational cost. Therefore, adaptively replacing the crack-tip region with DeepONet, as proposed in \cite{goswami2022physics}, can significantly improve simulation efficiency.

\subsection{Neural operator coupling as surrogate of adaptive mesh refinement: An essential key in multiscale modeling}
\label{thick_cylinder}

In preceding examples, our framework has been validated using domains with uniform mesh refinement to demonstrate the efficiency of the spatial and temporal coupling. This section extends our investigation to applications with non-uniform mesh density, a critical consideration for multiscale modeling problems. Adaptive mesh refinement (AMR) allows for selectively increasing the resolution in regions exhibiting fine-scale features (smaller length scale), such as stress concentrations near crack tips in fracture mechanics. However, both the identification of regions requiring refinement and the subsequent numerical solution of these densely meshed domains incur substantial computational costs.

Our approach addresses this challenge by strategically replacing computationally intensive fine-mesh regions with neural operator surrogates. The current implementation demonstrates this capability for static loading conditions with predetermined location of the ML-subdomain (i.e. we are aware of the locations that require refinement). This represents a foundational step toward our future research objective: developing a fully adaptive framework for static and dynamic problems where the neural operator subdomain automatically expands to encompass evolving regions that would otherwise require progressive mesh refinement.

\begin{figure}[H]
\begin{center}
\includegraphics[width=1\textwidth, height=10cm]{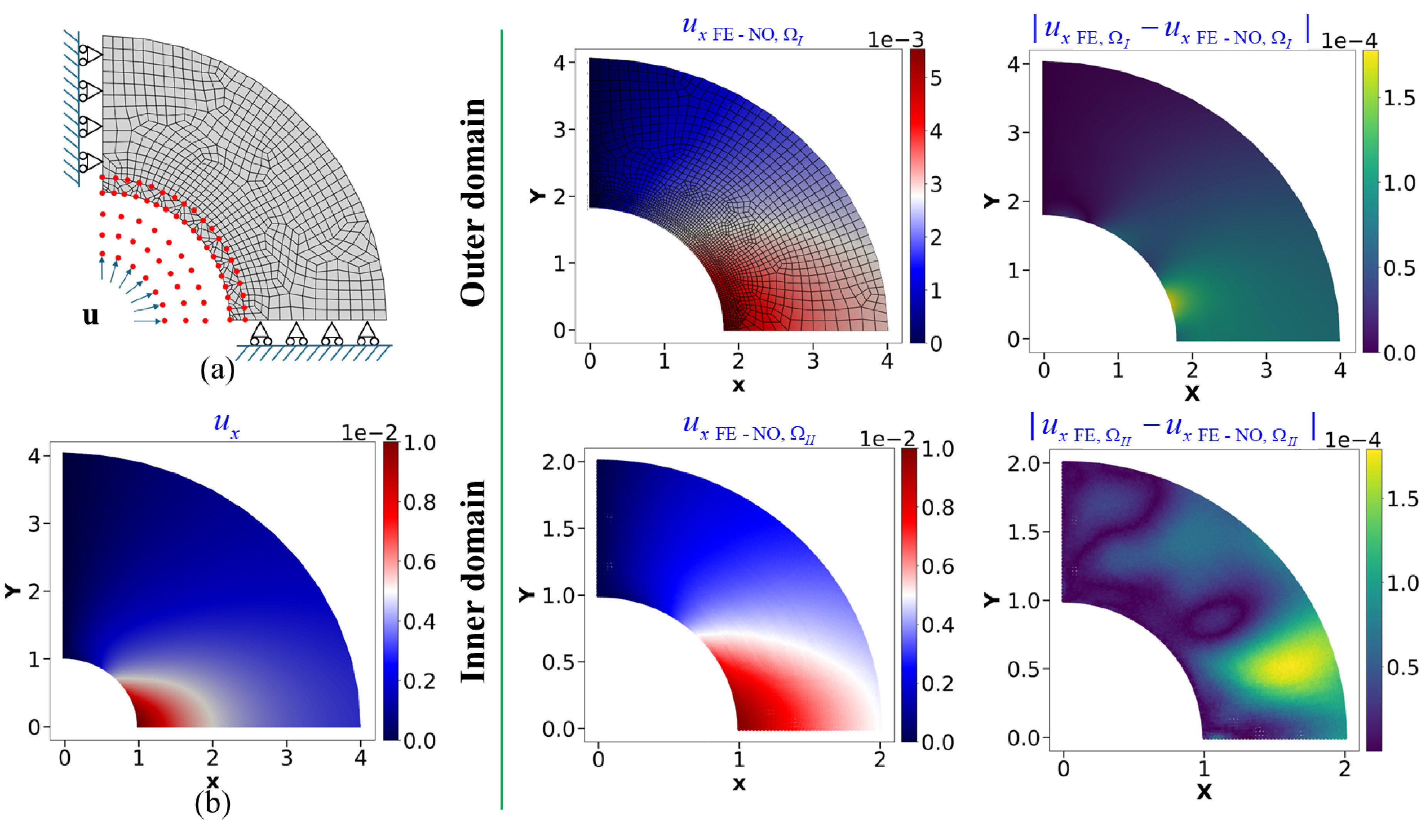}
\caption{Response in $x$-direction (\review{$u_x$}) of the linear elastic coupling model for thick cylinder: (a) Schematic of decomposed domains for the spatial coupling framework, where the left and bottom edge have support roller boundary conditions and inner circular edge is subjected to an applied displacement $\textbf{u} = (0.01\times x^2, 0.01 \times y^2 )$; (b) Ground truth displacement \review{$u_x$} obtained by solving the intact domain using FEniCSx. Middle column shows the \review{$u_x$} in $\Omega_I$ (first row) and $\Omega_{II}$ (second row) for FE-NO coupling at iteration $j = 19$, the last column displays the absolute error relative to the ground truth.}
\label{Fig:cylinder_u}
\end{center}
\end{figure}

\begin{figure}[H]
\begin{center}
\includegraphics[width=1\textwidth, height=10cm]{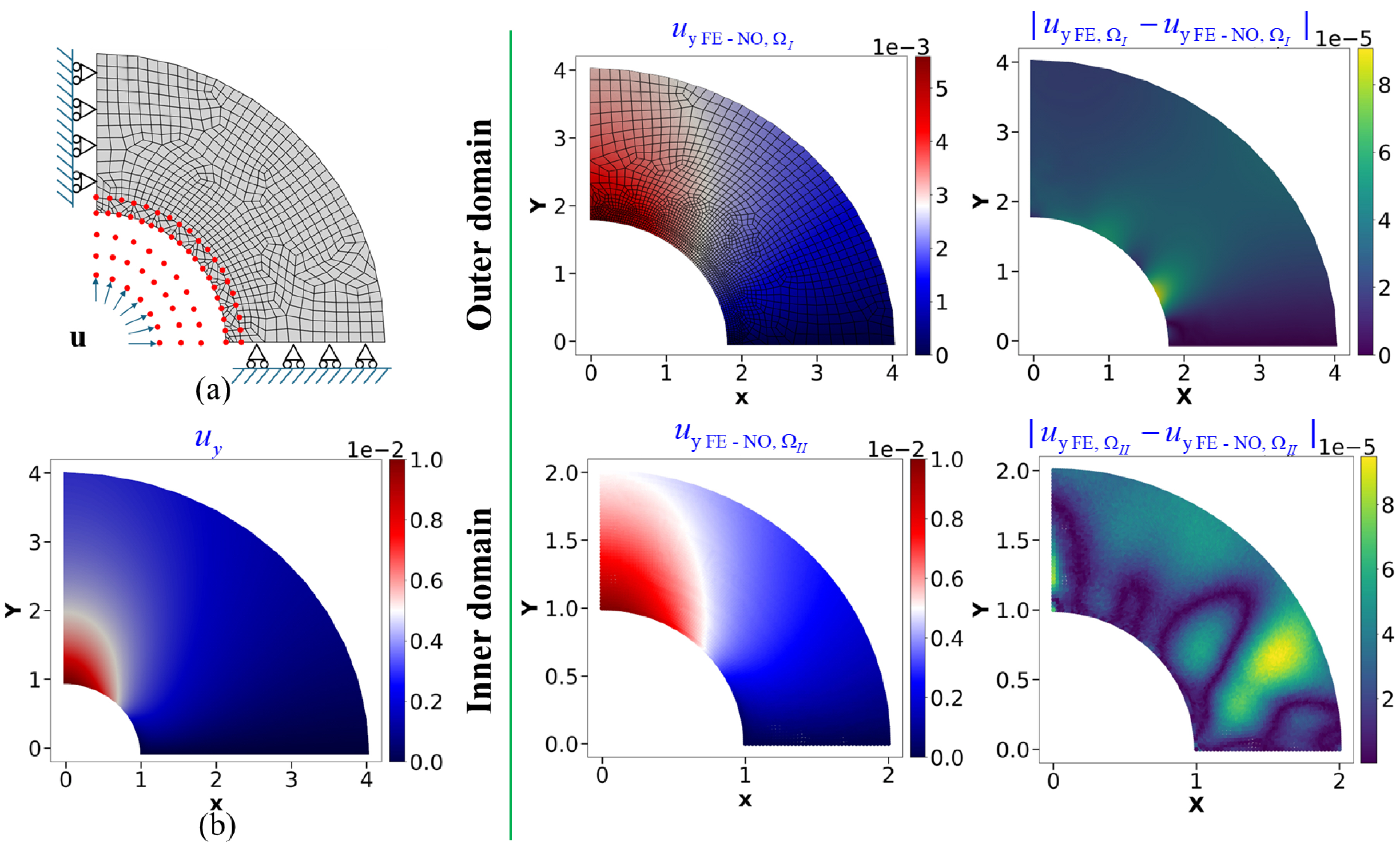}
\caption{Response in $y$-direction (\review{$u_y$}) of the linear elastic coupling model under static loading conditions: (a) Schematic of decomposed domains for the spatial coupling framework, where the left and bottom edge have support roller boundary conditions and inner circular edge is subjected to an applied displacement $\textbf{u} = (0.01\times x^2, 0.01 \times y^2 )$; (b) Ground truth displacement \review{$u_y$} obtained by solving the intact domain using FEniCSx. Middle column shows the \review{$u_y$} in $\Omega_I$ (first row) and $\Omega_{II}$ (second row) for FE-NO coupling at iteration $j = 19$, the last column displays the absolute error relative to the ground truth.}
\label{Fig:cylinder_v}
\end{center}
\end{figure}

In this subsection, we examine a linear elastic thick cylinder with an inner radius of 1 and an outer radius of 4 under static loading conditions. A non-uniform mesh with varying mesh densities is used in this example: the coarse-mesh region is modeled using the FE, while the fine-mesh region is modeled using the neural operator.
To simplify the analysis, the full cylinder domain is reduced to a quarter-cylinder by imposing roller support boundary conditions along the left and bottom edges. The schematic of the decomposed domains for the FE-NO coupling approach is shown in Figure \ref{Fig:cylinder_u}(a). The outer cylindrical region, $\Omega_I$, extending from a radius of 1.8 to 4, is discretized and solved using the FE method, while the inner cylindrical region, $\Omega_{II}$, spanning a radius of 1 to 2, is resolved using the neural operator. A non-uniform, displacement-controlled loading, $\textbf{u}$, is imposed on the inner circular boundary, $\partial \Omega_{II}$, and is defined as:
\begin{equation}
    \textbf{u} = (0.01\times x^2, 0.01 \times y^2),\text{  } (x,y) \in \partial\Omega_{II} \label{loading_u}
\end{equation}
The resulting displacement fields in the $x$ and $y$ directions, obtained from FE analysis over the entire domain, are presented in Figures \ref{Fig:cylinder_u}(b) and \ref{Fig:cylinder_v}(b), respectively. For training of the PI-DeepONet operators, \review{$G^{u_x}_{\boldsymbol{\uptheta_1}}$} and \review{$G^{u_y}_{\boldsymbol{\uptheta_2}}$}, the loss functions $\mathcal{L}_{\text{bcs}}$ and $\mathcal{L}_{\text{res}}$ are formulated similar to those described in Section \ref{section_static}, following Eq. \ref{static_PI_NO_res0}–\ref{static_PI_NO_bcs1}. Due to the roller support boundary conditions imposed along the left and bottom edges, an additional loss function, $\mathcal{L}_{bcs\_fix}$, is introduced to enforce the following constraints:
\begin{align}
    &G^{u_x}_{\boldsymbol{\uptheta}_1}(0, y) = 0, y\in \partial\Omega_{II} \label{bc_cylidner_fix0} \\
    &G^{u_y}_{\boldsymbol{\uptheta}_2}(x, 0) = 0, x\in \partial\Omega_{II} \label{bc_cylinder_fix1}
\end{align}
The PI-DeepONet was trained for $2 \times 10^6$ iterations, with the corresponding loss plot shown in Fig.\ref{Fig:cylinder_DeepONet}(a). Here, $\mathcal{L}_{bcs\_fix}$ is the loss function defined by Eq.\ref{bc_cylidner_fix0}–\ref{bc_cylinder_fix1}. Notably, $\mathcal{L}_{bcs\_fix}$ can be regarded as a specialized contribution to the broader boundary condition loss $\mathcal{L}_{bcs}$. At each training iteration, 1600 collocation points were randomly sampled across the entire domain to compute the loss functions. This dense sampling strategy ensures that the trained PI-DeepONet achieves sufficient resolution to accurately predict results comparable to those obtained from FE analysis on highly refined meshes.

To evaluate the accuracy of the surrogate model, the trained neural operators were tested under the non-uniform displacement loading $\mathbf{u}$ specified in Eq.\ref{loading_u}. The predictions, \review{${u_x}_{\text{NO}, \Omega_{II}}$} and \review{${u_y}_{\text{NO}, \Omega_{II}}$}, were compared against the FE ground truth solutions. The absolute errors were found to be negligible, with magnitudes of $9.5 \times 10^{-5}$ for \review{$|{u_x}_{\text{FE}, \Omega_{II}} - {u_x}_{\text{NO}, \Omega_{II}}|$} and $8.5 \times 10^{-5}$ for \review{$|{u_y}_{\text{FE}, \Omega_{II}} - {u_y}_{\text{NO}, \Omega_{II}}|$}, as illustrated in the last column of Fig.\ref{Fig:cylinder_DeepONet} \review{in \ref{APP_I}}. Given this high level of accuracy, the trained neural operators are deemed suitable for integration within the FE-NO coupling framework.

As a linear elastic system under static loading, the thick cylinder model follows the same computational methodology and governing equations detailed in Section \ref{section_static}. However, the mesh density distribution differs significantly from those used in Sections~\ref{section_static}, \ref{section_quasi_static}, \ref{section_elasto_dynamic}, and \ref{section_puzzle}, as a non-uniform mesh is employed in this case, in contrast to the uniform mesh density used in the previous examples. Specifically, $\Omega_{I}$ is discretized with a coarser mesh compared to the meshes used in the earlier sections, while $\Omega_{II}$ is resolved with a much finer mesh, as illustrated in Figures~\ref{Fig:cylinder_u}(b) and~\ref{Fig:cylinder_v}(b).

Figure \ref{Fig:cylinder_u} (middle column) shows the converged solution of \review{${u_x^{19}}_{FE-NO}$} for both domains $\Omega_{I}$ and $\Omega_{II}$ in iteration $j = 19$.

% The prescribed loading condition in Eq. \ref{loading_u}, induces a displacement concentration in $u_x$, with the maximum value of $\textbf{u} = (0.01, 0)$ occurring at $(x, y)=(1,0)$. Figure \ref{Fig:cylinder_u}(b) illustrates this concentration pattern along the horizontal axis within the radial range of 1 to 2. 

% The NO in $\Omega_{II}$ successfully captures this displacement concentration. 

The maximum absolute errors, quantified as \review{$|{u_x}_{FE} - {u_x^{19}}_{FE-NO, \Omega_{I}}|$} and \review{$|{u_x}_{FE} - {u_x^{19}}_{FE-NO, \Omega_{II}}|$}, both peak at approximately $1.7 \times 10^{-4}$. This slightly exceeds the error levels observed in Figure~\ref{Fig:cylinder_DeepONet}, reflecting the minor accuracy reduction typically introduced by domain decomposition methods (DDM). Moreover, the largest errors are mostly concentrated near the $\Omega_I$–$\Omega_{II}$ interface, indicating that the dominant contributions to the total error arise from interface-related discrepancies inherent in DDM.

% While some displacement concentration persists along the horizontal direction in $\Omega_{I}$, its magnitude remains relatively small and smoothly distributed, in contrast to the strongly localized concentration observed within the confined region of $\Omega_{II}$.

% This observation justifies the use of coarser meshes in $\Omega_{I}$. 

% coarse mesh resolution remains appropriate for the FE-NO coupling framework. 

Figure~\ref{Fig:cylinder_v} similarly presents the converged solutions for the vertical displacement components, \review{${u_y^{19}}_{\text{FE-NO}, \Omega_{I}}$} and \review{${u_y^{19}}_{\text{FE-NO}, \Omega_{II}}$}, in its middle column.
% Both the reference finite element solution ($v_{FE}$) and the coupled FE-NO solution in $\Omega_{II}$ exhibit the expected displacement gradient at $(x, y) = (0, 1)$, which results from the applied loading conditions. 
The corresponding absolute errors, shown in the last column, demonstrate excellent agreement with the reference solution, with maximum errors of approximately $9 \times 10^{-5}$. This level of accuracy is comparable to that achieved by the standalone NO model in $\Omega_{II}$ without DDM, as discussed previously. These results confirm that the pre-trained NO successfully captures the displacement field in $\Omega_{II}$, further validating the effectiveness of our FE-NO coupling approach.

Previous studies \cite{jia2019adaptive, goswami2020adaptive, hakimzadeh2025phase} have consistently demonstrated that accurate simulation of sharp gradients near cylindrical inner boundaries requires substantial mesh refinement. Adaptive refinement techniques enable appropriate variations in discretization density across subdomains \cite{krishnan2022adaptive, gupta2022adaptive}. However, these methods involve an iterative computational cycle comprising pre-processing, solution, error estimation, marking, refinement, and resolution steps \cite{li1997adaptive, goswami2020adaptive}. This computationally intensive process typically demands multiple iterations before achieving satisfactory accuracy.

The implementation of NOs presents a transformative alternative to conventional refinement methodologies. The example in this subsection indicated that by randomly sampling points throughout the computational domain at each iteration, NOs are capable of predicting solutions on highly dense point grids. They only require that dense coordinates $(x, y)$ be provided as input to the trunk network. 

Therefore, here we briefly demonstrated that our method remains robust across varying mesh densities and does not require mesh conformity, unlike traditional numerical solvers. In our future work, we will deploy the proposed method for studying local behaviour like fracture analysis. \review{In our future work, we aim to extend our framework to tackle multiphysics coupling and phase-field fracture problems. This extension will build upon prior efforts such as \cite{pantidis2025integrated} and \cite{snyder2023domain}, which utilize PINNs or data-driven as surrogates. In contrast, our approach leverages physics-informed neural operators, enabling greater generalization across input functions (loading and boundary conditions), and incorporates a time-marching scheme developed in this work to robustly handle time-dependent systems. A key innovation lies in the integration of velocity fields into the Branch network architecture, which significantly reduces autoregressive error accumulation during long-time simulations-a common limitation of conventional PINN-based models.}

\review{\section{Computational Cost}
\label{sec:compu_cost}}

\review{To illustrate the computational efficiency of our framework, we conducted a breakeven analysis using a quasi-static hyperelasticity example. Table~\ref{tab:breakeven} summarizes the key parameters involved in this comparison between the ML-based solver and the traditional FE solver. All measurements were taken on a single NVIDIA RTX 6000 Ada GPU.}

\begin{table}[ht]
\centering
\caption{Key parameters for the breakeven analysis of the ML-based method versus the traditional FEM solver for quasi-static hyperelastic problem.}
\label{tab:breakeven}
% set both text and rules to blue:
\begin{tabular}{@{}ll@{}}
\toprule
\textbf{Parameter}                                & \textbf{Value} \\ 
\midrule
Offline cost (training)         & 14820\,s     \\
ML inference time per simulation                  & 154\,s         \\
Traditional solver time per simulation            & 197\,s         \\
Breakeven number of simulations                   & 345            \\
\bottomrule
\end{tabular}
\end{table}

\review{In this setup, the physics-informed neural operator was trained entirely using governing PDEs, with an offline training time of 14,820\,s. Once trained, the FE-NO coupling approach required 154\,s per simulation, while the traditional FE-FE solver took 197\,s per simulation. This leads to a breakeven point at approximately 345 simulations—after which the ML-accelerated solver becomes computationally more efficient than the conventional approach. }

\review{While training can be substantially accelerated using multi-GPU or multi-node distributed strategies, we did not pursue such scaling here, as our main novelty lies in the coupling methodology rather than training efficiency. Notably, in our elastodynamics tests, only 100 training samples were used (due to the requirement of including initial velocity fields), and the time to generate these samples was negligible given our physics-informed approach. Moreover, advances in GPU hardware and distributed training promise further reductions in training time in future work.}

\section{Conclusion}
\label{sec:summary}

In this work, we propose a hybrid framework that integrates PI-DeepONet with FE solvers to simulate a range of mechanical systems across multiple regimes, including linear elasticity in the static regime, hyperelasticity in the quasi-static regime, and elasto-dynamics. We develop two specialized DeepONet architectures tailored to these settings-one for static and quasi-static problems, and another for dynamic systems. The dynamic architecture is further enhanced with an innovative time-marching technique that drastically reduces error accumulation observed in traditional ML solvers. Our framework supports efficient domain decomposition, with inter-domain communication facilitated by the Schwarz alternating method across overlapping boundaries. The primary contributions of this research are threefold:\vspace{-6pt}
\begin{enumerate} 
\item We propose a spatial coupling framework that enables a pre-trained ML model to replace localized regions requiring high-resolution discretization. The model is trained using the governing physics of the system, eliminating the need for additional data generation.\vspace{-6pt}
\item We incorporate traditional time discretization techniques from numerical solvers - specifically, the Newmark-Beta method, into the DeepONet architecture. This integration results in a high-accuracy surrogate operator-learning model with significantly reduced error accumulation in dynamical simulations.\vspace{-6pt}
\item We present a dynamic subdomain optimization strategy that adaptively expands machine learning subdomains during simulation, providing significant flexibility for reducing computational costs. \vspace{-6pt}
\end{enumerate}

While our current investigations focus on solid mechanics, this coupling framework can readily extend to other physical systems, potentially transforming computational mechanics in terms of efficiency, scale, and generalization capabilities. The rapid inference capabilities of pre-trained DeepONet models, coupled with their ability to capture complex nonlinear mappings, make the FE-NO framework particularly effective for replacing computationally intensive nonlinear components.

This work represents just the beginning of exploring the potential of hybrid coupling frameworks. Future research directions include: (1) developing automated criteria for adaptive machine learning subdomain evolution based on physical indicators, (2) extending the framework to three-dimensional problems, (3) incorporating uncertainty quantification within the coupling framework, and (4) applying the methodology to coupled multi-physics problems such as fluid-structure interaction. For practical applications, particularly in biological processes where simulations must accommodate diverse geometries and morphologies, DeepONet's capabilities can be further enhanced through transfer learning and agnostic learning techniques. This allows the model to generalize across varying cases with minimal additional training, significantly broadening its applicability across scientific and engineering domains.

\section*{Author contributions}
\noindent Conceptualization: WW, SG \\
Investigation: WW, MH, SG, HHR\\
Visualization: WW, MH, SG\\
Supervision: SG, HHR\\
Writing - original draft: WW\\
Writing-review \& editing: WW, MH, SG, HHR

\section*{Acknowledgements}
The authors (WW and HHR) would like to acknowledge the support by the Hong Kong General Research Fund (GRF) under Grant Numbers 15213619 and 15210622, and by an industry collaboration project (HKPolyU Project ID: P0039303). SG is supported by 2024 Johns Hopkins Discovery Award and National Science Foundation Grant Number 2438193. The authors acknowledge the computational resources provided by the University Research Facility in Big Data Analytics (UBDA) at The Hong Kong Polytechnic University. The authors would like to acknowledge Dr. Yue Yu from Lehigh University for all the insighful discussions.

\section*{Data and code availability}
\noindent All codes and datasets is made publicly available at {\small\url{https://github.com/Centrum-IntelliPhysics/Time-Marching-Neural-Operator-FE-Coupling}.

\section*{Competing interests}
\noindent The authors declare no competing interest

\bibliographystyle{elsarticle-num}  
\bibliography{ref} 

\renewcommand{\thetable}{A\arabic{table}}  
\renewcommand{\thefigure}{A\arabic{figure}} 
\makeatother
\setcounter{figure}{0}
\setcounter{table}{0}
\setcounter{section}{1}
\setcounter{page}{0}
\appendix 

\section{Additional details}
\label{APP_I}

\review{To assess the generalization capability of the neural operator within the inner subdomain, we report the average relative L$_2$ error across five randomly selected, unseen inputs for each physics setting. These results, summarized in Table \ref{tab:ARE} demonstrate that the physics-informed neural operator (PINO) maintains high accuracy, with relative $\ell_2$ errors consistently below 3\% across all test cases. This highlights both the robustness and generalization capability of our hybrid framework.}

\review{We do not report scalar errors over the full domain, as the outer subdomain is solved using the finite element method and treated as the ground truth in our coupling framework. Moreover, the outer subdomain is re-simulated for every downstream task. Accordingly, test errors are evaluated only for the PINO model operating in the inner subdomain—where learning and prediction take place. This is consistent with our objective of validating the PINO's effectiveness specifically in the regions where it replaces traditional solvers. }

\begin{table}[H]
\centering
\caption{\review{Mean relative L$_2$ error of 5 unseen test samples evaluated using the trained physics-informed neural operator.}}
\label{tab:ARE}
{\color{blue}
\begin{tabular}{@{}lc@{}}
\toprule
\textbf{Problem}                                & \textbf{Average Relative Error} \\ 
\midrule
Static Loading (Elastic)                           & 2.17\%       \\
Quasi-static Loading (Hyperelastic)                  & 1.09\%          \\
Elastodynamic Loading            & 1.38\%          \\
\bottomrule
\end{tabular}
}
\end{table}

 \begin{figure}[H]
\begin{center}
\includegraphics[width=1\textwidth, height= 7.5cm]{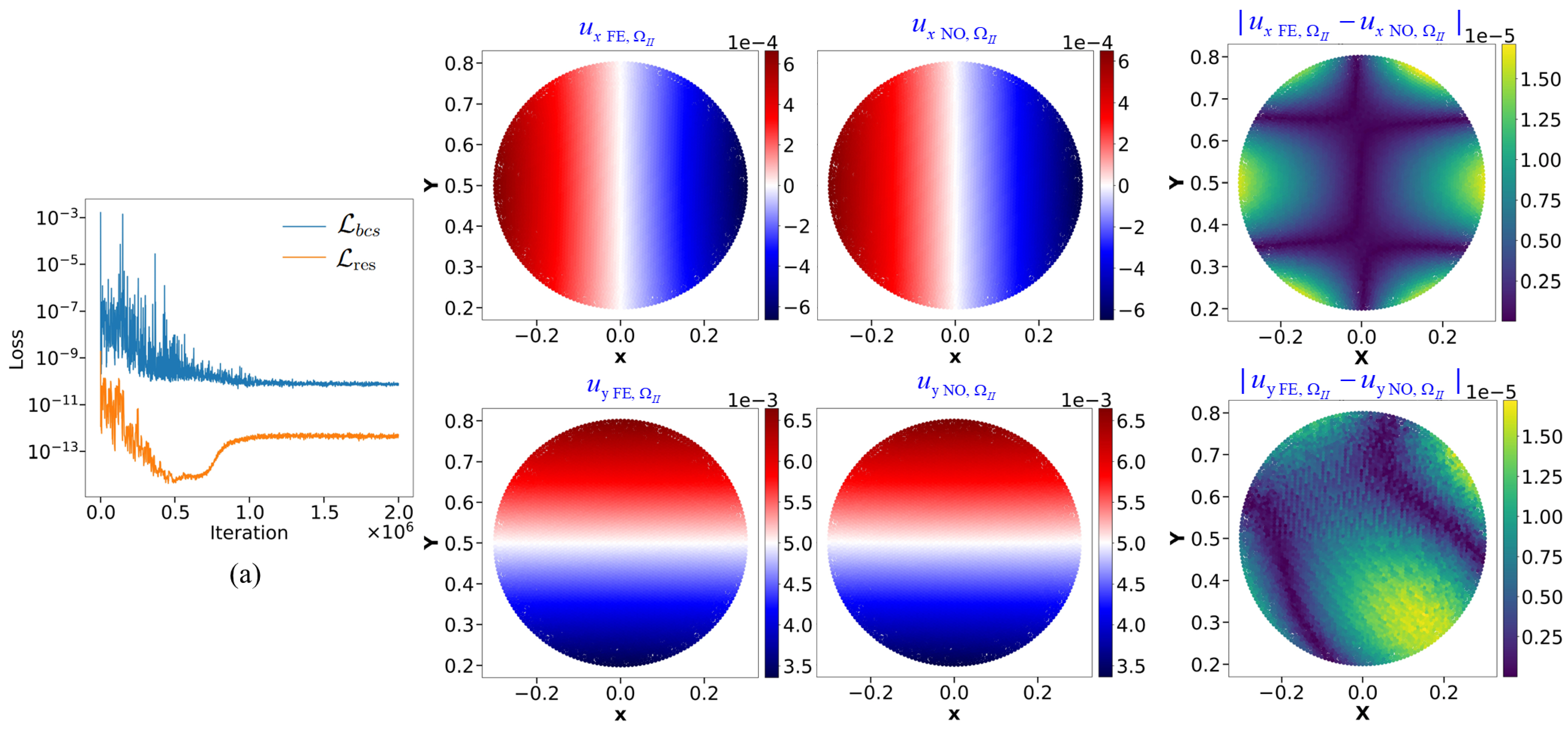}
\caption{Linear-elastic DeepONet pretraining: (a) Loss plot of static PI-DeepONet; The \review{${u_x}_{FE,\Omega_{II}}$} and \review{${u_y}_{FE,\Omega_{II}}$} are the ground truth for $x$-displacement and $y$-displacement in $\Omega_{II}$, using FEniCSx; The \review{${u_x}_{NO,\Omega_{II}}$} and \review{${u_y}_{NO,\Omega_{II}}$} are the prediction from  static PI-DeepONet, along with the absolute error \review{$|{u_x}_{FE, \Omega_{II}} - {u_x}_{NO, \Omega_{II}}|$}  and\review{ $|{u_y}_{FE, \Omega_{II}} - {u_y}_{NO, \Omega_{II}}|$}.}
\label{Fig:static_DeepONet}
\end{center}
\end{figure}

\begin{figure}[H]
\begin{center}
\includegraphics[width=1\textwidth, height= 7.5cm]{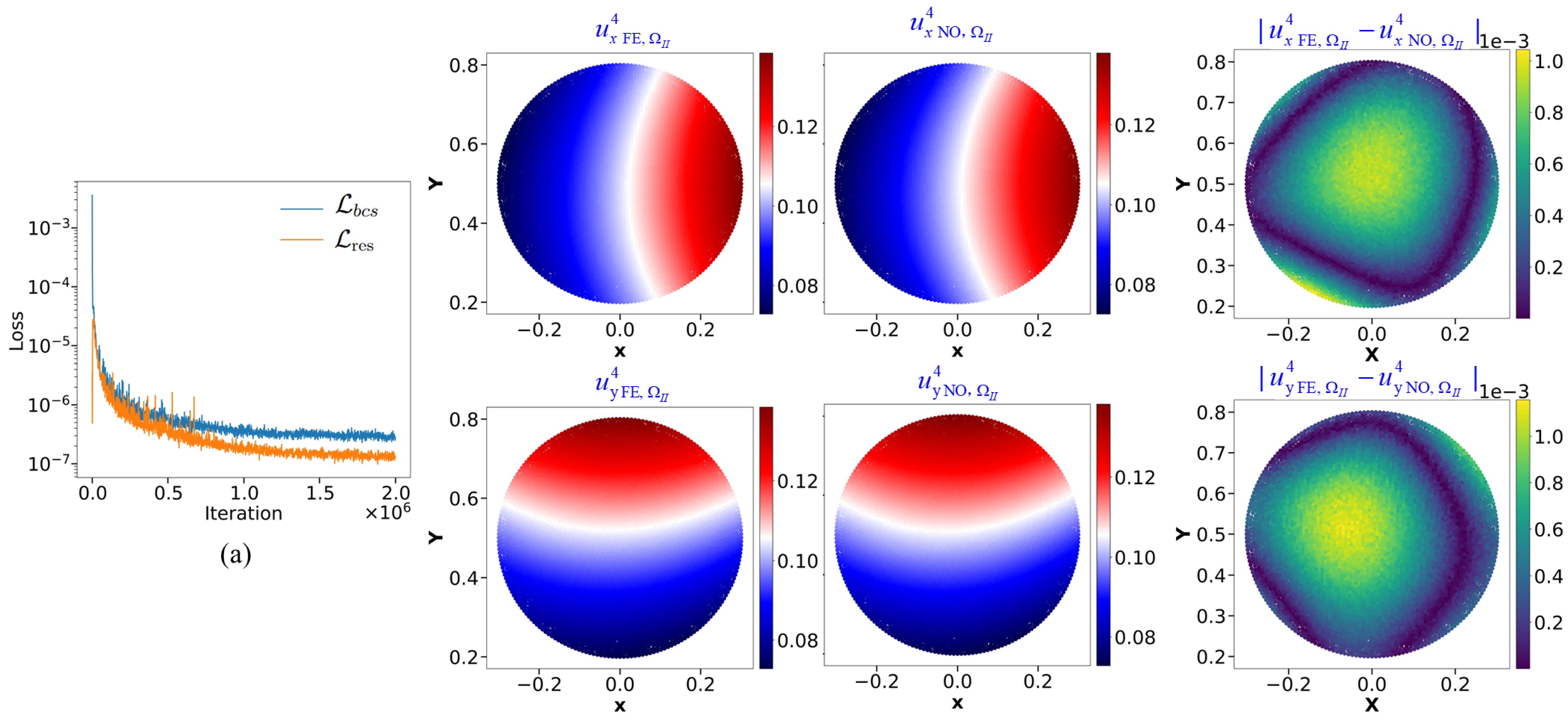}
\caption{Hyperelastic DeepONet pretraining: (a) Loss plot of hyperelastic PI-DeepONet; The \review{${u_x^4}_{FE,\Omega_{II}}$} and \review{${u_y^4}_{FE,\Omega_{II}}$} are the ground truth for $x$-displacement and $y$-displacement in $\Omega_{II}$, using FEniCSx; The \review{${u_x^4}_{NO,\Omega_{II}}$} and \review{${u_y^4}_{NO,\Omega_{II}}$} are the prediction from  static PI-DeepONet, along with the absolute error \review{$|{u_x^4}_{FE} - {u_x^4}_{NO, \Omega_{II}}|$} and \review{$|{u_y^4}_{FE} - {v_y^4}_{NO, \Omega_{II}}|$}.}
\label{Fig:hyper_DeepONet}
\end{center}
\end{figure}

\begin{figure}[H]
\begin{center}
\includegraphics[width=1\textwidth, height= 8cm]{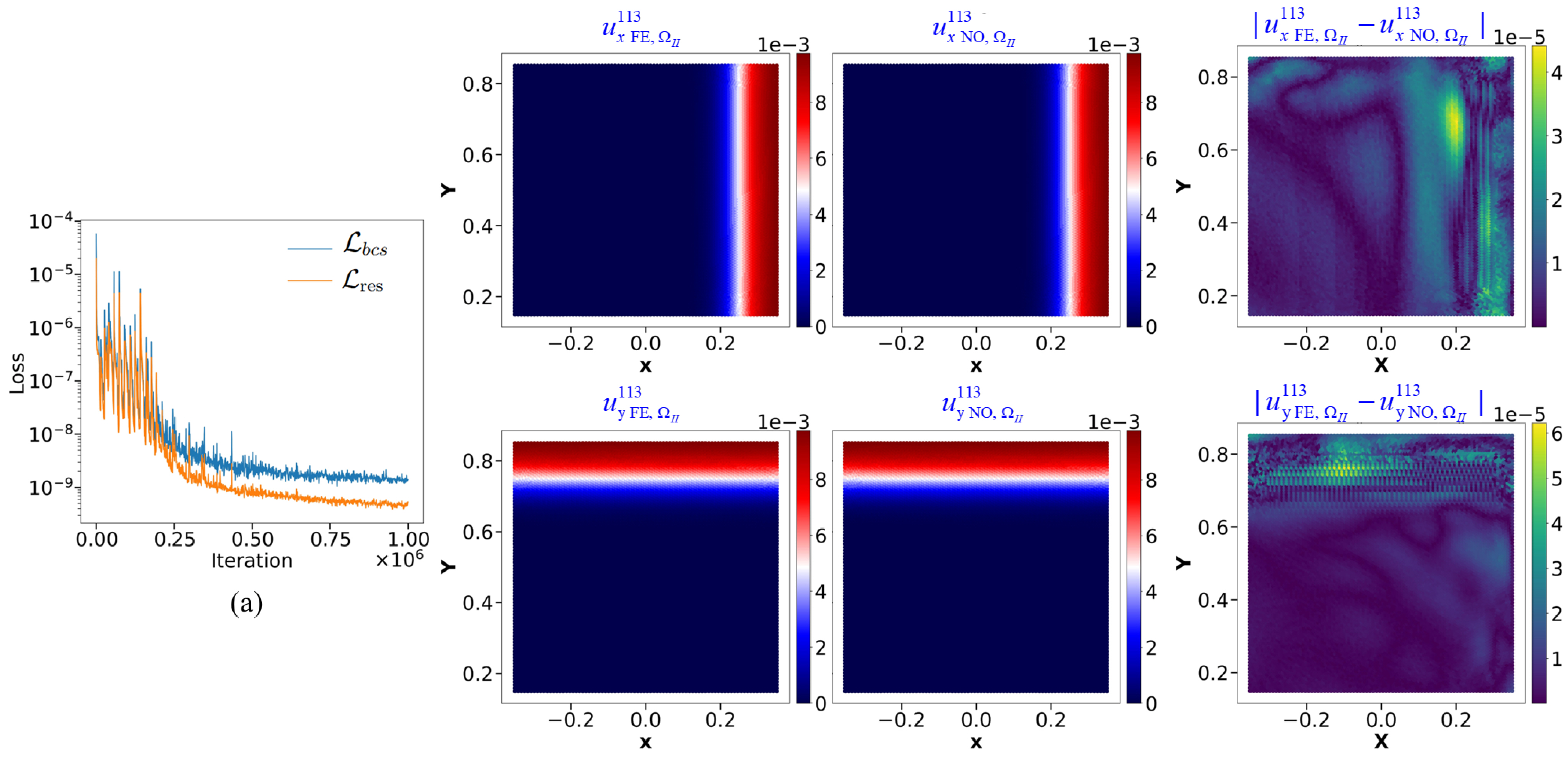}
\caption{Elasto-dynamic DeepONet pretraining: (a) Loss plot of time-advancing PI-DeepONet; The \review{${u_x^{113}}_{FE,\Omega_{II}}$} and \review{${u_y^{113}}_{FE,\Omega_{II}}$} are the ground truth for $x$-displacement and $y$-displacement in $\Omega_{II}$ at time step $n = 113$, using FEniCSx; The \review{${u_x^{113}}_{NO,\Omega_{II}}$} and \review{${u_y^{113}}_{NO,\Omega_{II}}$} are the prediction from time-advancing PI-DeepONet, along with the absolute error \review{$|{u_x^{113}}_{FE, \Omega_{II}} - {u_x^{113}}_{NO, \Omega_{II}}|$}  and \review{$|{u_y^{113}}_{FE, \Omega_{II}} - {u_y^{113}}_{NO, \Omega_{II}}|$}.}
\label{Fig:elasto_dynamic_DeepONet}
\end{center}
\end{figure}

\begin{figure}[H]
\begin{center}
\includegraphics[width=1\textwidth, height= 7.5cm]{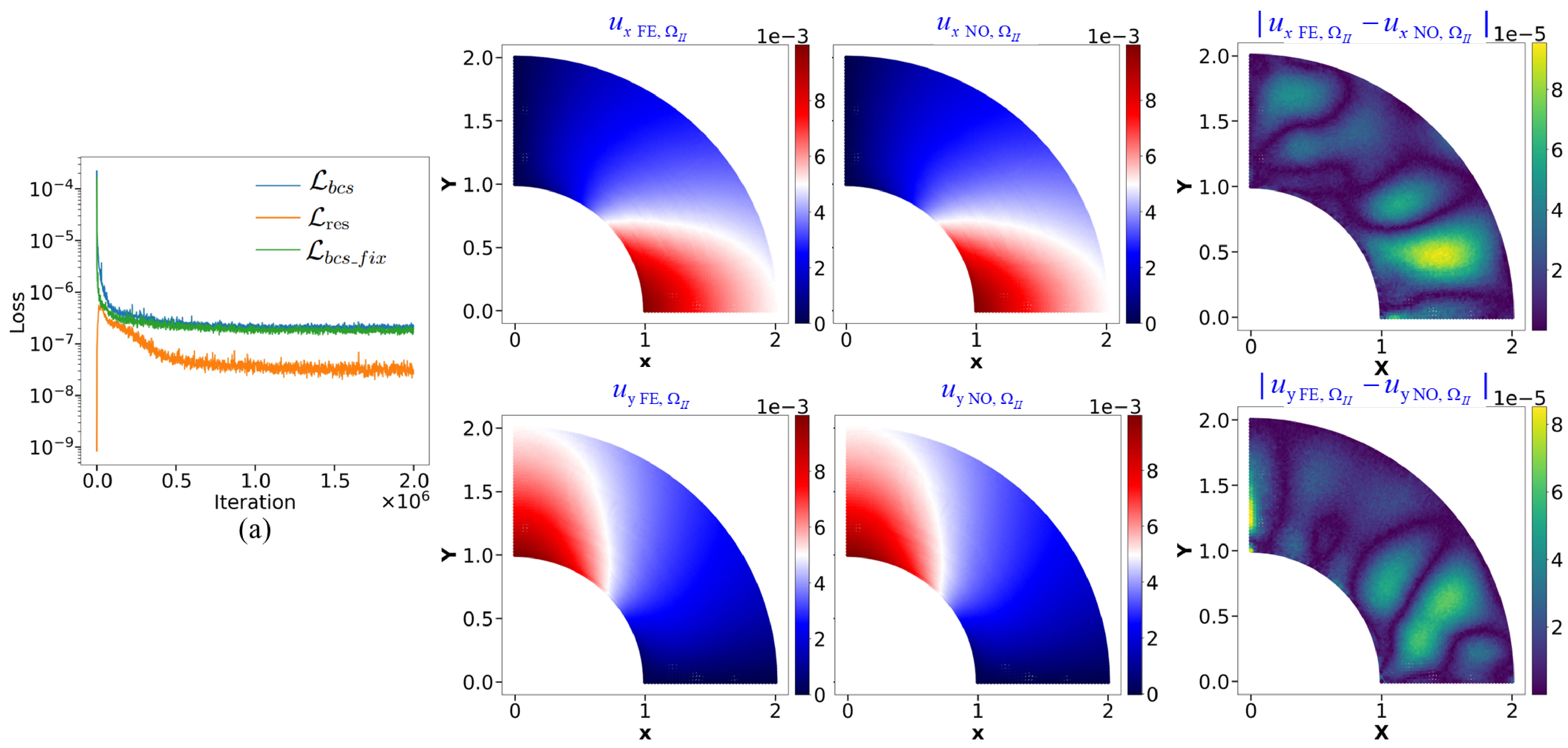}
\caption{Linear-elastic thick cylinder DeepONet pretraining: (a) Loss plot of time-advancing PI-DeepONet; The \review{${u_x}_{FE,\Omega_{II}}$} and \review{${u_y}_{FE,\Omega_{II}}$} are the ground truth for $x$-displacement and $y$-displacement in $\Omega_{II}$ using FEniCSx; The \review{${u_x}_{NO,\Omega_{II}}$} and \review{${u_y}_{NO,\Omega_{II}}$} are the prediction from time-advancing PI-DeepONet, along with the absolute error \review{$|{u_x}_{FE, \Omega_{II}} - {u_x}_{NO, \Omega_{II}}|$}  and \review{$|{u_y}_{FE, \Omega_{II}} - {u_y}_{NO, \Omega_{II}}|$}.}
\label{Fig:cylinder_DeepONet}
\end{center}
\end{figure}

\section{Non-dimensionalized Equations}
\label{APP_II}
\review{Substituting Eqs. \ref{strain} and \ref{stress} into Eq. \ref{static_equa}, the non-dimensional equilibrium equation for a linear elastic material under static loading (with no external force) is derived as:
\begin{equation}
    \frac{(\lambda + \mu)}{E_{el} L} \nabla^*(\nabla^* \cdot \mathbf{u}^*) + \frac{\mu}{E_{el} L} {\nabla^*}^2 \mathbf{u}^* = \mathbf{0} \nonumber
\end{equation}
where $^*$ denotes a non-dimensional quantity or operator; $L$ and $E_{el}$ represent the reference length ($1~\text{cm}$) and reference elastic Young's modulus ($2 \times 10^6~\text{Pa}$), respectively.\\
Similarly, the non-dimensional equilibrium equation under dynamic loading is given by:
\begin{equation}
    \frac{(\lambda + \mu)}{E_{el} L} \nabla^*(\nabla^* \cdot \mathbf{u}^*) + \frac{\mu}{E_{el} L} {\nabla^*}^2 \mathbf{u}^* = \frac{L^3 \rho}{m_0 t_0^2} \ddot{\mathbf{u}}^* \nonumber
\end{equation}
where $m_0 = 1~\text{g}$ and $t_0 = 4 \times 10^{-4}~\text{s}$ are the reference mass and reference time, respectively.\\
For quasi-static loading at time step $n$ (in the absence of external forces), the non-dimensional form of Eq. \ref{static_equa_P1} is expressed as:
\begin{equation}
    \frac{1}{L} \nabla^* \cdot \left[ \frac{\mu}{E_{hp}} (\mathbf{I} + \nabla^* \mathbf{u}^{n*}) + \frac{\left( \lambda \log J - \mu \right)}{E_{hp} J} \text{Cof}(\mathbf{I} + \nabla^* \mathbf{u}^{n*}) \right] = \mathbf{0} \nonumber
\end{equation}
where $\text{Cof}(\cdot)$ denotes the cofactor matrix, and $E_{hp} = 1~\text{Pa}$ is the reference hyperelastic Young's modulus.}

\end{document}